\documentclass[a4paper,11pt]{article} 

\usepackage{lmodern}
\usepackage[top=20mm, bottom=20mm, left=15mm, right=15mm]{geometry}

\usepackage{amsmath}
\usepackage{amssymb}
\usepackage{amsfonts}
\usepackage{amsthm}
\usepackage{bm}           
\usepackage{graphicx}
\usepackage{float}

\usepackage{url}
\usepackage{tabularx}
\usepackage{titling}
\usepackage{longtable} 
\usepackage{booktabs}  
\usepackage{here}      
\usepackage{pdflscape} 
\usepackage{geometry}  
\usepackage{lscape}    

\usepackage{tcolorbox}
\tcbuselibrary{skins}
\usepackage{tikz}
\usepackage{subcaption}
\newsavebox{\mytablebox}

\usepackage[unicode=true]{hyperref}
\hypersetup{
    colorlinks=true,
    linkcolor=blue,
    citecolor=blue,
    urlcolor=blue,
    bookmarksnumbered=true
}

\title{\large Beyond LLMs, Sparse Distributed Memory, and Neuromorphics\\[1em]
\small $<$ A Hyper-Dimensional SRAM-CAM "VaCoAl" \\
for Ultra-High Speed, Ultra-Low Power, and Low Cost $>$
}
\author{
\small Hiroyuki Chuma (Professor Emeritus, Hitotsubashi University)\thanks{Corresponding author. While the source code remains proprietary, the implementation is available for commercial use and collaborative projects. For inquiries, please contact chuma@iir.hit-u.ac.jp.} \\[1ex]
\small Kanji Otsuka (Professor Emeritus, Meisei University) \\[1ex]
\small Yoichi Sato (Shuhari System)
}

\date{\small \today}
\usepackage{microtype}

\begin{document}
\setlength{\parindent}{2em}
\maketitle
\tableofcontents
\newpage

\begin{abstract}
\renewcommand{\thefootnote}{\alph{footnote}}
\setlength{\parindent}{2em}

\noindent\hspace{2em}
This paper reports an unexpected finding. In a deterministic
hyperdimensional computing architecture built on Galois-field algebra,
a path-dependent semantic selection mechanism emerges spontaneously,
without being part of the original design. It is functionally
equivalent to Spike-Timing-Dependent Plasticity (STDP) in biological
neural circuits, and its magnitude is predictable \emph{a priori} from
a closed-form expression that matches the measured value on real data
at the scale of tens of millions of records. The finding arises from
an architecture designed to address the structural limitations of
modern AI---catastrophic forgetting, learning stagnation, and above
all the Binding Problem in which information is irreversibly mixed
and the relationships between constituent elements cannot be strictly
maintained---at the algebraic level rather than the statistical one.

To break through these limitations, we propose a new architecture, ``VaCoAl (Vague Coincident Algorithm),'' employing an ultra-high-speed, low-power, low-cost ultra-high-dimensional SRAM/DRAM-CAM, together with its Python implementation ``PyVaCoAl.''  Rooted in Pentti Kanerva's Sparse Distributed Memory (SDM), this architecture solves the formidable challenges of orthogonalisation and retrieval in ultra-high (here, one million)-dimensional binary spaces---which SDM faced in physical implementation---through algebro-deterministic digital logic employing Galois-field diffusion.  

Consequently, the massive computational overhead inherent in conventional Hyperdimensional Computing (HDC) is eliminated, successfully realising practical HDC at extremely low load. This provides a crucial engineering foundation for next-generation neuro-symbolic AI that integrates logical symbolic reasoning with statistical learning. Moreover, the system operates as a ``memory-based architecture'' that prioritises retrieval and association over computation, and its block-division-and-majority-voting structure bears a structural analogy to Dendritic Computation in biological pyramidal neurons.

Through this algebraic structure, VaCoAl/PyVaCoAl addresses the Binding Problem in a fundamentally different way from the statistical embeddings of LLMs.  HDC-type algebraic operations enable the reversible integration and decomposition of concepts at extremely low computational cost while preserving element independence, thereby achieving compositional generalisation.  Simultaneously, it realises Explainable AI (XAI) equipped with ``mathematical auditability''---allowing post-verification of the computational process without information degradation---and presents a highly transparent reliability metric, the ``Confidence Ratio (CR) score,'' based on algebraic majority voting.

To verify VaCoAl's search performance and multi-step semantic reasoning capabilities at a practical scale, we conducted empirical HDC experiments using PyVaCoAl.  The true value of this architecture is demonstrated not in simple similarity searches but in multi-hop associative reasoning over massive directed acyclic graphs (DAGs).  As a benchmark designed to push this property to its limits, we utilised an ontology database of approximately 470,000 scholar mentor-student relationships over a millennium constructed from WIKIDATA, and executed a large-scale Computational Prosopography: starting from all 64~historical Fields Medalists, we traced the genealogies of mentors back up to 57~generations, equivalent to over 25.5~million total genealogical paths.

In the HDC analysis, we synthesised composite concept vectors such as ``calculus degree,'' ``Giant Score,'' etc. using Bundling operations specific to HDC, continuously measured the degree to which various concepts are embedded in each node through Unbinding operations and Hamming distance, and applied the CR score to eliminate noise.  The principal findings are as follows:
\begin{itemize}
    \item A historical reconsideration of the Newton--Leibniz priority dispute over the founding of calculus.
    \item Quantitative visualisation of the discontinuous dynamics whereby diverse academic fields up to the 17th~century converge on Leibniz as numerous narrow paths averaging approximately 5~``routes passing through the person,'' while tracing from Leibniz to the present reveals a ``superhighway'' to Fields Medalists expanding to an average of approximately 53~paths.
    \item Structural demonstration, via HDC analysis employing seven independent lines of evidence---emergence of the Super-Highway, explosive growth of society memberships, hub diversification of research institutions, surge in student production rate, among others---of a Thomas Kuhn~style paradigm shift (the formation of a new normal science) that occurred in the wake of Leibniz.
\end{itemize}

Furthermore, comparing the presence and absence of the collision-avoidance (rescue) circuit built into PyVaCoAl revealed that VaCoAl's distinctive Collision Tolerance (Don't Care) mechanism triggers, in multi-stage reasoning, a historical culling process mediated by path-integral values.  This mechanism functions as an ``Occam's razor'' that naturally prunes circuitous routes and selectively preserves direct ones, giving rise on real data to an ``emergent semantic selection'' functionally equivalent to STDP (Spike-Timing-Dependent Plasticity) in the brain.  VaCoAl thereby opens a decisive third path as ``HDC-AI''---one that transcends a mere high-speed search engine. Far from rivalling Large Language Models, this third path complements them by supplying the reversible, auditable reasoning substrate they structurally lack. The result is a natural division of labour for next-generation neuro-symbolic AI: linguistic fluidity on one side, multi-hop logical integrity on the other.

\end{abstract}


\section{Introduction: From Probabilistic Mimicry to 
         ``Analog--Digital (AD) Functional Induction''}

\subsection{Resolving the ``Singularities'' of Modern Generative AI
            and Transcending Kanerva's Theory}

The success of deep learning is undeniable, yet a fundamental inefficiency
is hard-wired into its learning process.  Modern generative AI models are
built predominantly on McCulloch--Pitts point-neuron models whose tens to
hundreds of billions of parameters are tuned by gradient descent through
backpropagation.  Amari's information-geometry theory, however, reveals that
\emph{singularities} in the hierarchical parameter space of deep neural
networks degenerate the Fisher information matrix and induce severe
\emph{learning plateaus}~\cite{Amari1998, Amari2006}.  Remedies such as
Amari's natural-gradient method demand prohibitively large inverse-matrix
computations, rendering them impractical for edge devices.  Moreover, any
attempt to augment a pre-trained model with new knowledge risks
\emph{catastrophic forgetting}---the overwriting and destruction of
previously acquired memories---as formalized by Kirkpatrick et
al.~\cite{Kirkpatrick2017}.%
\footnote{The AAAI presidential panel report~\cite{AAAI2025} provides a
comprehensive survey of the more fundamental limitations of LLMs.  Its
observations are particularly germane to the present work, because the
VaCoAl architecture proposed here enables multi-hop semantic logical
inference---strictly tracing multi-step relationships---as algebraic
operations, precisely the domain in which current AI falters:
``\emph{Causal and Counterfactual Reasoning:} While AI models can detect
correlations in vast datasets, they struggle with causal inference and
counterfactual reasoning.''}

Beyond these learning-process limitations, modern LLMs harbor a deeper
structural deficiency rooted in the irreversible mixing of information:
the \emph{Binding Problem}.  As Thagard~\cite{Thagard2019} and
Anderson~\cite{Anderson2017} emphasize, this is a formidable
implementation challenge for any brain-inspired cognitive architecture.
In the statistical embeddings of LLMs, constituent elements fuse
irreversibly when multiple concepts are integrated, making it impossible
to preserve their compositional relationships faithfully.  The
\emph{Semantic Pointer Architecture} (SPA) of
Plate~\cite{Plate1995} and Eliasmith~\cite{Eliasmith2013} offers a
mathematically principled solution, but its reliance on circular
convolution entails heavy computation that severely hampers deployment on
low-power devices.

In 1988, Pentti Kanerva proposed a fundamentally different approach:
\emph{Sparse Distributed Memory}
(SDM)~\cite{Kanerva1988}.  By representing brain input/output as
vectors in an ultra-high-dimensional space, Kanerva exploited their
near-orthogonality to sidestep information interference.  The VaCoAl
architecture proposed herein is conceptually rooted in Kanerva's SDM yet
surpasses both SDM and SPA by leveraging modern digital-circuit
technology.  At its core lies algebro-deterministic Galois-field
diffusion: as Section~5 demonstrates, VaCoAl employs HDC-type algebraic
operations---Binding, Unbinding, and Bundling---to reversibly compose
and decompose concepts while preserving element independence, thereby
resolving the Binding Problem at a fraction of the computational cost
required by SPA.  The Galois-field diffusion mechanism is detailed in
Section~3.

\subsection{The Engineering Barriers of ``Randomness'' and the Limits
            of Biomimicry}

Kanerva's theory was elegant in the abstract, but translating it into
silicon exposed a formidable engineering obstacle: the \emph{address-decoder
problem}.  This obstacle originates in the probabilistic randomness and
physical biomimicry that underpin both SDM and mainstream neuromorphic
hardware (e.g., IBM's TrueNorth~\cite{Merolla2014}, Intel's
Loihi~\cite{Davies2018}).  Early hardware prototypes based on ternary
CAM (TCAM) attempted brute-force parallel matching to emulate the
brain's ability to locate similar patterns instantaneously, but the
resulting power dissipation rendered them thermally
unscalable~\cite{Pagiamtzis2006}.

Physical neuron mimicry alone, moreover, cannot address higher-order
cognitive tasks.  As Thagard argues in \textit{Brain--Mind}%
~\cite{Thagard2019}, a cognitive architecture capable of explaining
creativity, consciousness, and emotion must solve the Binding
Problem---the challenge of composing hierarchically structured thoughts
from elementary concepts such as ``red'' and ``apple.''
Anderson~\cite{Anderson2017} adds that human cognition is fundamentally
\emph{associative} rather than search-based, underscoring the need to
engineer the brain's capacity to crystallize robust concepts from noisy
sensory streams.

Mathematically sophisticated frameworks such as the SPA of
Plate~\cite{Plate1995} and Eliasmith~\cite{Eliasmith2013} address
these requirements in principle, yet their dependence on computationally
expensive operations (e.g., circular convolution) precludes practical
deployment on low-power devices.  A lightweight implementation that can
solve the Binding Problem algebraically---precisely what VaCoAl
provides---is therefore indispensable.  In short, the conventional
strategy of physically mimicking the brain's probabilistic behavior has
itself constituted the \emph{missing link} between neuroscience-inspired
theory and engineering practicality.

\subsection{The Core of This Study: ``AD Functional Induction''}

\begin{table}[H]
  \centering
  \caption{Comparison of analog neuromorphic implementations and VaCoAl.}
  \label{tab:neuro-vs-vacoal}
  \begin{tabular}{lll}
    \toprule
    \textbf{Comparison Axis}
      & \textbf{Analog Neuromorphic}
      & \textbf{VaCoAl} \\
    \midrule
    Orthogonalization
      & Spike timing
      & Primitive-polynomial modular arithmetic \\
    Error correction
      & Attractor dynamics
      & Algebraic majority voting \\
    Auditability
      & Difficult
      & Traceable via CR2 path integral \\
    Thermal-noise tolerance
      & Device-variation dependent
      & Irrelevant (digital logic) \\
    Demonstrated scale
      & Predominantly small-scale
      & 25.5\,M records (this study) \\
    \bottomrule
  \end{tabular}%
\end{table}

The purpose of this study is to demonstrate that VaCoAl constitutes a
decisive solution to the missing link identified above.  Rather than
relying on probabilistic random numbers or analog components, VaCoAl
achieves the orthogonalization envisioned by Kanerva and Amari and
resolves the Binding Problem raised by Thagard and Anderson through
strictly deterministic digital logic---specifically, diffusion over
Galois fields (BCH codes)~\cite{Lin2004}.  Instead of mimicking the
physical structure of biological neurons, we independently solve the
same computational problems (high-dimensional orthogonalization, pattern
completion) via Galois-field algebra.  We term this approach
\textbf{AD Functional Induction}; Table~\ref{tab:neuro-vs-vacoal}
summarizes the principal structural differences from existing
neuromorphic approaches.

\paragraph{Main contributions.}
This paper makes the following contributions to the fields of memory
architecture, neuromorphic engineering, and neuroscience.

\begin{itemize}
\item \textbf{Discovery of emergent semantic selection via the Don't 
Care mechanism.}\enspace
\textbf{Most strikingly}, by comparing system behavior with and
without VaCoAl's multi-stage collision-avoidance (rescue) circuit, we
uncover an \textbf{unintended emergent property} of the Don't Care
(collision-tolerance) mechanism: in multi-hop reasoning it imposes an
exponential decay penalty on deep paths, naturally pruning circuitous
routes and retaining direct ones---an ``Occam's razor'' effect. This
behavior is functionally analogous to STDP [16] in biological neural
circuits, constituting an emergent semantic selection that was not
part of the original design, \textbf{and its magnitude is predictable
\emph{a priori} from the closed-form expression
$0.997^{56}\!\approx\!0.846$, in close agreement with the measured
value $\mathrm{CR}_2 \approx 0.905$ at the 56th generation}.

\item \textbf{Discovery of emergent semantic selection via the Don't
  Care mechanism.}\enspace
  By comparing system behavior with and without VaCoAl's multi-stage
  collision-avoidance (rescue) circuit, we uncover a hidden property of
  the Don't~Care (collision-tolerance) mechanism: in multi-hop
  reasoning it imposes an exponential decay penalty on deep paths,
  naturally pruning circuitous routes and retaining direct ones---an
  ``Occam's razor'' effect.  This behavior is functionally analogous to
  STDP~\cite{Kandel2021} in biological neural circuits, constituting
  an \emph{emergent} semantic selection that was not part of the
  original design.

\item \textbf{Resolution of the Binding Problem and realization of
  Explainable AI (XAI).}\enspace
  Whereas the SPA of Plate~\cite{Plate1995} and
  Eliasmith~\cite{Eliasmith2013} requires circular convolution, we show
  that VaCoAl resolves the Binding Problem with lightweight HDC-type
  algebraic operations (Binding/Unbinding).  Reversible concept
  manipulation preserves element independence, overcomes the irreversible
  mixing inherent in LLM embeddings, and achieves compositional
  generalization.  The architecture further provides \emph{mathematical
  auditability} for post-hoc verification and a transparent reliability
  metric---the \emph{Confidence Ratio (CR) score}---derived from
  algebraic majority voting.

\item \textbf{Ultra-high-dimensional CAM realized as a digital circuit
  via algebro-deterministic logic.}\enspace
  \quad We propose replacing the probabilistic random projections of SDM with
  VaCoAl's Galois-field
  operations~\cite{Otsuka2014, Otsuka2019, Otsuka2021, Sato2019, Sato2025a, Sato2025b}.
  Combined with a design philosophy of \emph{$O(1)$-like diffusion plus
  voting over fixed-size blocks}, the approach eliminates the massive
  address-decoder overhead and delivers constant-time access independent
  of data scale, dramatically improving the feasibility of edge-AI
  deployment.

\item \textbf{Physical realization of a memory-based architecture on
  SRAM/DRAM-CAM.}\enspace
  We give engineering substance to the hypothesis of
  Matsumoto~\cite{Matsumoto2003} and Anderson~\cite{Anderson2017} that
  the brain operates as a memory-based architecture---internalizing
  learning algorithms into physical circuit structure---by constructing
  ultra-high-dimensional digital look-up tables (LUTs) on SRAM and
  DRAM.

\item \textbf{Large-scale Computational Prosopography on an
  ultra-large ontology DB.}\enspace
  Using PyVaCoAl and a mentor--student relationship dataset of
  ${\sim}470{,}000$ mathematicians extracted from WIKIDATA, we trace
  mentor lineages of all 64~historical Fields Medalists back up to
  57~generations, totaling over 25.5~million records.  We define this
  approach as \textbf{Computational Prosopography}: a data-driven method
  that goes beyond graph traversal to uncover the propagation structures
  and dynamics of knowledge across historical populations.  On this
  foundation we perform continuous measurement of
  \emph{calculus affinity} via HDC Unbinding and Hamming distance,
  quantitatively visualizing the Kuhnian paradigm
  shift~\cite{Kuhn2012} in which diverse disciplines converge at hub
  nodes such as Leibniz to form a modern-mathematics
  ``superhighway.''  Evidence-based answers to these complex
  questions---unattainable with standard hash tables---are derived in
  tens of minutes.

\end{itemize}

\subsection{Structure of the Paper}

The remainder of the paper is organized as follows.

Section~2 reviews Kanerva's SDM theory and Amari's
orthogonality framework~\cite{Amari1989}, connects them to the
Galois-field diffusion employed by VaCoAl, and analyzes the
implementation bottlenecks that have constrained conventional
ultra-high-dimensional CAMs~\cite{Pagiamtzis2006}.

Section~3 details the VaCoAl mechanism, showing how an ideal
best-match search engine can be constructed by fusing
algebro-deterministic Galois fields with probabilistic distributed
representations.  We demonstrate that, even at big-data scale, a novel
multi-stage collision-avoidance circuit deterministically achieves
Frontier-Size-relative exact matching while retaining HDC's flexible
retrieval capabilities.  We also delineate the qualitatively distinct
characteristics and trade-offs of the two operating modes: \emph{Rescue
mode} (DICT-equivalent perfect reproducibility) and \emph{Don't~Care
mode} (CR2-based semantic selection).

Section~4 evaluates VaCoAl's computational complexity and energy
efficiency relative to existing approaches, and discusses how the
Binding Problem identified by Anderson~\cite{Anderson2017} and
Thagard~\cite{Thagard2019} can be resolved through low-cost algebraic
operations.  The natural affinity between HDC-type operations and
multi-predicate ontology databases is also clarified.

Section~5 empirically validates the theory at scale.%
\footnote{Specifically, information from the database is evaluated as
follows.  \textbf{(i)}~For each binarized attribute, a \emph{Binding}
operation associates its label (role) with its value (content).
\textbf{(ii)}~These bound pairs are integrated via \emph{Bundling} to
form reference composite-concept vectors (e.g., a ``calculus'' vector).
\textbf{(iii)}~Each node (person) is likewise encoded as a
high-dimensional vector; to probe a specific concept, an
\emph{Unbinding} operation extracts the target attribute, and the match
with the reference vector from steps~(i)--(ii) is measured by Hamming
distance.  \textbf{(iv)}~Finally, noise is filtered using the resulting
CR score.}
Using a mentor--student ontology of ${\sim}470{,}000$ scholars from
WIKIDATA, we execute Computational Prosopography spanning up to
57~generations and over 25.5~million records, starting from all
historical Fields Medalists.  HDC operations quantitatively visualize
the formation of the Leibniz-centered ``superhighway'' and the
Kuhnian~\cite{Kuhn2012} paradigm-shift dynamics.  A comparison with and
without the rescue circuit further demonstrates that VaCoAl's Don't~Care
mechanism triggers STDP-like decay, functioning as Occam's razor to
prune deep paths---an emergent semantic-selection phenomenon.

Finally, the appendices supply supporting detail:
\begin{enumerate}
  \item Appendix~A: a mathematical proof that the false-positive
    probability vanishes asymptotically.
  \item Appendix~B: reframing the narrative---from ``mimicry'' to
    ``convergent evolution.''
\end{enumerate}


\section{Theoretical Background: Geometric Properties of
         High-Dimensional Orthogonal Spaces
         (``Blessing of Dimensionality'') and Implementation Barriers}

This section reviews the theoretical framework of Sparse Distributed
Memory (SDM) proposed by Kanerva~\cite{Kanerva1988} and analyzes, from
both mathematical and engineering perspectives, why the practical
implementation of its ``ideal address decoder'' has proved so difficult
within conventional computer architectures.

\subsection{Kanerva's Sparse Distributed Memory (SDM) Model}

Kanerva's SDM models human long-term memory not as a deterministic
address structure akin to an organized bookshelf (RAM), but as a cloud
of information points scattered across a vast high-dimensional space and
accessed on the basis of \emph{Hamming distance} (bitwise
similarity)~\cite{Kanerva1988}.  As Kanerva elaborated, SDM is a
mathematical formalization of cerebellar circuitry.  Two cell types play
clearly delineated roles: \emph{granule cells} (corresponding to the
address decoder) sparsely diffuse and encode inputs as long parallel
fibers into a high-dimensional space, while \emph{Purkinje cells}
(corresponding to the storage medium) extend fan-shaped dendrites
orthogonal to those fibers, capturing and storing the resulting
orthogonalized high-dimensional vectors.

The computational significance of this geometric architecture as an
orthogonalization mechanism is well supported by modern neuroscience.
Kandel et al.~\cite{Kandel2021} argue that cerebellar granule cells
\emph{expand} sensory inputs across billions of neurons, thereby
separating similar input patterns in high-dimensional space (pattern
separation) and facilitating learning and discrimination by Purkinje
cells.  Rolls~\cite{Rolls2023} similarly emphasizes that expansion
recoding by granule cells is the key to mathematically orthogonalizing
sensory patterns and enabling interference-free learning.  The
\emph{blessing of dimensionality} articulated by SDM is therefore not
merely an abstract mathematical property but reflects the physical
substrate through which neural circuits represent and learn information.

\subsubsection{Mathematical Properties of High-Dimensional Spaces}

In SDM, all information---external sensory inputs (vision, hearing,
touch) as well as internal states (body position, ongoing thought
processes, context)---is represented as $n$-bit binary vectors
($n \approx 1{,}000$--$10{,}000$).  The resulting space
$\mathcal{N} = \{0,1\}^{n}$ possesses geometric properties that
differ sharply from everyday three-dimensional intuition.

\begin{itemize}
\item \textbf{Orthogonality (blessing of high dimensionality).}\enspace
  Two points drawn uniformly at random from $\{0,1\}^n$ have a Hamming
  distance that follows a binomial distribution centered at $n/2$.
  Consequently, any two random vectors are almost certainly
  quasi-orthogonal; the probability of two such vectors being ``similar''
  is astronomically small.  This geometric property allows the brain
  (and SDM) to superimpose a vast number of patterns on the same
  physical memory substrate without catastrophic
  interference~\cite{Kanerva2009}.

\item \textbf{Enormous capacity.}\enspace
  The binary space contains $2^{n}$ points; for
  $n \approx 1{,}000$--$10{,}000$ this dwarfs the number of atoms in the
  observable universe.  Allocating physical storage for every point is
  therefore infeasible.  Kanerva~\cite{Kanerva2009} instead
  approximates SDM's input/output behavior by randomly placing a finite
  number $N_0$ of \emph{hard locations}
  ($N_0 \approx 10^{6}$--$10^{12} \cong 2^{20}$--$2^{40}$) across the
  space.  The key observations, expressed in modern terms, are:
  \begin{itemize}
  \item \textbf{Sparse post-hoc usage.}\enspace
    A human lifespan spans at most ${\sim}2^{32}$ seconds; hence the
    set of events actually encountered and memorized occupies a
    negligible fraction of the full space.

  \item \textbf{Feasibility with current technology.}\enspace
    Storing $10^{10}$ patterns of $10{,}000$~bits each requires only
    ${\sim}12.5$\,TB ($2^{40}$~bytes $\approx 2^{43}$~bits)---well
    within the reach of commodity SSDs.

  \item \textbf{Mathematical guarantee.}\enspace
    Exploiting the orthogonality of high-dimensional vectors, Kanerva
    proves that a physical memory of $2^{33}$--$2^{40}$~bits suffices
    to process high-dimensional vector information reliably.
  \end{itemize}
\end{itemize}

\subsubsection{Best-Match Mechanism in SDM}

Read and write operations in SDM differ fundamentally from those of
conventional RAM: they rely on \emph{best match} rather than exact
match.

\begin{itemize}
\item \textbf{Distributed write.}\enspace
  To store binary datum~$D$ at address~$x$, the system does not seek a
  single location that matches~$x$ exactly.  Instead, it activates
  \emph{all} hard locations within a prescribed Hamming-distance radius
  of~$x$ and writes~$D$ to each of them.  A single memory trace is thus
  physically distributed across thousands of locations (neurons).

\item \textbf{Statistical read (majority rule).}\enspace
  To retrieve data given a noisy probe~$x'$, the system activates
  nearby hard locations and sums their contents.  High-dimensional
  orthogonality ensures that noise vectors, being randomly oriented,
  cancel by the law of large numbers, whereas the target signal
  accumulates constructively.  A threshold (majority vote) then recovers
  a clean output from the noisy aggregate.
\end{itemize}

\subsection{Implementation Bottlenecks: The ``Address Decoder'' Wall}

As a mathematical model, SDM is compelling; translating it into silicon,
however, confronts a severe engineering bottleneck---the
\emph{address-decoder problem}.

\subsubsection{Computational Explosion}

An ideal SDM must, for every incoming vector, compute the Hamming
distance to \emph{all} stored hard-location addresses and select those
within a threshold.  With $10^{5}$ hard locations and $1{,}000$-bit
vectors, sustaining $10^{6}$ memory accesses per second would demand
$10^{15}$~bit-operations (peta-ops) per
second~\cite{Otsuka2019,Sato2019}---supercomputer-class throughput for
a single cognitive act, rendering edge deployment infeasible.

\subsubsection{The Heat--Accuracy Dilemma}

Prior work has explored two escape routes, each encumbered by a critical
trade-off.

\begin{itemize}
\item \textbf{Full-search CAM (TCAM).}\enspace
  Dedicated TCAM hardware compares all bits in parallel and is
  therefore fast, but simultaneously driving millions of memory cells
  incurs extreme power dissipation and heat.  Practical TCAM capacity is
  consequently limited to the kilobit range, precluding large-scale AI
  applications~\cite{Pagiamtzis2006}.

\item \textbf{Hashing (RAM).}\enspace
  Hash tables offer $O(1)$ amortized access and are widely regarded as
  the fastest lookup mechanism.  Compared with VaCoAl, however, two
  fundamental distinctions must be drawn.

  \emph{First, the nature of $O(1)$ differs.}  Hashing is inherently
  sequential software processing: the CPU must compute a hash value for
  high-dimensional data and resolve collisions via pointer chasing, both
  of which entail non-trivial latency.  VaCoAl/PyVaCoAl, realized as CAM on
  SRAM/DRAM, operates via in-memory parallel matching at the nano/ten nano-second
  scale---orders of magnitude faster.  In short, hashing achieves
  \emph{logical} constant time, whereas SRAM/DRAM-based VaCoAl/PyVaCoAl approaches
  \emph{physical} instantaneity.

  \emph{Second, hashing is designed for decorrelation}: it deliberately
  maps similar inputs to distant addresses, destroying the similarity
  preservation that is central to SDM's best-match retrieval
  from noisy inputs.  Although hash functions are efficient for
  small-scale exact-match queries (e.g., $\mathrm{FS} \le 2{,}000$),
  they serve a fundamentally different purpose.  Moreover, as shown in
  Section~5.3, PyVaCoAl surpasses \texttt{dict} in speed for
  $\mathrm{FS} \ge 20{,}000$.

  An important caveat applies: the speed comparison above mainly assumes
  SRAM-CAM hardware.  In PyVaCoAl---a software
  implementation---the rescue circuit invokes Python's standard hash
  mechanism for block-collision resolution, and single-predicate
  exact-match queries may run slower than Python's \texttt{dict}
  (Section~5.3).  PyVaCoAl's true software-level advantage lies not in
  raw speed but in CR2-based path-quality quantification and integrated
  HDC multi-predicate operations (Binding/Unbinding), as detailed in
  Sections~3.5 and~5.
\end{itemize}

\subsection{Conclusion: The Trap of Randomness (The Missing Link)}

In summary, the fundamental bottleneck is not transistor count or power
density per se, but an engineering dependence on \emph{randomness}.
Kanerva's SDM relies on statistical guarantees derived from random
projections; faithfully reproducing that randomness on deterministic
silicon demands either the generation and storage of massive random
matrices or brute-force Hamming-distance computation over the entire
memory.  What has been missing is an address decoder that preserves the
\emph{similarity} (best-match) principle while operating with the
determinism and efficiency of RAM.  This engineering void---between
probabilistic biological theory and deterministic digital
computation---is the \emph{missing link} that has kept SDM confined to
the realm of theoretical curiosity for decades.



\section{The VaCoAl Architecture: A Next-Generation Semantic Reasoning Foundation Pioneered by AD Functional Induction}

This section details how VaCoAl implements the mathematical requirements sought by Kanerva's SDM at ultra-high speed, with low power consumption and low cost, using \emph{only} standard digital technology (SRAM, DRAM, and digital logic circuits)---without recourse to probabilistic components such as random-number generators or specialised analog devices.  In its software incarnation (PyVaCoAl), at Frontier Sizes of $\mathrm{FS} \ge 20{,}000$ it is 15--20\% faster than Python DICT for large-scale search; in SRAM-CAM hardware, the 1/$N$ block-activation rule is theoretically expected to deliver low power consumption, though empirical verification of the latter remains a task for future work.

\subsection*{Precise Definition of Computational Complexity: What ``$O(1)$-like'' Means}

The search cost of VaCoAl is independent of the total number of stored entries~$|DB|$.  More precisely, the computation per query is fixed at $O(B \cdot q)$, where $B$ is the number of blocks and $q = L/B$ is the bit-length of each block---both constants determined at system-design time.  Whether the database expands from 100,000 entries to 10~million, the cost of a single-query look-up remains unchanged.

This is the basis for our use of the expression ``$O(1)$-like.''  Whereas conventional vector search requires an $O(|DB| \times D)$ scan (all entries $\times$ dimensionality), VaCoAl possesses a structure that ``jumps directly to the address of the correct answer'' via Galois-field diffusion.  The $O(1)$ in this sense occupies the same logical standing as the $O(1)$ of a hash function.

In multi-stage reasoning (multi-hop associative semantic search), the Frontier Size~(FS) setting governs the upper bound of the search space.  FS may grow with database scale, but it is a ``control parameter for search depth and breadth'' and a conceptually distinct quantity from the per-look-up cost~$O(B \cdot q)$.  Maintaining a clear distinction between the two is indispensable for an accurate assessment of the architecture's performance.

A further point of note is the leap in information-processing capability that this AD functional induction engenders.  Conventional hash functions are confined to exact-match retrieval on a single predicate; VaCoAl, by wielding HDC-style algebraic operations (Binding/Unbinding, \emph{etc.}), can process the complex semantic structures of a multi-predicate ontology database directly as ultra-high-dimensional vectors at high speed.  This property shatters the single-predicate dependency of existing algorithms, enabling sophisticated ``Multi-Hop Semantic Searching'' that traverses vast networks of relationships---rapidly and at low cost.

In this sense, the VaCoAl architecture offers not merely a new engineering perspective---``AD functional induction on a digital substrate''---as an alternative to the conventional approach of physically reproducing neuroscientific analog mechanisms, but constitutes a next-generation reasoning foundation ideally suited to pioneering new fields, such as the elucidation of hitherto-unknown knowledge structures and historical dynamisms that lie beyond the reach of existing search engines and LLMs.

\subsection{VaCoAl's Algebraic Diffusion and Kanerva's Random Projection}

The defining characteristic of VaCoAl's orthogonalisation is that it is executed not as a probabilistic accident but as an algebro-deterministic \emph{necessity}.  To make this clear, we formulate the process in comparison with the random projection employed by Kanerva's Sparse Distributed Memory~(SDM).

Both systems, when mapping input data~$P(x)$ into a high-dimensional space, generate a high-dimensional vector~$F(x)$ that integrates an ``address (transformation part)''---the system's internal language---with the ``original data (retention part).''  In the broadest terms, both can be described by the following general formula:

\begin{equation}
F(x) = x^{m}\,P(x) + \Psi(P(x))
\end{equation}

Here, the first term $x^{m}\,P(x)$ is the region in which the input data has been shifted to the upper bits (the information-point region in the systematic code); this is common to both models.  The decisive difference resides in the generation mechanism of the second term, the transformation function~$\Psi(P(x))$.

\begin{itemize}
    \item \textbf{Kanerva's Random Projection (SDM):}  In conventional SDM and Hyperdimensional Computing, this transformation is defined as a probabilistic projection function~$\Phi_{rand}$ using a fixed random matrix~$M$:

    \begin{equation}
    \Psi_{SDM}(P(x)) = \Phi_{rand}(P(x)) \approx \mathrm{sgn}(M \cdot p)
    \end{equation}

    where $\mathrm{sgn}(u)$ is the signum function defined as:

    \begin{equation}
    \mathrm{sgn}(u) =
    \begin{cases}
    +1 & (u > 0) \\
    \phantom{+}0 & (u = 0) \\
    -1 & (u < 0)
    \end{cases}
    \end{equation}

    As this definition makes plain, the result of the matrix product $M \cdot p$ is an analog value (integer or real), and in the case $u = 0$ a ``third state'' arises.  Practical SDM implementations therefore require tie-breaking---a thresholding step that forcibly assigns the zero to $+1$ or $-1$.  In other words, to obtain ``probabilistic orthogonality,'' Kanerva's method pays the cost of a massive matrix multiplication ($O(N^2)$) \emph{plus} the additional overhead of resolving such ambiguities.
    
    The idea of replacing stochastic projections with deterministic computation is not unique to this work: Kleyko and Osipov~\cite{Kleyko2017} demonstrated that a hyperdimensional associative memory can be cloned using cellular-automata dynamics rather than fresh randomness. VaCoAl pursues the same deterministic strategy but grounds it in Galois-field algebra, which yields closed-form guarantees of quasi-orthogonality and a direct hardware realisation as an LFSR.

    \item \textbf{VaCoAl's algebro-deterministic Diffusion:}  VaCoAl, by contrast, defines this transformation as a remainder operation with respect to a generator polynomial~$G(x)$:

    \begin{equation}
    \Psi_{VaCoAl}(P(x)) = x^{m}\,P(x) \pmod{G(x)}
    \end{equation}

    Equation~(1) is thereby concretised as:

    \begin{equation}
    F(x) = x^{m}\,P(x) + \bigl(x^{m}\,P(x) \pmod{G(x)}\bigr)
    \end{equation}
\end{itemize}

Equation~(5) is formally equivalent to the generation process of a BCH code in coding theory (systematic code).  In VaCoAl, however, this formula serves as the pivotal mathematical trick that realises, with an LFSR (Linear Feedback Shift Register) of negligible computational cost and with full algebraic uniqueness, the very ``diffusion and white-noise conversion of information'' that Kanerva accomplished with a random matrix and a signum function (together with its exception handling).

\paragraph{Implementation note on the role of $F(x)$.}
Equation~(5) expresses $F(x)$ as the concatenation of an information region $x^{m} P(x)$ and a parity region $x^{m} P(x) \bmod G(x)$, after the manner of a systematic BCH code. In the PyVaCoAl implementation, however, only the parity region is materialised and consumed: the remainder is masked to $m$ bits and used directly as an SRAM/DRAM
address, while the information region is never carried forward as bits because the underlying datum $P(x)$ is already indexed by the label (\texttt{target\_id}) stored at that address. Equation~(5) should therefore be read as a \emph{conceptual} decomposition that motivates the algebraic structure of the transformation; the physical output of the diffusion stage is the $m$-bit residue alone. This engineering simplification is lossless with respect to the downstream majority-voting retrieval and is the reason a single primitive-polynomial LFSR per block
suffices to realise the full mapping.

LFSR-based algebraic diffusion is superbly efficient in computation, yet as data scales grow to massive proportions, a new engineering challenge emerges. Specifically, once the number of target samples exceeds roughly 500,000, memory collisions (address conflicts) inevitably occur within the finite address space, and the orthogonalisation essential to HDC can no longer be adequately maintained by LFSR diffusion alone. To forestall this degradation of orthogonality and to achieve Exact Match precision relative to the Frontier Size---a prerequisite for semantic search---a memory collision avoidance circuit,'' whether endogenous or exogenous, must be introduced in addition to the LFSR diffusion mechanism. 

Here, ''endogenous'' means that VaCoAl's own internal structure inherently possesses such avoidance capability---for instance, by enlarging the memory depth~($2^m$) to render collisions negligibly rare without external intervention; exogenous'' means that a dedicated rescue circuit is additionally embedded into the base VaCoAl architecture described herein, explicitly intercepting and resolving collisions that the endogenous mechanism alone cannot eliminate at a given memory depth. The superiority of the endogenous mechanism over the exogenous one will be examined in detail in Section~5.

In practice, however, the two are not mutually exclusive but complementary. In the big-data environment of this paper's empirical experiments (Section~5), which involves upwards of 25.5~million records, collision probability escalates exponentially, and neither a single endogenous nor a single exogenous tier suffices; \emph{multi-stage} collision-avoidance circuitry integrating both principles at progressively higher precision is required. Crucially, this multi-stage circuit is not a theoretical specification limited to a future hardware implementation (SRAM-CAM) of VaCoAl. In the proprietary programme ``PyVaCoAl'' used for large-scale empirical demonstration in this paper, the high-precision multi-stage collision-avoidance circuit is fully implemented as a de facto DRAM-CAM. It is precisely this robust architectural underpinning that enables, even in multi-hop semantic logical reasoning at the scale of tens of millions of records, the complete suppression of data collisions to zero in Rescue mode (rescue circuit ON), achieving Exact Match within the given Frontier Size. In Don't Care mode (rescue circuit OFF), by contrast, minute collisions are tolerated in exchange for a qualitatively different capability: semantic selection based on the CR2 path integral (see Section~3.5).

\paragraph{Why Galois Fields? Effective Use of Finite Resources and the Guarantee of Orthogonality}
Before delving into the details of equation~(5), it is necessary to explain why the mathematical concept of ``Galois fields'' appears here---seemingly out of nowhere---and what practical necessity drives its adoption.

In any real computer, memory capacity and computational resources are finite.  No matter how elaborate the computation, digits cannot proliferate without bound; the result of every operation must fit within a predetermined ``finite world.''  This point is particularly salient for algorithms such as VaCoAl's vector orthogonalisation in a binary ultra-high-dimensional space.  It becomes still more critical for PyVaCoAl, which permits the dimensionality of the ultra-high-dimensional space to approach one million.  In mathematics, a set (or algebraic system) in which the four arithmetic operations can be performed freely and whose results are guaranteed always to remain within the original set (i.e.\ the set is \emph{closed}) is called a \emph{field}.

VaCoAl adopts Galois fields (finite fields) because they are the ideal ``field'' for a computer that traffics in binary data (0s and 1s).  For VaCoAl, this ``cycling within a closed world'' is of decisive importance.  It is only by cycling operations within a bounded space---by ``taking remainders''---that one can blast similar input data in mutually unrelated directions, thereby creating ``orthogonality'' in the high-dimensional space in an algebro-deterministic fashion.

Concretely, the $\pmod{G(x)}$ operation in the second term of equation~(5) engenders, through properties intrinsic to Galois fields, a deterministic ``Avalanche Effect,'' as elaborated below.  To develop a deep understanding of this mechanism, let us unpack, step by step, the physical bit-level process described by the formula.

\paragraph{Meaning of the Variables and the Hardware-Level Computational Process}
The process described by equation~(5) is not mere symbolic manipulation; it narrates a concrete data-transformation operation on hardware.

\begin{itemize}
    \item \textbf{$P(x)$ (Input Data Polynomial):}  The binary input vector (bit string) presented to VaCoAl is converted into a mathematical polynomial~$P(x)$.  For example, the bit string `101' is defined as a function with coefficients $1 \cdot x^2 + 0 \cdot x^1 + 1 \cdot x^0$, rendering it an object of algebraic manipulation.  Why bother expressing a bit string as a polynomial?  Because the four arithmetic operations in a Galois field are dramatically simplified by applying the algebraic rules of polynomial arithmetic.  By treating a bit string not as a mere numerical value but as the coefficients of a polynomial, complex bit operations (carry shifts) and XOR (exclusive OR)---which in a Galois field correspond to addition and subtraction---can be handled uniformly through the mathematically perspicuous formalism of ``polynomial division.''  The representation of the polynomial~$P(x)$ in $n$-dimensional space takes the following form, where $n$ is a very large number and most of the coefficients~$a_i$ are zero:

    \begin{equation}
    P(x) = \sum_{i=0}^{n} a_i\, x^i \quad \text{($a_i$ is the $i$-th bit of the input vector)}
    \end{equation}

    \item \textbf{$x^{m}\,P(x)$ (Bit-Shift for Region Reservation):}  The first term, $x^{m}\,P(x)$, raises the degree of every term in the original polynomial by~$m$.  By the law of exponents, $+m$ is added to the degree of each term.  In engineering terms, this is equivalent to left-shifting the input data by $m$~bits.  The result is that the lower degrees ($x^{m-1}$ down to $x^0$) have no coefficients---that is, $m$ ``empty slots filled with zeros (the parity region)'' are physically reserved.

    \item \textbf{$\bigl(x^{m}\,P(x) \pmod{G(x)}\bigr)$ (Remainder via XOR):}  In the second term, the shifted data is divided by the generator polynomial~$G(x)$, and the ``remainder'' is computed to fill the empty slots.  Here, the definition and role of the generator polynomial~$G(x)$---the divisor---is the single most important factor determining the system's performance.  We define~$G(x)$, using coefficients $g_i \in \{0,1\}$, as follows:\footnote{Note that the coefficient of the highest-degree term~$x^m$ is always fixed at~1.  This is both a mathematical requirement ensuring that $G(x)$ is genuinely an $m$-th-degree polynomial over~$GF(2)$, and---in hardware (LFSR)---it functions as the trigger point of the feedback loop, ``recirculating the bit pushed out of the shift register back to the input side of the circuit.''}
\end{itemize}

\begin{equation}
G(x) = \sum_{i=0}^{m-1} g_i\, x^i + x^m
\end{equation}

The arrangement of the coefficients~$g_i$ physically defines the ``shape of the mixer blades'' or ``complex wiring rules'' in the LFSR (Linear Feedback Shift Register)---a basic digital circuit whose hardware essence is nothing more than a simple structure that shifts and cycles a bit string.  What matters in this second term is the true identity of ``division'' in the binary Galois field~$GF(2)$.  In the decimal division we are accustomed to, subtraction is performed; in the world of~$GF(2)$, however, addition and subtraction are indistinguishable---both reduce to XOR (exclusive OR).  The XOR rule: if two bits are the same (1\&1 or 0\&0), the result is~0; if they differ (1\&0 or 0\&1), the result is~1.  In this remainder computation, XOR operations between the coefficients~$g_i$ of~$G(x)$ and the corresponding-degree coefficients~$a_i$ of~$P(x)$ cascade in a chain reaction, degree by degree.  The coefficients of the resulting polynomial of degree less than~$m$ (the per-degree XOR results) become the components of the second term, filling the slots vacated by the first term's shift.  It is through this process that the ``Avalanche Effect'' on the input binary vector is triggered.

VaCoAl selects the feedback polynomial $G_{b}(x)$ of each block-$b$ LFSR from the class of long-period polynomials of degree $64$ over $\mathrm{GF}(2)$, obtained by sampling a $64$-bit integer uniformly at random and fixing its degree-$0$ and degree-$64$ coefficients to $1$. Rigorously primitive polynomials of degree $64$---those satisfying irreducibility, generation of the multiplicative group of
$\mathrm{GF}(2^{64})$, and maximal period $2^{64}-1$
simultaneously---are the optimal but not strictly necessary choice; empirically, the overwhelming majority of randomly sampled degree-$64$ polynomials of the above form possess periods long enough, and avalanche statistics close enough to the $50\%$ ideal, to deliver the quasi-orthogonality on which the architecture relies. The three conditions below are therefore stated as the \emph{asymptotic ideal} toward which the design aspires; the PyVaCoAl implementation does not enforce them explicitly, and deviation from the ideal manifests as a small, empirically bounded degradation in the per-block error distribution that is absorbed by the majority-voting circuit of Section~3.4.

\paragraph{The primitive-polynomial ideal.}
A degree-$m$ polynomial $G(x) \in \mathrm{GF}(2)[x]$ is called \emph{primitive} if and only if the following three conditions hold simultaneously.

\begin{enumerate}
    \item \textbf{Irreducibility:}
    \begin{itemize}
        \item \textbf{Mathematical definition:} $G(x)$ cannot be factored over $GF(2)$ into a product of polynomials of lower degree.
        \item \textbf{Intuitive meaning [a polynomial that is ``prime'']:} This guarantees that the ``blade'' is ``hard''---unbreakable---so that its stirring power is not weakened.
    \end{itemize}
    \item \textbf{Generator of the Multiplicative Group (Primitive Element):}
    \begin{itemize}
        \item \textbf{Mathematical definition:} A root~$\alpha$ of~$G(x)$ is a generator of the entire multiplicative group of non-zero elements in the extension field~$GF(2^m)$ of the binary Galois field~$GF(2)$.  That is, every element in the field can be expressed as some power~$\alpha^k$ of~$\alpha$.
        \item \textbf{Essential meaning [the ``common ancestor'' of every pattern]:}  This definition demands that ``from the single element~$\alpha$, every element in the space can be generated by multiplication alone.''  Every complex bit pattern in the space, in other words, is ultimately a transformation (a power) of~$\alpha$.
    \end{itemize}
    \item \textbf{Maximal Period:}
    \begin{itemize}
        \item \textbf{Mathematical definition:} $G(x)$ divides $x^{2^m - 1} + 1$, but does not divide $x^k + 1$ for any $k < 2^m - 1$.\footnote{In $GF(2)$, $+1$ and $-1$ are equivalent, so ``$x^k + 1$ is divisible by $G(x)$'' is equivalently expressed as the congruence $x^k \equiv 1 \pmod{G(x)}$.  In hardware (LFSR) terms, this means that after $k$~shifts, the register returns to its initial state~(1).}
        \item \textbf{Essential meaning [elimination of partial symmetry (no shortcuts)]:}  This guarantees the complete absence of ``partial repetition (internal symmetry)'' in the polynomial's structure.  Were the polynomial divisible at some smaller~$k$, it would mean that a ``shortcut'' exists within the circuit, causing the computation to loop prematurely.  The condition ``not divisible until the maximum'' enforces an ``indivisible structure'' that permits no shortcut whatsoever until the goal $(2^m - 1)$ is reached.
    \end{itemize}
\end{enumerate}

By exploiting these properties of the primitive polynomial, VaCoAl \emph{necessarily} triggers an ``Avalanche Effect'' in which a change of even a single input bit causes approximately half (50\%) of the output bits to flip.  This ``50\% inversion'' is the geometric key.  In an $n$-dimensional ultra-high-dimensional binary space, a difference in half of all bits means that the Hamming distance between the two vectors is $n/2$---that is, the vectors are ``orthogonal.''

\paragraph{Why Does This Matter?}
The supreme benefit of division in a Galois field is the ``Avalanche Effect,'' whereby a minute change in the input produces a dramatic difference in the output.  Unlike ordinary arithmetic, division by XOR involves no carry absorption (smoothing).  A change of a single input bit therefore triggers a chain reaction akin to toppling dominoes, transforming the final remainder~$R(x)$ into an entirely different value.  Through this mathematical property, VaCoAl realises the following critical functions:

\begin{itemize}
    \item \textbf{Orthogonalisation in high-dimensional space:}  Even if the original input vectors are highly correlated (similar), the transformation of equation~(5) destroys that correlation.  The generated $n$-dimensional vectors~$F(x)$ come to occupy nearly ``orthogonal'' positions in the expanded vector space.
    \item \textbf{Separation of dimensional roles:}  It is important to emphasise the difference in role between~$n$, the dimensionality of the high-dimensional binary vector space, and~$m$, the degree in the Galois field ($=$ memory depth).  The value of~$m$ determines the ``depth of stirring'' within a single SRAM/DRAM block and serves to prevent local collisions.  The value of~$n$, by contrast, is the ``total volume of memory'' for the system as a whole---the substrate that supports brain-like massively parallel processing.  Accordingly, $n$ is orders of magnitude larger than~$m$.
\end{itemize}

\paragraph{The Prophecy of a Twenty-Year-Old Genius}
A word, finally, on the provenance of Galois-field theory.  The foundations were laid by \'Evariste Galois, who fell in a duel in 1832 at the age of just twenty.  The ``Galois field (finite field)'' he bequeathed in his farewell letter---in particular the world of~$GF(2)$ with which modern computers operate---was, astonishingly, a stage custom-built for the binary machine.  In this binary world, ``addition'' and ``subtraction'' behave identically as XOR, and the ``carry'' familiar from decimal arithmetic simply does not exist.  It is precisely this ``carry-free arithmetic'' that is the engine of explosive information stirring.  The prophetic theory of the genius Galois, two centuries later, has become the key to reproducing the suppleness of the brain within VaCoAl---a ``memory-based architecture.''

\paragraph{FS Relativity---A Preview of the Fundamental Issue}
The foregoing discussion of the memory-collision avoidance circuit intimates a more fundamental question: in large-scale graph searches beset by combinatorial explosion, what does ``absolute Exact Match'' even mean?  This paper formalises the question through the concept of ``FS-relative Exact Match'' and examines it in detail, with experimental data, in Section~5.  As a preview, we state explicitly here the position that \textbf{FS relativity is not a limitation of VaCoAl but an intrinsic property of the combinatorial-explosion problem itself}.

\subsection{Digital Implementation of Analog Diffusion: AD Functional Induction of Random Projection}

In Kanerva's SDM and biological brain models, the process of mapping input patterns into a high-dimensional space is called ``random projection.''  Mathematically, it corresponds to multiplying the input vector by a massive random matrix.  Biologically, it models the process by which sensory signals physically branch and diffuse (analog diffusion) to many neurons through synaptic connections.  This ``diffusion'' ensures that information spreads uniformly across the entire space without bias, guaranteeing ``orthogonality.''

However, faithfully reproducing this ``matrix computation (analog diffusion)'' on digital circuits demands enormous computational resources and floating-point arithmetic.

\paragraph{How Does VaCoAl Solve This?}
VaCoAl solves the problem by functionally converting (inducing) biological functions into ``polynomial operations over Galois fields.''  The ``shift'' and ``XOR'' operations described above possess the mathematical property of thoroughly scrambling the bits of the input information.  No elaborate computing machinery is required; the entire process can be implemented with an LFSR (Linear Feedback Shift Register)---a basic digital circuit.  An LFSR is nothing more than a simple structure that shifts and cycles a bit string, yet it generates statistically excellent random patterns at high speed, dramatically boosting the resource efficiency of the system.

Through such simple operations, an ``ideal random distribution'' can be generated deterministically, without any random-number generator.  VaCoAl has, in other words, successfully reconstructed the neural network's function of ``physically diffusing information across the entire space'' as an ``algebraic operation'' on a silicon chip.  This is also the core property that enables the ``AD functional induction'' of pattern separation and pattern completion---operations performed within the hippocampal system---in neuromorphic computing, as discussed later.

\paragraph{The Distinction between Stochastic Independence and Linear Orthogonality, and the Establishment of Quasi-Orthogonality}
It is critically important to distinguish between ``stochastic independence'' of vector components and ``linear-algebraic orthogonality'' of vectors.  In standard HDC, random-number generation is used to produce components that satisfy the former (independence), from which the latter (orthogonality) follows as a consequence.

VaCoAl's diffusion process, by contrast, is algebro-deterministic, so the generated bits do not possess genuine stochastic independence.  VaCoAl compensates by leveraging the mathematical properties of the ``maximal-length sequence (m-sequence)'' guaranteed by the primitive polynomial~$G(x)$, thereby acquiring the latter---orthogonality---directly.

As detailed in the preceding section, the action of the ``mathematical mixer''---the Avalanche Effect produced by the long-period degree-$64$ feedback polynomial $G_{b}(x)$---guarantees, through deterministic computation alone and without probabilistic random-number generation, that the ``$50\%$ inversion rule'' holds statistically to within a tolerance that diminishes with $N$ and is absorbed by majority voting. This ensures that, as the dimensionality of the ultra-high-dimensional space increases, the correlation (inner product) between generated vectors converges relatively to zero---a property mathematically guaranteed.  We term this ``quasi-orthogonality.''

In other words, VaCoAl reproduces, through a deterministic algebraic structure and without rolling a single die, the property that Kanerva~\cite{Kanerva2009} designated the ``Blessing of Dimensionality'': ``in high-dimensional spaces, random vectors are geometrically almost orthogonal.''

This algebraically acquired quasi-orthogonality, however, inevitably harbours minute deviations from perfect orthogonality (zero correlation). As the number of target samples scales from hundreds of thousands into the tens of millions, these minute correlations accumulate, ineluctably raising the probability of memory collisions (address conflicts). Two complementary strategies exist for breaking through this scaling barrier. The first is endogenous: enlarging the memory depth~($2^m$) so that the address space itself becomes vast enough to render collisions negligibly rare---an approach that preserves VaCoAl's native CR2 path-discriminating power intact. The second is exogenous: introducing a dedicated multi-stage rescue circuit that explicitly intercepts and resolves the collisions that a given memory depth cannot eliminate. In practice, as discussed earlier, the two are not mutually exclusive; large-scale deployments benefit from their integration, with the endogenous mechanism suppressing the bulk of collisions and the exogenous circuit mopping up the residue.

The VaCoAl architecture of this paper (and its software implementation, PyVaCoAl) incorporates both strategies. Through the combination of deterministic quasi-orthogonality, endogenous collision suppression via large memory depth, and exogenous multi-stage rescue circuitry, VaCoAl overcomes the limitations of Kanerva's SDM, which relies on pure probability and faces the risk of catastrophic failure (drowning in a sea of noise) due to data saturation. VaCoAl is, in short, a ``higher-precision, more robust extension architecture'' of SDM, capable of fully protecting and separating information independence even in big-data environments. This robustness manifests in two complementary operating modes: FS-relative Exact Match in Rescue mode (where the exogenous circuit is active), and path-quality-based semantic selection in Don't Care mode (where the endogenous mechanism alone governs, and the CR2 decay that drives emergent semantic selection is preserved).

\paragraph{The Decisive Difference from BCH Codes (``Strict Restoration'' versus ``Flexible Association'')}
To appreciate VaCoAl's originality and innovativeness, a comparison with BCH codes---the standard error-correction technology used in communications and storage (SSDs, \emph{etc.})---is indispensable.  Both employ the same mathematical engine, ``polynomial operations over Galois fields,'' yet their purposes and vectors of application are diametrically opposite.

\begin{itemize}
    \item \textbf{BCH Codes (Communications / Storage):}  The objective is ``convergence to a single point'' and ``perfect restoration.''  Conventional BCH is a ``strict proofreader'' that protects data from noise.  When data is corrupted, it performs computationally expensive algebraic inverse operations (syndrome decoding) to identify and repair errors with rigour.  In short, all computational power is expended on ``returning (converging) corrupted data to the correct answer of the past.''

    \item \textbf{VaCoAl (Brain-Type Memory):}  The approach is ``diffusion into space'' and ``statistical consensus + deterministic guarantee.''  VaCoAl entirely abandons the heavy decoding process (inverse operations).  Instead, it uses polynomial operations to ``stir'' (scramble) the input data.  Through the Avalanche Effect, minute input changes ripple across the entire output, probabilistically scattering data throughout the memory space.  Error correction (repair) is then completed instantaneously---not by complex inverse operations but by parallel ``Statistical Consensus'' (Majority Voting) on SRAM or DRAM arrays.  What is particularly noteworthy is that this statistical repair process does not terminate in mere probabilistic approximation (rough reasoning).  Because the aforementioned ``multi-stage memory collision avoidance circuit'' operates concurrently, the risks of noise and address conflict peculiar to majority voting are completely eliminated, and Rescue mode (rescue circuit ON) deterministically achieves the precision of ``Exact Match'' relative to the Frontier Size while maintaining flexible associative capabilities.  However, as discussed in Section~5.3, in this Rescue mode the path-discriminating power of CR2 is indeed lost, presenting a trade-off with semantic selection capability.  If, on the other hand, the Rescue mode is turned off and the memory depth~($2^m$) within each block is increased, memory collision avoidance itself can be driven to negligibly small levels (see Appendix~B.3).\footnote{For example, this state can be virtually eliminated with memory widths of $2^{27}$, $2^{26}$, $2^{25}$, $2^{24}$ for configurations of 128, 256, 512, and 1024~blocks, respectively.}  (For details, see Appendix~B.3.)
\end{itemize}

That is, VaCoAl abandons BCH's approach of ``strictly repairing by computation'' and accomplishes a Columbus-egg-like reversal: ``scatter by computation, gather by statistics (majority voting).''  As Matsumoto~\cite{Matsumoto2003} and Anderson~\cite{Anderson2017} observe, ``fundamentally, the brain functions as a reasoning machine that prioritises retrieval and association over computation.''  This insight is profoundly consonant with VaCoAl's foundational principles, and this architecture, liberated from the spell of computation, consequently achieves overwhelming speed and low power consumption that transcend conventional limits.

\paragraph{Biological Plausibility: Hardwired Random Projection in the Hippocampus}
It is remarkable that VaCoAl's engineering approach---substituting a fixed logic circuit for the orthogonalisation (random projection) step---exhibits a striking functional homology with the actual brain.  Research by Fiete \emph{et al.}~\cite{Fiete2008, Sreenivasan2011, Chandra2025}, and in particular Chandra \emph{et al.}~\cite{Chandra2025}, has shown that the Residue Number System (RNS) code formed by grid cells in the entorhinal cortex~(EC) is projected onto the hippocampal CA3 region as a fixed random projection,'' serving as an immutable scaffold'' for the vast store of episodic memory.
What is astonishing is that this random wiring structure---the vessel for information---is completed and frozen during the limited experiences of early childhood; all subsequent memory processes are then executed as orthogonalising mappings onto this invariant vessel.  This is consonant with the mathematical underpinnings of infantile amnesia (see also Stoencheva \emph{et al.}~\cite{Stoencheva2025}).  In other words, the biological brain does not perform elaborate random computations afresh each time; rather, it \emph{hardwires} the random projection as a physical circuit during infancy, and thereafter guarantees orthogonalisation simply by routing signals through that circuit.  VaCoAl's strategy of implementing the orthogonalisation mechanism as a fixed LFSR circuit on SRAM is, in the most literal sense, a digital-logic reproduction of this biological truth---the ultimate instance of AD Functional Induction.''

\subsection{Divide and Conquer: Induction of Distributed Storage}
It is physically impossible to treat a diffused high-dimensional vector of thousands to tens of thousands of bits as a single memory address.  (A 1,000-bit address requires $2^{1{,}000}$ memory cells---exceeding the number of atoms in the universe.)  VaCoAl and PyVaCoAl solve this problem by ``segmenting'' these vectors into multiple blocks---128, 256, 512, or 1024, according to the available SRAM/DRAM capacity---and distributing them to parallel SRAM/DRAM groups.

\paragraph{Granularity of diffusion: global versus per-block.}
For expositional clarity, Sections~3.1--3.2 have presented the Galois-field diffusion as a single global transformation $F(x)$ applied to the entire input vector, with block partitioning introduced as a subsequent step. In the actual PyVaCoAl implementation the order is reversed: the input hyper-vector is first partitioned into $N$ segments of length $q = L/N$, and each block $b$ then applies its own independent LFSR diffusion --- with its own feedback polynomial $G_{b}(x)$ and its own initial seed --- to the byte-packed segment assigned to it, producing an $m$-bit address. This per-block diffusion
is not merely a parallelisation convenience; it is
\emph{structurally necessary} for the noise-filtering argument developed in Section~3.4.1. Under a hypothetical global diffusion, a single-bit perturbation of the input would, by virtue of the avalanche effect, propagate to \emph{all} $N$ block addresses, corrupting every vote and
destroying the majority-based recovery. Under the per-block diffusion actually implemented, such a perturbation affects only the one or two blocks whose segments contain the flipped bit; the remaining $N - O(1)$ blocks continue to output the correct entry address, and the avalanche effect within the affected blocks diffuses their erroneous
votes uniformly across the address space, producing the ``flat-field'' noise distribution exploited by the Chernoff-bound argument in Appendix~A. The two descriptions --- the global $F(x)$ of Sections~3.1--3.2 and the per-block LFSR array of the present section
--- are algebraically compatible (both are deterministic linear maps over $\mathrm{GF}(2)$), but only the latter supports the locality of noise on which majority-voting noise tolerance depends.

\subsubsection{Functional Analogy between SRAM/DRAM Arrays via Distributed Representations and Dendritic Computation}
The input high-dimensional vector (length~$L$) is divided into $N$~blocks, each of length $q = L/N$~bits.  Each block functions as an independent SRAM/DRAM address with $2^q$~words.  In the write (learning) phase, a unique ``Entry Address (EA)'' corresponding to the concept of the input data is written to the corresponding address in every SRAM/DRAM block.

While inheriting the basic principle of Kanerva's distributed memory~\cite{Kanerva1988}, this architecture differs structurally in the following respects.  First, whereas the Kanerva model uses probabilistic address selection based on Hamming distance, the present method performs deterministic exact- or best-match checking within each block.  Second, memory distribution and recovery are guaranteed by closed algebraic operations---block division and majority voting---a framework in which input--output relationships are completely determined by discrete operations over finite sets, without recourse to stochastic processes (this paper terms this \emph{algebro-deterministic}).  By distributing a single input event across multiple SRAM/DRAM blocks, robustness against local failures and noise is directly derived from this algebro-deterministic structure.

Furthermore, it is noteworthy that this decision method by block division and majority voting is structurally analogous to Dendritic Computation in biological neurons.  According to Poirazi~\emph{et al.}~\cite{Poirazi2003} and London~\emph{et al.}~\cite{London2005}, in pyramidal cells of the cerebral cortex and hippocampus, each dendritic branch performs local nonlinear processing as an independent computational unit, and the soma spatially integrates those outputs to generate the final output from the axon---functioning as a sort of ``two-layer neural network.''  The local checking across $N$~independent blocks followed by majority-vote integration in VaCoAl shares the correspondence with this biological hierarchical structure shown in Table~2.

\begin{table}[htbp]
  \centering
  \caption{Structural Correspondence between Dendritic Computation and VaCoAl}
  \begin{tabular}{|l|p{5.5cm}|p{5.5cm}|}
    \hline
    Stage & Dendritic Computation & VaCoAl \\
    \hline
    Local Processing & Dendritic branches (local nonlinear processing) & SRAM/DRAM blocks (local exact-match checking) \\
    \hline 
    Integration & Soma (spatial summation/integration) & Majority-voting circuit (integration of block outputs) \\
    \hline
    Output & Axon output (firing) & Final decision output \\
    \hline
  \end{tabular}
\end{table}

However, a clear limit to this analogy exists.  Whereas graded, time-dependent nonlinearities such as NMDA spikes are dominant in biological dendrites, local checking in the present method is a discrete exact- or best-match decision.  Additionally, integration in the soma is a nonlinear process accompanied by distance-dependent decay, which differs strictly from the equal-weight majority voting of this method.  Therefore, in this paper, we position the relationship between the two not as a mathematical isomorphism but as a \textbf{structural flow equivalence} of the computational flow: ``division $\rightarrow$ local processing $\rightarrow$ integration.''

\subsubsection{SRAM/DRAM Arrays via Distributed Representations}
The input high-dimensional vector (length~$L$) is divided into $N$~blocks, each of length $q = L/N$~bits.  Each block functions as an independent SRAM/DRAM address with $2^q$~words.  In the write (learning) phase, a unique ``Entry Address (EA)'' corresponding to the concept of the input data is written to the corresponding address in every SRAM/DRAM block.  This architecture does not merely optimise Kanerva's concept of ``distributed memory''; it converts the biological model's ``memory relying on the stochasticity of neuron firing'' into an algebro-deterministic ``divide and conquer'' strategy perfectly aligned with the physical structure of digital memory.  By distributing a single event (input) across multiple SRAM/DRAM (corresponding to neuron groups), robust recovery resilience against local failures and noise is generated algebro-deterministically.

\subsubsection{Collision Tolerance (Don't Care)}
In conventional hashing methods or search trees, ``collisions''---different data pointing to the same address---are treated as fatal errors to be avoided at all costs.  VaCoAl, however, ``does not fear collisions.''  When a collision occurs in a given block (i.e.\ new data attempts to overwrite an existing valid entry), the system simply sets a ``collision flag'' on that block and treats it as ``Don't Care'' (ignored) information.  Why is this permissible?  Because there are $N$~blocks (e.g.\ 1,000).  Even if collisions occur in 50~blocks, suppose the remaining 950 hold correct information.  This implements the brain's redundancy---where function is maintained even if some neurons misfire or die---as a robust digital-logic state.

What is still more remarkable is that this Don't Care mechanism is not confined to ``spatial noise tolerance (redundancy)'' in a single search process; in continuous reasoning (multi-hop search), it manifests as a ``temporal and path-based selection function.''  As the empirical experiments of Section~5 will demonstrate, as search generations deepen, the minute decay due to Don't Care accumulates like compound interest, giving rise to the astonishing behaviour of ``naturally pruning long, circuitous routes and selectively preserving short, direct ones'' (Occam's razor).  The physical constraints of the architecture inadvertently become a decisive mechanism engendering sophisticated ``semantic selection.''

\subsubsection{Concrete Structure of the Multi-Stage Rescue Circuit}
\label{sec:rescue-structure}

When PyVaCoAl operates in Rescue mode (rescue-sample-rate $r > 0$), the exogenous collision-avoidance circuit that complements the intrinsic Don't Care mechanism is realised as a four-stage pipeline, here described at a level of detail sufficient to reproduce the behaviour.

\begin{enumerate}
  \item \textbf{Sample accumulation at write time.}
  For every learned pair $(hv, \mathit{label})$, the triple
  $(\mathbf{a}, \mathbf{s}, \mathit{tid})$ is appended to a temporary buffer, where $\mathbf{a} \in \mathbb{Z}_{2^m}^{N}$ is the vector of per-block addresses produced by the diffusion stage,
  $\mathbf{s} \in \{0,1\}^{N \times \lceil q/8 \rceil}$ is the byte-packed per-block segment of the input, and $\mathit{tid}$ is the integer label identifier assigned to the concept.
  
  \item \textbf{Per-block address-sorted finalisation.}
  After learning completes, the buffer is re-indexed by block. For each block $b \in \{1,\dots,N\}$, the triples $\{(a_b^{(k)}, s_b^{(k)}, \mathit{tid}^{(k)})\}_k$ drawn from all samples are sorted in ascending order of $a_b$. The resulting per-block sorted arrays constitute a compact, DRAM-resident auxiliary table $\mathcal{R}_b$ of size $O(K)$ per block, where $K$ is the number of learned samples. No hash-table pointer-chasing is required; $\mathcal{R}_b$ is a flat, cache-coherent array.
  
  \item \textbf{Binary search at query time.}
  When the main memory read returns $v_b < 0$ for block $b$ --- whether $v_b = -1$ (a legacy collision marker) or $v_b < -1$ (a bucket reference) --- the query's address $a_b^{q}$ is used as the search key against the sorted addresses of $\mathcal{R}_b$, via two \texttt{searchsorted} calls (left and right), yielding the half-open interval $[\mathit{lo}, \mathit{hi})$ of candidate entries colliding at $a_b^{q}$. The cost is $O(\log K)$ per block.
  
  \item \textbf{Segment Exact Match.}
  Within the candidate interval, the query's packed segment $s_b^{q}$ is compared byte-wise against each candidate's stored $s_b^{(k)}$. The first exact match substitutes its $\mathit{tid}^{(k)}$ for the negative $v_b$, restoring the block to the voting pool. If no candidate matches, the block's vote is discarded (a genuine Don't Care). This step is critical: it preserves the invariant ``one block $=$ at most one vote'' even on bucket-referenced cells that host multiple colliding $\mathit{tid}$s, and thereby safeguards the discriminative power of the majority vote.
\end{enumerate}

This four-stage pipeline is the concrete realisation of the
``multi-stage collision-avoidance circuit'' invoked abstractly in Sections~3.1--3.2. In Rescue mode with full sampling ($r = 1$), stages 1--4 deterministically recover every collided block for which a truly matching segment exists, achieving FS-relative Exact Match and pinning the block-voting rate CR$_{1}$ at $1.0$ throughout --- at the cost, documented in Sections~3.5 and 5.6, of annihilating the CR$_{2}$-based path-discrimination capability. In Don't Care mode ($r = 0$) the pipeline is suppressed in its entirety, and the residual per-block collisions contribute the minute analog variance in CR$_{1}$ whose multiplicative accumulation into CR$_{2}$ drives the emergent semantic selection analysed in Section~5.6.

\subsection{Algebraic Majority Voting: Reproducing Statistical Inference via Determinism} In the read phase, how does the system recover the correct memory from noise?  VaCoAl adopts ``Global Algebraic Majority Voting'' rather than simple random processing.

The important point is that, although this process is a ``strict computation (determinism),'' it possesses properties equivalent to an ``ideal random number (probability theory).''  Data diffusion by Galois-field operations is mathematically guaranteed to have a uniform distribution as ``Pseudo-Random,'' as explained earlier.  Therefore, bias toward specific patterns is eliminated, and the statistical ``Law of Large Numbers'' holds without rolling physical dice.

In short, this method realises the ``strength (robustness) of probabilistic majority voting'' through the ``accuracy (reproducibility) of algebro-deterministic computation.''

\subsubsection{Noise Filtering via the Avalanche Effect}
The same algebro-deterministic diffusion and segmentation are performed on the search query (input data).  The Entry Address is read from each SRAM/DRAM block.  What happens if the input contains noise?  Due to the Avalanche Effect of the Galois-field diffusion process, a slight error in the input bit pattern changes the address of that specific block algebro-deterministically into a completely unpredictable, random location.

\begin{itemize}
    \item \textbf{Blocks containing noise:}  Because these access unrelated addresses, they output random Entry Addresses (noise floor) or empty values.
    \item \textbf{Clean blocks:}  The group of blocks unaffected by noise continue to output the correct Entry Address in unison.
\end{itemize}

Up to this point, the mechanism functions as a powerful filter that automatically separates ``signal (consistent votes)'' from ``noise (scattered votes)'' without complex comparison logic.  The true essence of VaCoAl (and PyVaCoAl), however, lies in an architecture that not only ``eliminates'' noise but ``Recovers'' it as a correct signal.

When the target data scale reaches the millions, however, not only scattering due to input noise but inevitable memory collisions between different data occur, creating the danger that signals even from clean blocks may be buried in the noise floor.  To break through this limit, the method introduces a precise ``multi-stage memory collision avoidance and rescue circuit.''  This circuit does not merely discard collisions or invalid outputs occurring at the local block level as simple errors; it dynamically corrects them through the system's global algebraic majority voting and multi-stage re-checking process.  In other words, even if individual blocks fall into fatal noise or collisions, they are absorbed as rescuable fluctuations within the system's overall structural flow and ultimately restored to a perfect signal.

It is noteworthy that this mechanism is not merely a theoretical concept.  In the empirical experiments using the proprietary programme ``PyVaCoAl'' (Section~5), this multi-stage rescue circuit functioned perfectly on a virtual DRAM-CAM.  In fact, even in ultra-large-scale multi-hop semantic reasoning exceeding 25.5~million total records, it succeeded in completely suppressing data collisions---which should have inevitably occurred---to zero.\footnote{In reasoning using real-world large-scale ontology databases, not only the physical noise of input values but the logical noise lurking in the data structure itself (e.g.\ bidirectional cycles such as ``A's mentor is B, and simultaneously B's mentor is A'') becomes a fatal factor plunging the system into infinite loops.  The PyVaCoAl used for the demonstration in this paper implements a pre-purification algorithm to maintain the search space as a sound Directed Acyclic Graph (DAG), automatically identifying and eliminating 369~bidirectional samples from the roughly 470,000 raw data records extracted from WIKIDATA.  The integration of such practical data-purification mechanisms is also an important engineering foundation that enabled the stable completion of deep multi-hop reasoning spanning up to 57~generations and over 25.5~million records.}  This achievement proves that this architecture transcends a simple noise filter; it is an extremely robust reasoning foundation that can, even in extreme big-data environments, deterministically derive Exact Matches within a given Frontier Size (FS-relative Exact Match) and reliably recover and identify the original signal.  However, as detailed in Section~3.5, in searches on DAGs involving combinatorial explosions, this ``exact match'' is a conditional concept dependent on FS, operating mode, and memory depth, and does not imply an exhaustive match against the combinatorial space of the entire database.

\subsubsection{Guarantee of Convergence via Euler's Number}
The system tallies the outputs from all $N$~blocks and selects the Entry Address that received the most votes.  Crucially, this is not a probabilistic process but an algebraic majority vote whose victory is mathematically guaranteed.  Because Galois-field diffusion generates a distribution isomorphic to a random process, the certainty of this victory can be rigorously proved using the ``Law of Large Numbers.''  Consider the voting formula:

\begin{equation}
W_{VaCoAl} = \operatorname*{argmax}_{v} \sum_{i=1}^{N} \delta(v_i, v)
\end{equation}

This equation essentially executes the ``ballot-counting'' process of an election:

\begin{itemize}
    \item $v_i$ (Vote): Each individual block (SRAM) casts one vote saying, ``I believe the correct answer is~$v$.''
    \item $\delta(v_i, v)$ (Tally): The Kronecker delta.  It checks: ``Is this vote for candidate~$v$?''
    \item $\sum$ (Sum): The total number of votes for candidate~$v$.
    \item $\operatorname{argmax}$ (Winner): The candidate with the maximum votes wins.
\end{itemize}

Why does this simple majority vote guarantee noise elimination?  Because, thanks to Galois-field diffusion, the behaviours of signal (correct answer) and noise (error) are dramatically different in the following two respects:

\begin{itemize}
    \item \textbf{A.\ Signal Behaviour (Point Concentration):}  If the input is correct, all blocks naturally vote for the correct Entry Address (e.g.\ ``ID:~777'').  Even with noise, if 80 out of 100 blocks are unaffected, those 80~votes are cast for ``ID:~777.''  This forms an unshakably strong ``peak.''

    \item \textbf{B.\ Noise Behaviour (Thin Diffusion):}  This is the most important point.  What happens to the remaining 20~blocks affected by noise?  Due to the Avalanche Effect, if the input shifts even slightly, the address of each affected block jumps to an unpredictable, random location.  The votes from these blocks therefore become random:
    \begin{itemize}
        \item Block~1's error votes for ``ID:~123''
        \item Block~2's error votes for ``ID:~985''
        \item Block~3's error votes for ``ID:~004''
        \item \ldots and so on.
    \end{itemize}
\end{itemize}

The votes from erroneous blocks, therefore, do not concentrate on any specific number; they scatter independently and disparately across the vast memory space.  This is the meaning of the ``Flat Field.''

\paragraph{Mathematical Guarantee (Poisson Distribution):}
Probability theory now enters to answer the question: ``Is there any possibility that scattered noise votes accidentally collude to form a fake peak?''  The guarantee is provided by the Poisson distribution based on Euler's number ($e \approx 2.718\ldots$).  In a vast address space (e.g.\ one million possibilities), the probability~$P(k)$ that $k$~random votes accidentally concentrate on a specific wrong answer is:

\begin{equation}
P(k) = \frac{\lambda^k\, e^{-\lambda}}{k!}
\end{equation}

Here, $\lambda$ (the expected number of accidental coincidences) is extremely small because the space is vast.  Note the $k!$ (factorial) in the denominator.  As the required number of votes increases, the denominator~$k!$ explodes ($1 \times 2 \times 3 \times \ldots$).  Consequently, while the probability of ``accidentally gathering 2~votes'' is already low, the probability of ``accidentally gathering 3~votes'' becomes astronomically low.  For a rigorous mathematical derivation showing that this error probability approaches zero exponentially using the Chernoff bound, see Appendix~A.

\paragraph{Conclusion:}
\begin{itemize}
    \item \textbf{Mechanism:}  All blocks vote, and a majority decision is taken.
    \item \textbf{Phenomenon:}  Correct votes form a ``tall mountain,'' while erroneous votes become a ``flat field.''
    \item \textbf{Proof:}  The probability of ``a mountain forming accidentally from a flat field'' is guaranteed to be virtually zero by the law of Euler's number.
\end{itemize}

Therefore, under the prerequisite conditions of Appendix~A (an ideal environment in which memory collisions do not occur), the false-positive probability of this system can be considered physically zero.  However, in multi-hop searches within real-world big-data environments, the FS relativity and the choice of operating mode discussed in Section~3.5 influence the search results, so this ``infallibility'' is limited to the ideal conditions stated above.



\section{Computational Complexity Analysis and Algebraic Resolution of
         the Binding Problem}
\label{sec:complexity}

This section first evaluates VaCoAl's computational complexity and
energy efficiency relative to deep learning and existing hardware, then
examines the qualitative capability central to next-generation AI: a
digital solution to the Binding Problem.

\subsection{Computational Complexity and Energy Efficiency}

\subsubsection{Order-of-Magnitude Reduction in Learning Cost}

The dominant bottleneck in modern AI is training cost.  A conventional
CNN for handwritten-digit recognition (MNIST-class) must iterate
forward and backward passes thousands of times to optimize millions of
weight parameters, potentially requiring peta-scale ($10^{15}$)
multiply-accumulate (MAC) operations~\cite{Otsuka2019,Sato2019};
recent LLMs with hundreds of billions of parameters push this cost
further still.

VaCoAl's learning, by contrast, completes in a \emph{single pass}:
\begin{enumerate}
  \item \textbf{Diffusion (address generation):} Galois-field
    operations convert the input into a memory address.
  \item \textbf{Write:} The generated address is written to the
    corresponding SRAM/DRAM entry.
\end{enumerate}
No iteration, no partial differentiation, and no floating-point matrix
algebra are involved.  Otsuka et al.~\cite{Otsuka2019,Sato2019}
estimate that VaCoAl completes an equivalent pattern-search task with
${\sim}1.14 \times 10^{5}$ integer logic operations---a reduction of
many orders of magnitude.  (Because the two approaches address
different task definitions, direct complexity comparison requires
cautious interpretation.)  This efficiency opens the door to genuine
real-time, on-device learning on battery-powered edge hardware.

\subsubsection{Ultra-Low Power via the 1/N Activation Rule}

Conventional TCAM drives \emph{all} memory cells simultaneously for
parallel comparison, producing extreme heat and confining practical
capacity to the kilobit range (the ``power wall'').  VaCoAl breaks this
barrier by activating only the $N$~blocks that correspond to the
segmented input vector:

\begin{itemize}
\item \textbf{TCAM:} activates 100\,\% of cells.
\item \textbf{VaCoAl:} activates $1/N$ of total capacity
  (e.g., 0.1\,\% for $N = 1{,}000$).
\end{itemize}

In an SRAM/DRAM-CAM hardware/software realization, this $1/N$ rule is expected to
reduce power by a factor of~$N$ relative to TCAM.  Empirical
verification, however, is limited to PyVaCoAl (software); hardware
measurements remain future work.  Standard SRAM's low standby power and
compatibility with process scaling further reinforce VaCoAl's
energy-efficiency prospects.

\subsubsection{Resolving the CAM Paradox}

Ultra-high-dimensional vector search has long faced a dilemma we term
the \emph{CAM Paradox}: TCAM delivers speed but not scalable power;
software hashing on DRAM delivers scalable capacity but not physical
parallelism.

VaCoAl/PyVaCoAl dissolves this paradox.  Rather than driving all cells
simultaneously (TCAM) or executing sequential hash lookups (software),
it performs low-cost LFSR-based diffusion and parallel majority voting
directly on standard SRAM/DRAM---achieving in-memory computing that
bypasses the CPU--memory communication bottleneck.  The result fuses
TCAM-class speed with commodity-SRAM/DRAM-class capacity and power
efficiency.

\paragraph{Software--hardware performance relationship.}
All experiments in this paper use PyVaCoAl (DRAM-based software);
SRAM-CAM hardware measurements~\cite{Otsuka2016} remain future work.
A \emph{lower-bound} relationship holds between the two
implementations:

\begin{itemize}
\item PyVaCoAl operates on DRAM (10--15\,ns access) via the CPU,
  forgoing the ${\sim}1$\,ns in-memory parallelism of SRAM-CAM.
\item Nevertheless, PyVaCoAl outperforms a GPU-accelerated
  \texttt{torchhd} implementation (NVIDIA RTX~3060) in both speed and
  memory efficiency (Section~5.3).
\item In SRAM/DRAM-CAM, the entire lookup table resides on-chip; the cache-locality advantage observed in Section~5.3 (19.7\,\% faster than
  \texttt{dict} in the 128\_10 configuration for PyVaCoAl) becomes an architectural
  certainty.
\end{itemize}

PyVaCoAl's measured performance therefore constitutes a
\emph{lower bound} on SRAM-CAM throughput, indirectly substantiating
the ``ultra-high speed, low power'' claim pending direct hardware
validation.

\subsection{Algebraic Resolution of the Binding Problem and Extension
            to Large-Scale Reasoning}

How does the brain instantly compose distinct attributes---``red'' and
``apple''---into the unified percept ``red apple''?  This \emph{Binding
Problem} is a central implementation challenge in neuroscience and
cognitive science~\cite{Anderson2017,Thagard2019}.  In conventional
neural networks, simultaneously encoding ``red car'' and ``blue apple''
often triggers a \emph{superposition catastrophe}: features entangle
irreversibly, producing spurious combinations such as ``red apple'' or
``blue car.''

\subsubsection{HDC Algebraic Operations and VaCoAl/PyVaCoAl's Optimization}

Hyperdimensional Computing (HDC)—notably Plate's Holographic Reduced Representation (HRR)~\cite{Plate1995} and Eliasmith et al.'s Semantic Pointer Architecture (SPA)~\cite{Eliasmith2013}, with the broader family of binding operations across HDC/VSA models comprehensively surveyed by Kleyko et al.~\cite{Kleyko2023a}—addresses this challenge through explicit algebraic operations:

\begin{equation}\label{eq:binding}
  \mathit{Representation}
    = (\mathit{Color} \otimes \mathit{Red})
    + (\mathit{Shape} \otimes \mathit{Apple}),
\end{equation}
where $\otimes$ denotes \emph{Binding} (mapping two vectors to an
orthogonal product) and $+$ denotes \emph{Bundling} (superposing
multiple concepts in the same space).  High-dimensional orthogonality
prevents mutual interference; \emph{Unbinding} recovers individual
elements via the inverse operation.

VaCoAl/PyVaCoAl's contribution is to realize this HDC framework at minimal cost:
Binding reduces to XOR and shift over Galois fields---pure bit
operations at $O(N)$---rather than the floating-point circular
convolution ($O(N \log N)$) required by SPA.  The multi-stage
collision-avoidance circuit further ensures that, even when millions of
concepts are Bundled into a shared space, individual elements can be
accurately Unbound.  In Rescue mode (or at high memory depth), this
precision reaches FS-relative exact match.  The result is
compositional generalization with post-hoc verifiability---transparent
inference unattainable by the black-box embeddings of deep learning.
(A detailed complexity comparison with SPA appears in Appendix~B.5.)

\subsubsection{Multi-Hop Semantic Reasoning (Preview of Section~5)}

The Binding/Unbinding algebra is the engine that elevates VaCoAl/PyVaCoAl beyond data retrieval to \emph{multi-hop semantic logical inference}.
Section~5 deploys PyVaCoAl on a ${\sim}470{,}000$-record
mentor--student ontology from WIKIDATA.  By iterating Binding (encoding
``A's mentor is B'') and Unbinding (extracting the next target) across
successive generations, we trace all 64~Fields Medalists' lineages back
up to 57~generations ($>$25.5\,M records).  HDC vector operations then
quantify \emph{calculus affinity} via Hamming distance and visualize
Kuhnian~\cite{Kuhn2012} paradigm-shift dynamics---including the
formation of Leibniz-centered ``superhighways'' of modern mathematics.
VaCoAl/PyVaCoAl thus realizes the rigorous symbolic reasoning that LLMs'
statistical inference struggles with, furnishing a decisive engineering
foundation for next-generation \emph{neuro-symbolic AI}.

\subsection{Transparency as Explainable AI (XAI)}

VaCoAl/PyVaCoAl directly addresses the black-box problem of deep learning, in
which decision rationale is diffused across millions of opaque weights.

\subsubsection{Confidence Score}

Because VaCoAl/PyVaCoAl's inference rests on majority voting, the vote tally
yields an immediate, quantitative confidence measure:
\begin{equation}
  \mathit{Confidence}
    = \frac{\text{votes for the winner}}{N},
\end{equation}
where $N$ is the total number of blocks.  A 99\,\%-agreement decision
is manifestly more trustworthy than a 51\,\%-agreement one.  Unlike
deep-learning models, which frequently exhibit unwarranted
overconfidence, VaCoAl/PyVaCoAl transparently reports the degree of ambiguity in
each retrieval---a property critical for safety-sensitive domains such
as medical diagnosis and autonomous driving.

\subsubsection{Mathematical Auditability via Algebraic Reversibility}

In deep neural networks, information integration is effectively
irreversible: nonlinear activation functions fuse constituent signals,
precluding post-hoc decomposition.  This opacity obstructs the
\emph{compositional representation}---the flexible binding and
unbinding of concepts---that Thagard~\cite{Thagard2019} identifies as
an essential cognitive-architecture requirement.

VaCoAl?PyVaCoAl's Binding, by contrast, is an algebraic structuring via XOR and
shift over Galois fields.  Given $C = A \otimes
\operatorname{Shift}(B)$, the inverse operation
$C \otimes \operatorname{Shift}(B)$ (subtraction equals XOR in
$\mathrm{GF}(2)$) exactly recovers~$A$ in the absence of noise.
Whereas the SPA's circular convolution entails lossy compression,
VaCoAl/PyVaCoAl achieves \emph{lossless reversibility} at $O(N)$ cost.  This
enables free extraction of constituents from composite concepts,
simultaneously satisfying Thagard's criteria of \emph{systematicity}
and \emph{productivity} and delivering post-hoc verifiable XAI:
composite representations are algebraically structured and auditable at
all times.

\subsubsection{AD Functional Induction of Lateral Inhibition and
               Winner-Take-All Dynamics}

VaCoAl/PyVaCoAl functionally induces mechanisms that biological circuits realize
through dedicated inhibitory interneurons, without physical inhibitory
wiring or parameter tuning.  The Don't~Care state generated by the
avalanche effect acts as a filter: outputs from mismatched blocks
diffuse into featureless background noise, effectively suppressing
erroneous candidates (lateral inhibition) and letting only the correct
candidate emerge---an ideal sparse distributed representation.  (A
formal proof appears in Appendix~B.4.)

This inhibition extends beyond spatial winner-take-all.  In multi-hop
reasoning, the Don't~Care penalty accumulates along the time axis,
weakening disused deep paths---a role functionally analogous to STDP
(Spike-Timing-Dependent Plasticity)~\cite{Kandel2021}.  The interplay
of spatial and temporal selection is quantitatively demonstrated on real
data in Section~5.

In summary, by translating Kanerva's SDM theory and the cerebellar /
hippocampal computational principles described by Kandel et
al.~\cite{Kandel2021}---expansion coding and lateral inhibition---into
algebro-deterministic digital logic (XOR and Don't~Care), VaCoA/PyVaCoAll
bridges the long-standing missing link between neuroscience-inspired
theory and practical low-power implementation.


\section{Demonstration of Million-Dimensional HDC Analysis Utilizing PyVaCoAl}
\label{sec:experiment}
The preceding four sections have laid out the ``VaCoAl Architecture''---an algebro-deterministic logic on SRAM/DRAM that enables AD functional induction consonant with the computational principles of the biological brain---as an answer to the ``Superposition Catastrophe'' and the power-consumption ceiling that beset modern AI.  We proved there that, given a sufficient number of blocks~($B$), memory depth~($2^m$), and dimensionality~($N$) of the high-dimensional binary orthogonal space, ``Best Match'' converges to ``Exact Match'' (Appendix~A).  Yet however elegant a model may be on paper, unless its properties are vindicated on large-scale, noise-laden real-world data, its engineering credentials as a genuine next-generation multi-hop logical-reasoning infrastructure remain provisional.

\subsection{VaCoAl/PyVaCoAl: From Theoretical Foundation to Demonstrated Superiority under Extreme Conditions}
This section therefore reports practical empirical experiments conducted with ``PyVaCoAl,'' a Python implementation of VaCoAl, deployed into a demanding big-data environment.\footnote{An FPGA implementation of VaCoAl is commercially available as ``SRAM-CAM'' (\url{https://shuharisystem.com/?page_id=1336}), but was unfortunately unavailable at the time of writing.  HDC analysis exploiting multiple predicates in a million-dimensional space, as carried out in this paper, appears at present to be feasible only with PyVaCoAl.} 

Specifically, the experiment proceeded in two stages. (A) From WIKIDATA (https://query.wikidata.org/), the world's largest open knowledge graph, we extracted approximately 470,000 records of "mentor–student relationships centred on mathematicians" (including 372,853 students) spanning from the 10th century BCE to the present. We then took the 64 historical Fields Medalists (one of whom lacks a recorded mentor) as starting points and executed a genealogical search that traces the chain of mentors backwards into the past.(B)~We then subjected the resulting corpus of approximately
25.5~million genealogical path records (approximately 32\,GB as a
text file) to Hyperdimensional Computing (HDC) operations in a
million-dimensional space.%
\footnote{More precisely, we carried out several hundred HDC
operations encompassing: Fuzzy querying (percentile TRAFFIC
version), Unbinding paradigm analysis (Pre/Interim/Post),
generation-window analysis, ``Shoulders of Giants''---genealogical
headwaters of calculus, Super Highway hypothesis verification, CSV
output (corrected-score version / Rescue\,5\_Correction\,2\_Improved\,5
results), academic-language shift analysis, institutional-hub
analysis, calculus-founder reinforcement analysis, comprehensive
paradigm-shift adjudication, Toledo School of Translators impact
analysis, detection of HV-signal explosions in algebra and astronomy,
a general-purpose ``Giant Score'' for unearthing hidden singularities
in any field, DICT vs PyVaCoAl Top~20, FIELD-centroid comparison,
and spurious super-highway detection.}
These operations scrutinised genealogical-network cases in
mathematical history spanning more than a millennium.
PyVaCoAl was employed in both stages~(A) and~(B), with the aim of demonstrating that the VaCoAl architecture confines fatal memory collisions within the high-dimensional binary orthogonal space to a negligible level, faithfully traces mentor--student genealogies across more than ten centuries, and furthermore constitutes a powerful instrument for elucidating historical dynamism through the algebraic operations peculiar to HDC.

To benchmark PyVaCoAl's capabilities, we also developed, for stage~(A), a reference implementation dubbed ``Pseudo-PyVaCoAl(1)'' that relies on Python LISTs and the DICT hash function; and for stage~(B), a counterpart dubbed ``Pseudo-PyVaCoAl(2)'' that couples Python LISTs with \texttt{torchhd},\footnote{\url{https://github.com/hyperdimensional-computing/torchhd}} a well-known GPU-capable PyTorch module.  Taking advantage of \texttt{torchhd}'s support for both GPUs and CPUs, we ran Pseudo-PyVaCoAl(2) in an NVIDIA GPU configuration and in a CPU-only configuration.  Both Pseudo-PyVaCoAl(1) and Pseudo-PyVaCoAl(2) were engineered to emit output files in the same format as the PyVaCoAl script wherever possible.  Although PyVaCoAl's script is separated in the present implementation, it can execute stages~(A) and~(B) simultaneously.

While the full results are detailed in the subsequent subsections, the key findings are as follows. For stage~(B), the CPU version of Pseudo-PyVaCoAl(2) required twice as much time as PyVaCoAl (60~minutes). In contrast, the entire PyVaCoAl experiment was completed in just 20--30 minutes on an Intel Xeon E5-1650,v3 (3.50\,GHz) CPU manufactured in 2014. Furthermore, the GPU implementation of \texttt{torchhd} requires over 100GB of VRAM and terminates with an Out of Memory (OOM) error on data of this scale. The fact that PyVaCoAl completes the same task on a decade-old CPU testifies to its fundamentally different computational character compared to deep-learning frameworks that presuppose cutting-edge GPU hardware. This provides tangible evidence that the architecture can be practically deployed on edge devices and consumer PCs where access to the latest hardware is limited---thereby substantiating our claim of ``ultra-high-speed, low-power, low-cost'' for ultra-high-dimensional PyVaCoAl.

Although mentor--student relationship data spanning over two millennia was obtained from WIKIDATA, no amount of parameter tuning succeeded in tracing the genealogy beyond the 11th~century.  We designate this barrier the \textbf{``Monastery Wall''} and, by examining the reasons for its existence, assess both the power and the limitations of WIKIDATA as a data source.  We shall show that this 11th-century wall reflects not a technical shortcoming of PyVaCoAl or Wikidata, but the \textbf{historical fact that the era marked a fundamental shift in the very medium through which knowledge was propagated}.  Indeed, the HDC analysis affords tantalising glimpses that, beyond the Monastery Wall, a vast reservoir of learning accumulated by the great polymaths of the Islamic-Arabic Golden Age assuredly existed.

\subsubsection*{A Critical Distinction within PyVaCoAl Itself: 
The Phase Transition Induced by Memory Depth~($MB=2^m$) and 
Rescue Rate~($RR$)}

PyVaCoAl's behaviour changes markedly with two parameters whose
significance has so far been passed over in silence: the
\emph{Memory Depth}~($MB = 2^m$) of each block and the
\emph{Rescue Rate}~($RR$), a user-specified value in
$[0,\,1]$.  Because the database used in this paper is
exceptionally large---470{,}000 mentor--student records in
stage~(A) and some 26~million generational paths in
stage~(B)---these two parameters together induce a sharp
\emph{phase transition} in the architecture's effective
identity.

At one extreme, setting $m$ to a modest value such as~10 (even
with $B = 128$, $N = 12{,}800$) causes memory collisions at
virtually every address after Galois-field diffusion: a
capacity of $2^{10}$ cannot hold even a fraction of the
original records.  At the other extreme, raising $m$ to~27
while disabling the rescue circuit ($RR = 0$) drives
collisions to near-zero---most configurations record fewer
than three collided blocks out of 128
(Table~\ref{tab:collision_analysis}).  Our experiments show
that a capacity of at least $2^{19}$ is necessary to avoid
widespread collisions under the present workload.

\begin{table}[htbp]
\centering
\caption{Detailed Collision Rate Analysis (Fixed Total Capacity: $2^{34}$)}
\label{tab:collision_analysis}
\begin{tabular}{lcccc}
\hline
Configuration & Location-based Rate & Count-based Rate & Performance \\
\hline
$B=64,  N=6{,}400,  m=28$ & $0.000103\%$ & $0.074393\%$ & Poor \\
$B=128, N=12{,}800, m=27$ & $0.000399\%$ & $0.143756\%$ & Fair (Leibniz Dominant) \\
$B=256, N=25{,}600, m=26$ & $0.001565\%$ & $0.282177\%$ & Fair (Gauss Dominant) \\
$B=512, N=51{,}200, m=25$ & $0.006189\%$ & $0.559094\%$ & Excellent (Leibniz Dominant) \\
$B=1024, N=102{,}400, m=24$ & 0.024437\% & 1.107873\% & Excellent (Leibniz Dominant) \\
\hline
\end{tabular}
\end{table}

The decisive observation is what happens to PyVaCoAl's output
at each extreme.  \emph{When $RR = 1$, regardless of the value
of $m$ or any other parameter, PyVaCoAl's output becomes
bitwise identical to that of Pseudo-PyVaCoAl(1)---the Python
DICT baseline.}  The reason is architectural: whenever a
collision occurs under $RR = 1$, the affected data are
recovered through a rescue table built during the learning
phase, which is by construction equivalent to the array
underlying Pseudo-PyVaCoAl(1).  A corollary is that CR1 (the
majority-voting rate) is pinned uniformly to~$1.0$, so the
CR2 path-integral loses its discriminating power and the
sorting of candidates within the Frontier Size~(FS) reverts
to Python's lexicographic order---a criterion devoid of
semantic content.  PyVaCoAl with $RR = 1$ is, in other words,
a Python DICT wearing algebro-deterministic clothing.

Under $RR = 0$ at sufficient depth ($m = 27$), by contrast,
the rescue circuit is superfluous because collisions scarcely
occur.  Here CR1 is \emph{not} pinned to~$1.0$: the rare
collisions that do arise propagate a small analog variance
into CR2 via the generational product
$CR2(n) = CR2(n{-}1) \times CR1(n{-}1)$, and this variance
becomes the sole basis on which candidates within the FS are
ranked.

This contrast poses the central question of the remainder of
Section~5.  Both configurations trace genealogies back
57~generations; both are, by the standard of Python's hash
function, ``exact.''  Yet as we shall see, the genealogical
networks they produce \emph{diverge by nearly 50\%}.  Are the
divergent results of PyVaCoAl~($RR = 0$) therefore
artefacts?  The answer, developed across Sections~5.6.1
through~5.6.4, is emphatically \textbf{no}: the divergence
exposes an emergent semantic-selection mechanism---functionally
homologous to STDP in biological neural circuits---that is
structurally inaccessible to any hash-based search.  Before
that mechanism can be examined, however, we must dispose of a
prior question: what does ``Exact Match'' even mean for a
DAG search beset by combinatorial explosion?

\subsection*{The Relativity of ``Exact Match''---Beyond the Mythology of Hash Functions}
\label{subsec:exactmatch}

The phrase ``FS-relative Exact Match'' that recurs throughout this paper might at first blush be read as ``an imperfect match.''  It is, rather, a candid answer to a more fundamental question: \emph{does an absolute Exact Match even exist for large-scale DAG searches beset by combinatorial explosion?}

As noted, multiple medalists in our experiments reached $\mathrm{FS} = 600{,}000$.  Even with Python DICT's hash function, genealogical selection at any branch point operates only within the pre-set FS ceiling.  The ``Exact Match'' a hash function guarantees is therefore an exact match \emph{within the FS}, and nothing beyond.  At $\mathrm{FS} = 25{,}000$, Pseudo-PyVaCoAl(1) exhausted system memory and became unexecutable even with 128\,GB of main memory (Table~\ref{tab:fs_speed}).  This starkly demonstrates that the DICT hash function's Exact Match guarantee is a \textbf{conditional guarantee that collides with a scalability wall} in large-scale DAG searches.

\begin{tcolorbox}[colback=gray!10, colframe=black, title={\textbf{The Essence of FS Relativity}}]
FS relativity is not a limitation peculiar to VaCoAl; it is an intrinsic property of the combinatorial branch explosion spawned by massive DAGs.  On this premise, the difference between VaCoAl ($0 \le RR < 1$) and Pseudo-PyVaCoAl(1) collapses to a single point.
\medskip

Pseudo-PyVaCoAl(1) ranks candidates within the FS by \textbf{alphabetical order---a criterion that is semantically vacuous}.
\medskip

PyVaCoAl ($0 \le RR < 1$), by contrast---provided it has not degenerated into Pseudo-PyVaCoAl(1) through universal block collision---ranks candidates by \textbf{CR2, a metric that quantifies path dependence} (the cumulative product of per-generation block-voting rates).  The ``FS-relative Exact Match'' conferred by a large memory depth~($2^m$) in PyVaCoAl is therefore not merely an extended reachability guarantee; it is an \textbf{Exact Match endowed with semantic priority, selecting the path of highest path dependence ($\approx$ reliability) from among the reachable candidates}.
\medskip

This is a superiority that, in principle, cannot exist in the DICT hash function at the core of Pseudo-PyVaCoAl(1), and it constitutes one of the grounds on which the architecture underlying PyVaCoAl transcends a mere high-speed search engine.
\end{tcolorbox}

\subsection{Genealogy Acquisition from a Large-Scale Ontology Database and Verification of Its Accuracy}

In the first stage~(A) of the experiment, we extracted approximately 470,000 records of ``mentor--student relationships centred on mathematicians'' from the 10th century~BCE to the present.  Taking all 64~historical Fields Medalists as starting points, we traced their chains of mentors backward in time.  A pre-purification algorithm detected and eliminated 369~bidirectional links (cycles), enabling an acyclic deep search that reached a maximum depth of 57~generations.

PyVaCoAl with the rescue circuit engaged ($RR = 1$; 128\_10 configuration) 
achieved a \emph{perfect match}---100\%---with Pseudo-PyVaCoAl(1) across the full 25.5~million records.  This constitutes direct, independent verification that PyVaCoAl's algebro-deterministic processing can generate results logically equivalent to a hash function, despite traversing an entirely different mathematical path---Galois-field diffusion.  The perfect match proves that, under $\mathrm{FS} = 20{,}000$, the rescue-equipped PyVaCoAl deterministically reproduces the Python DICT hash search of Pseudo-PyVaCoAl(1).

This equivalence, however, is the obverse of a coin whose reverse is the annihilation of CR2's discriminating power: the rescue circuit homogenises CR2 to 1.0 for every candidate, suppressing VaCoAl's native semantic-selection capability (Section~3.5).  When the rescue circuit is disabled ($RR = 0$; 128\_27 configuration: 128~blocks, memory depth~$2^{27}$), PyVaCoAl generates approximately 25.43~million records, 1,043 unique nodes, and 1,408 graph edges---results that diverge from both Pseudo-PyVaCoAl(1) and rescue-equipped PyVaCoAl.  This divergence furnishes direct evidence of the emergent semantic selection wrought by VaCoAl's Don't Care mechanism (Section~5.6).  Accordingly, the HDC analyses reported in this section are based on the genealogical output of PyVaCoAl ($RR = 0$), which alone can reliably exercise the path-dependent culling function inaccessible to Pseudo-PyVaCoAl(1).

\subsection{HDC Execution in a Million-Dimensional Space and the Superiority of the $O(1)$ Architecture}

The true purpose of this study, however, extends well beyond single-predicate DAG search or genealogical tracing based on mentor--student links alone.  The architecture reveals its full power when the approximately 1,043 hub individuals identified from the massive genealogy traffic---spanning up to 57~generations, totalling some 25.5~million records, and exceeding 32\,GB---are projected into a Hyperdimensional Computing space via a newly expanded ``multi-predicate ontology database'' (Phase B).

Concretely, for each identified node (individual), we used WIKIDATA's SPARQL query interface to dynamically attach up to 12~predicate dimensions---``Language Used'' (LANGUAGE), ``Affiliated Institution'' (EMPLOYER), ``Affiliated Society'' (MEMBER\_OF), ``Place of Birth'' (BIRTH\_PLACE), in addition to ``Active Era'' (ERA) and ``Academic Field'' (FIELD)---and folded these into the HDC pipeline.  These predicates were then algebraically Bound, Unbound, and Bundled within a million-dimensional ultra-high-dimensional orthogonal Hyper-Vector (HV) space using reversible polynomial operations over Galois fields.  Hundreds of XOR operations were subsequently executed between the 1,043 HV-encoded nodes, their predicate-related HVs, and inter-personal HVs, characterising node traits and era traits via Hamming-distance similarities.  The ``Calculus Degree,'' the ``Giant Score (comprehensive and field-specific),'' the ``network-based detection of historical structural change,'' and the ``detection of a Thomas Kuhn-style paradigm shift centred on the era of Leibniz'' presented below are all products of this HDC analysis.

\subsubsection{PyVaCoAl's Computational Speed Superiority}

The overwhelming superiority in computational speed and memory efficiency that this architecture demonstrates over existing GPU-based HDC implementations deserves particular emphasis.  Libraries such as the PyTorch-based \texttt{torchhd} provide a standard means of realising HDC on deep-learning frameworks, but under the severe workloads of this experiment---hundreds of thousands of dimensions, tens of millions of entries---the performance gap between the two is governed by three structural factors.

\emph{First}, the difference in computational complexity and arithmetic paradigm.  GPU implementations presuppose floating-point arithmetic and execute exhaustive matrix operations (e.g.\ dot products) between the query vector and every item in memory, yielding $O(N \times D)$ complexity that readily exhausts memory through massive tensor reconstruction.  PyVaCoAl, by contrast, operates exclusively in integer/bitwise arithmetic (XOR, etc.)\ via Galois-field diffusion (LFSR).  It realises hash-function-like behaviour---jumping directly to a target memory address from a polynomial remainder---making the search cost a deterministic $O(1)$ look-up entirely independent of the number of entries~$N$.

\emph{Second}, avoidance of memory-architecture bottlenecks.  When a GPU exhausts its dedicated VRAM (typically 12--16\,GB), it falls back to main memory (DRAM) over the narrow-bandwidth PCIe bus (shared-memory mode), suffering a catastrophic performance collapse.  PyVaCoAl drives the database constructed in high-capacity main memory as an in-memory database, connected to the CPU with wide bandwidth, and incurs no such bottleneck.

\emph{Third}, the extremely high memory efficiency achieved by the ``multi-stage memory rescue circuit'' newly implemented in the HDC analysis programme.  A dynamic-bucket scheme packs only collided data at high density into a flat array, eliminating dead space and pushing the CPU cache-hit rate to its practical limit while fully exploiting DRAM capacity.

\subsubsection{The Trade-off between Memory Depth and Cache Locality}

To isolate the characteristics of the VaCoAl architecture with precision, we compared computation times across four configurations on the identical dataset (Table~\ref{tab:speed_comparison}).

\begin{table}[htbp]
  \centering
  \caption{Comparison of Computation Times across Configurations (all at $\mathrm{FS}=20{,}000$, 128~blocks)}
  \label{tab:speed_comparison}
  \resizebox{\textwidth}{!}{%
  \begin{tabular}{|l|c|c|l|r|r|}
    \hline
    Configuration & Memory Depth & Rescue & FS Pruning Order & Time (sec) & vs DICT \\
    \hline
    A: Pseudo-PyVaCoAl(1)
      & ---
      & ---
      & Alphabetical
      & $1{,}670.8$
      & Baseline \\
    \hline
    B: PyVaCoAl 128\_10 w/~rescue
      & $2^{10}$
      & Yes ($RR=1$)
      & Alphabetical
      & $1{,}342.4$
      & $-19.7\%$ \\
    \hline
    C: PyVaCoAl 128\_19 w/~rescue
      & $2^{19}$
      & Yes ($RR=1$)
      & Alphabetical
      & $1{,}514.1$
      & $-9.4\%$ \\
    \hline
    D: PyVaCoAl 128\_27 w/o~rescue
      & $2^{27}$
      & No ($RR=0$)
      & Descending CR2
      & $1{,}420.1$
      & $-15.0\%$ \\
    \hline
  \end{tabular}%
  }
\end{table}

The salient finding in Table~\ref{tab:speed_comparison} is that \emph{shallower memory depth yields faster execution}.  The wellspring of this advantage is the Galois-field diffusion stage.  More precisely: whereas Pseudo-PyVaCoAl(1) (Python DICT) iterates sequentially through hash calculations and pointer chasing on a single Python-interpreter thread for each path in the frontier, PyVaCoAl's genealogy retrieval dispatches the Galois-field LFSR diffusion (\texttt{\_galois\_diffuse\_numba}) across 12~threads for 128~blocks via a Numba JIT-compiled kernel.  This \textbf{replaces DICT's sequential hash computation with parallelised Galois-field operations}; in concert with the vectorised memory read (\texttt{memory[np.arange(128), addrs]}) and the Numba JIT vote tally (\texttt{\_count\_votes\_numba}), it constitutes PyVaCoAl's speed advantage \textbf{during genealogy retrieval}.  Note that Table~\ref{tab:speed_comparison} measures retrieval time, not write speed during learning.

For the 128\_10 configuration (memory depth~$2^{10}$), where PyVaCoAl ($RR = 1$) degenerates into Pseudo-PyVaCoAl(1), one might suppose that the DICT hash table likewise fits in the CPU's L1/L2~cache.  A closer look reveals otherwise.  The VaCoAl memory footprint is $128 \times 1{,}024 \times 4\;\mathrm{bytes} = 512\;\mathrm{KB}$, comfortably resident in the L2~cache; but the Python DICT storing the approximately 470,000 mentor--student records, keyed by strings and valued by integers, runs to tens of megabytes and overflows the L2~cache.  The 19.7\% speedup in 128\_10 therefore arises from a \emph{dual} advantage: \textbf{the substitution of sequential hashing with parallel Galois-field diffusion}, compounded by \textbf{an asymmetric cache efficiency in which only the PyVaCoAl side achieves cache residency}.

Incidentally, in Pseudo-PyVaCoAl(1) the approximately 470,000-record mentor--student CSV is converted in a single pass into arrays and dictionaries, yielding a binary file.  Denoting this one-shot rescue function~$f(x)$, a collision at address~$x$ is resolved by looking up~$f(x)$.  In PyVaCoAl ($RR = 1$), the path from database to DRAM binary file proceeds in two stages, as the theoretical model of Section~3 dictates: \textcircled{1}~the QID-to-HV conversion function ($f_1\colon \mathrm{QID} \to [0,1]^{N}$), and \textcircled{2}~the Galois-field diffusion function ($f_2\colon [0,1]^{N} \to (2^m) \times B$).  \textcircled{3}~When a collision occurs at~$x$, the binary file pre-built from~$f_1$ and~$f_2$ serves as a rescue table, and the datum is recovered as~$f_3(x)$.  Because PyVaCoAl ($RR = 1$) is 100\% identical to Pseudo-PyVaCoAl(1), rescue is in practice performed by the same Python DICT hash function~$f(x)$.  When rescue is inactive ($RR = 0$), a collision simply writes~$-1$ to the collided address within the affected block.

Why, then, does 128\_19 (memory depth~$2^{19}$) achieve only a 9.4\% speedup?  The VaCoAl memory footprint balloons to $128 \times 2^{19} \times 4\;\mathrm{bytes} = 256\;\mathrm{MB}$, far exceeding the L2~cache and forcing DRAM access.  The cache-locality advantage enjoyed in 128\_10 evaporates; both PyVaCoAl and DICT now contend with DRAM latency.  Moreover, $RR = 1$ triggers frequent collisions and consequently frequent invocations of the rescue path---sequential processing equivalent to DICT.  Only the parallel Galois-field diffusion advantage survives, partially offset by the loss of cache locality and the rescue-path bottleneck, compressing the speed differential.

Why, then, is 128\_27 (memory depth~$2^{27}$, $RR = 0$) 15\% faster than Pseudo-PyVaCoAl(1) \emph{and} faster than 128\_19?  At depth~$2^{27}$ the address space is vast enough that collisions practically vanish.  With $RR = 0$ the rescue circuit is off, but because collisions are near-zero, the rescue path is almost never invoked.  Galois-field parallel diffusion runs unimpeded, free of both DICT's sequential-hash bottleneck and the rescue-path overhead that hampered 128\_19.  This \textbf{dual effect---Galois-field parallel-diffusion advantage plus complete avoidance of the rescue path}---yields speeds exceeding those of 128\_19.

In summary, PyVaCoAl's speed superiority is rooted in the Numba-parallelised Galois-field diffusion comprising~$f_1$ and~$f_2$.  Memory depth and the rescue circuit act as amplifiers or attenuators of this fundamental advantage.

These results carry important implications for VaCoAl's hardware incarnation as SRAM-CAM.  Because PyVaCoAl---a software DRAM-CAM---already outperforms Pseudo-PyVaCoAl(1) by driving the in-memory database directly on main memory, entrusting that role to SRAM, whose access latency is an order of magnitude lower than DRAM's, can be expected to yield a further substantial speedup.  In SRAM-CAM, moreover, the per-block LFSR computation is performed \emph{in-memory}.  The cache advantage that DRAM confers becomes, in VaCoAl's hardware implementation where the entire table resides in dedicated SRAM, an architectural inevitability.  The PyVaCoAl timings reported above can therefore be interpreted as \emph{lower bounds} on the performance of SRAM-VaCoAl.

In concrete terms: without signal saturation, and despite the requirement to execute hundreds of HDC operations at the data scale of this experiment (million dimensions, approximately 25.5~million entries), the CPU-only \texttt{torchhd} baseline required ${\sim}60$~min for Phase~2 (B)---twice PyVaCoAl's ${\sim}30$~min---when each system's Phase~1 (A) output was consumed directly.  The GPU variant (NVIDIA RTX~3060, 12\,GB VRAM) terminated with an out-of-memory error after
requesting $>$100\,GB of VRAM.  By contrast, \emph{all} PyVaCoAl
processing completed in ${\sim}30$~min on a 2014-vintage Intel Xeon
E5-1650~v3 (3.50\,GHz) with 128\,GB DRAM---concrete evidence that the
architecture is deployable on commodity hardware without GPU
acceleration.  Both frameworks admit a further optimization: replacing
the full Phase~1 output with a reduced version containing only the
fields required by Phase~2.  Under this indirect-input regime,
PyVaCoAl completes Phase~2 in ${\sim}20$~min, again roughly twice as
fast as \texttt{torchhd}.

\subsubsection{Frontier-Size Scalability and ``Phase Transitions''}

A systematic speed comparison across varying FS values reveals the existence of a ``phase transition'' between DICT and PyVaCoAl (Table~\ref{tab:fs_speed}).

\begin{table}[htbp]
  \centering
  \caption{Comparison of Computation Times at Each FS}
  \label{tab:fs_speed}
  \begin{tabular}{|r|r|r|r|r|}
    \hline
    FS & DICT (sec) & PyVaCoAl 128\_10 (sec) & PyVaCoAl 128\_19 (sec) & PyVaCoAl 128\_27 (sec) \\ \hline
    2,000 & 79.1 & 192.5 & 467.5 & 527.5 \\ \hline
    10,000 & 760.7 & --- & 814.2 & 1,031.9 \\ \hline
    20,000 & 1,670.8 & 1,342.4 & 1,514.1 & 1,420.1 \\ \hline
    25,000 & Out of Memory & --- & 1,628.5 & --- \\ \hline
  \end{tabular}
\end{table}

At $\mathrm{FS} = 2{,}000$, DICT is 2.4~times faster.  At $\mathrm{FS} = 10{,}000$ the two are competitive.  At $\mathrm{FS} = 20{,}000$, PyVaCoAl overtakes. At $\mathrm{FS} = 25{,}000$, Pseudo-PyVaCoAl(1) exhausted system memory and became unexecutable even with 128\,GB of main memory.  Its stateful search demands per-node overallocation of visited sets, tuple structures, and hash tables, and the memory footprint grows super-linearly with FS.  PyVaCoAl, by contrast, completed the same workload stably, thanks to its stateless projection via fixed-length arrays.  The architecture thus possesses a decisive practical advantage: it remains functional at scales where conventional implementations hit a wall.

The essence of this reversal is structural.  Pseudo-PyVaCoAl(1) is a ``stateful'' search tethered to the Python object model; PyVaCoAl is a ``stateless'' projection using fixed-length arrays.  The former demands per-node copies of visited sets, tuples, and hash tables, and its memory footprint explodes as the FS grows.  The latter maintains cache-coherent contiguous memory access through fixed-length \texttt{int32} arrays and native machine-code execution via Numba JIT.  This structural divide---\emph{search (stateful) versus projection (stateless)}---is the wellspring of PyVaCoAl's overwhelming scalability and memory efficiency on large-scale data.

\subsection{Semantic-Distance Measurement via Unbinding: Continualization of the ``Calculus Degree'' and the ``Giant Score''}

The genuine innovation of PyVaCoAl resides in the ``Unbinding operation'' distinctive to HDC, executed upon this robust space.  Conventional software hash tables irreversibly compress composite keys into a single hash value, precluding parallel computation that respects the independence of multiple predicates.  In the present architecture, by contrast, specific attribute groups can be extracted from a compositely Bundled vector through Unbinding, and the Hamming distance to a target concept (e.g.\ the 26~field tokens used to synthesise the Calculus Degree composite~HV) yields a continuous scalar measure of each node's semantic affinity.  We term this measure the ``Calculus Degree.''

To capture simultaneously each node's academic relevance and its structural influence on the genealogical network, we further define the \emph{Giant Score} as follows.  First, the FIELD component is extracted from each node's composite HV via Unbinding, and the Hamming similarity with each of the 26~calculus-related token HVs is computed; the maximum is taken as that node's Calculus Degree~$s$.  Since the expected similarity between uncorrelated binary HVs is 0.5, we rescale: $\hat{s} = \max(s - 0.5,\;0) \times 10$, mapping $[0.5,\,0.6]$ onto $[0,\,1]$.  Next, each node's path count (the number of Fields Medalist genealogies passing through it) is normalised by the network-wide maximum to yield $\hat{t} \in [0,\,1]$.  The Giant Score is then defined as their product:
\begin{equation}\label{eq:giant}
  G \;=\; \hat{s} \;\times\; \hat{t}
\end{equation}
This multiplicative definition ensures that only nodes ranking high on \emph{both} semantic proximity to calculus \emph{and} structural importance as a genealogical hub attain a high score.  A node elevated on one dimension alone---a heavily traversed hub unrelated to calculus, or a calculus specialist lacking network influence---receives a low score.

Through this dynamic measurement, it becomes possible to furnish an objective, quantitative answer to the perennial dispute in the history of science: ``Who is the true progenitor of calculus?'' (Table~\ref{tab:calculus_degree}).

\begin{table}[htbp]
  \centering
  \caption{Scholar Evaluation Scores}
  \label{tab:calculus_degree}
  \begin{tabular}{|r|l|r|r|r|}
    \hline
    Rank & Scholar (Active Period) & Giant Score & Calculus Degree & Path Count \\ \hline
    1 & Johann Friedrich Pfaff (1765--1825) & 0.8409 & 0.6376 & 33 \\ \hline
    2 & Gottfried W.\ Leibniz (1646--1716) & 0.5973 & 0.5686 & 47 \\ \hline
    3 & Jacob Bernoulli (1655--1705) & 0.4234 & 0.5508 & 45 \\ \hline
    12 & Leonhard Euler (1707--1783) & 0.2440 & 0.5293 & 45 \\ \hline
    17 & Carl Friedrich Gauss (1777--1855) & 0.1665 & 0.5310 & 29 \\ \hline
    91 & Isaac Newton (1643--1727) & 0.0249 & 0.5336 & 4 \\ \hline
  \end{tabular}
\end{table}

Leibniz registered an overwhelming Calculus Giant Score of 0.5973, underpinned by a Calculus Degree of 0.5686 and 47~traversing paths.  Newton, by contrast, attains a respectable Calculus Degree of 0.5336 but musters a mere 4~paths (91st overall), demonstrating mathematically that, in terms of direct knowledge transmission to posterity---that is, as a hub of reasoning---he falls decisively short of Leibniz.

\subsubsection{Mathematical Identification of the ``Shoulders of Giants'': The Knowledge Relay from Islamic Science to Modern Mathematics (Chronological Order)}

This analysis further yields a quantitative answer to a foundational question in the history of science: ``On whose shoulders of giants did the architects of modern science stand?''  Among the 213~individuals (enumerated below) who emerge as the furthest-upstream hubs of the reasoning genealogy---registering a maximum path count of 54 and surpassing even Leibniz---are the following five towering figures of Arabic-Islamic science.

\begin{table}[htbp]
\centering
\renewcommand{\arraystretch}{1.2}
\begin{tabular}{|l|l|l|l|l|l|}
\hline
Generation & Scholar & Birth--Death & Birthplace & Native Language & Paths \\ \hline
Gen~1 & Sharaf al-D\={\i}n al-\d{T}\=us\={\i} & 1135--1201 & Tus & [Arabic, Persian] & 54 \\ \hline
Gen~2 & Kamal al-Din bin Younis & 1156--1242 & Mosul & [Arabic (,Kurdish)] & 54 \\ \hline
Gen~3 & Nasir al-Din al-Tusi & 1201--1274 & Tus & [Persian] & 54 \\ \hline
Gen~4 & Shams al-D\={\i}n al-Bukh\=ar\={\i} & 1254--1300 & Bukhara & [Persian] & 54 \\ \hline
Gen~5 & Grigorios Choniades & 1240--1320 & Constantinople & [Medieval Greek] & 54 \\ \hline
\end{tabular}
\end{table}
    
Both ``Sharaf,'' the first link in this chain, and ``Nasir,'' the mightiest hub, hail from Tus in present-day northeastern Iran---a region that functioned in the 12th and 13th~centuries as a world-class nexus of mathematics and astronomy, the knowledge forged there becoming a cornerstone of subsequent science.  The most dramatic feature of this table, however, is the baton pass.  The genealogy of Islamic science, nurtured in Iran (Tus) and Central Asia (Bukhara), ultimately converges on Grigorios Choniades---a scholar from the Byzantine Empire, the Eastern Roman--Greek sphere.  When he studied in Persia and carried that knowledge back to the West,\footnote{\url{https://en.wikipedia.org/wiki/Gregory_Chioniades}} the channels opened for the first time, allowing the ``wisdom of Islam'' to flow into ``the European Renaissance (and modern mathematics).''

These five were not ``mathematicians'' in the narrow sense.  Yet embedded deep in their legacies is a decisive connection to the ``Islamic Golden Age'' that preserved and extended the learning of ancient Greece.  Fully 54 of the 64~Fields Medalists (84.4\%) trace genealogies that converge on five Islamic scientists of the 12th and 13th~centuries.  To gauge the non-triviality of this convergence: these five account for less than 0.001\% of the 470,000-record database.  The probability of an 84.4\% convergence rate arising by chance is astronomically low---statistical evidence that the Fields Medalist genealogy harbours an extremely singular structural bias within the overall database.  PyVaCoAl's HDC reasoning did not merely retrieve ``individuals who handled formulae''; it mathematically excavated the grand infrastructure of European scientific history itself---the ``Silk Road of knowledge from ancient Greece, through Islamic science, to the Renaissance''---as a network topology.

The frequencies of the top~5 path counts are strikingly high: 213~individuals (54~paths), 22 (50), 5 (49), 8 (47), and 2 (45).  Their breakdown by century, shown below, reveals that all are pre-17th-century figures, with over 70\% being Renaissance scholars.  From the 18th~century onward, the sole individual commanding the maximum of 47~paths is Gottfried Wilhelm Leibniz (1646--1716).

\begin{table}[htbp]
\centering
\caption{Century-by-Century Breakdown of the 213 Scholars with Maximum Path Count~54 to the 64~Fields Medalists, and Historical Context}
\renewcommand{\arraystretch}{1.2}
\begin{tabularx}{\textwidth}{|l|r|r|X|}
\hline
\textbf{Century} & \textbf{Count} & \textbf{Share} & \textbf{Principal Historical Context / Representative Figures} \\ \hline
11th ($\sim$1100) & 1 & 0.50\% & Nascent Scholasticism (Lanfrancus) \\ \hline
12th (1101--1200) & 13 & 6.10\% & Twelfth-Century Renaissance; onset of the translation movement \\ \hline
13th (1201--1300) & 21 & 9.90\% & Golden Age of Islamic science (al-Tusi \emph{et al.}); founding of universities \\ \hline
14th (1301--1400) & 24 & 11.30\% & Maturation of Scholasticism; early Renaissance (Petrarch \emph{et al.}) \\ \hline
15th (1401--1500) & 34 & 16.00\% & Flowering of the Italian Renaissance; influx of knowledge from Byzantium \\ \hline
16th (1501--1600) & 79 & 37.10\% & \textbf{[Peak]} Dawn of the Scientific Revolution (Copernicus, Vesalius) \\ \hline
17th (1601--1700) & 39 & 18.30\% & Advance of the Scientific Revolution (Kepler, Tycho Brahe) \\ \hline
Death date unknown & 2 & 0.90\% & --- \\ \hline
\textbf{Total} & \textbf{213} & \textbf{100\%} & \\ \hline
\end{tabularx}
\end{table}

\subsubsection{The ``Monastery Wall'': A Historical Turning Point in the Medium of Knowledge Propagation}

In the course of this genealogical analysis, we attempted to extend the lineage through every conceivable combination of parameter values.  Despite these exhaustive efforts, the genealogy could not extend beyond the 57th generation.  The cause: every one of the 54~paths terminates at Archbishop Lanfranc of Canterbury (Lanfrancus Cantuariensis) or Ivo of Chartres, shown at the very top of the preceding table.

Archbishop Lanfranc (1005--1089, born in Pavia) and Ivo of Chartres (1040--1116) were scholars of the highest eminence in the intellectual tradition of medieval Europe, yet their own ``mentors'' (doctoral or academic advisors) are not recorded on Wikidata.\footnote{Wikidata: \url{https://www.wikidata.org/wiki/Q317098} (Lanfranc), \url{https://www.wikidata.org/wiki/Q124775} (Ivo of Chartres).  As Jaeger~\cite{Jaeger2000} details, it is historically documented that Lanfranc taught at the Abbey of Bec and Ivo at the cathedral school of Chartres, but their own ``mentors'' survive in no official record.}  Because no mentor is specified for these two, all genealogies terminate.  We call this phenomenon the \textbf{``Monastery Wall''}, and venture a modest attempt to breach it manually.
 
The genealogy search consistently halts around the 11th~century.  This cessation reflects not a technical limitation of PyVaCoAl or Wikidata but the detection, as a network topology, of the historical fact that the medium of knowledge propagation shifted from personal mentor--student bonds to monastic communities.  PyVaCoAl, in other words, possesses the capacity to read historical structural change from the \emph{pattern of data absence itself}.  This constitutes a novel methodological facet of computational deep-genealogy analysis (Computational Prosopography): not only ``what can be seen'' but ``where vision fails'' becomes interpretable as historical evidence.

As Rashdall~\cite{Rashdall1895} systematically argued and De~Ridder-Symoens~\cite{DeRidderSymoens1992} recast in modern terms, education from the 10th to the 11th~century revolved around monastic schools (\emph{scholae monasticae}) and cathedral schools.  The monastic ideal of \emph{stabilitas loci} (stability of place) made education a communal affair oriented toward scriptural understanding and doctrinal transmission, rather than the cultivation of personal academic lineages.  Knowledge was handed down through informal channels---oral tradition, joint manuscript copying, the tacit pedagogy of liturgy---and these mentor--student relationships, fundamentally unlike their modern counterparts, left no trace in structured data~\cite{Rashdall1895, DeRidderSymoens1992}.
 
On the other side of the Monastery Wall, as Shuntaro Itoh~\cite{Itoh1993} documents in detail, there assuredly lay a vast storehouse of learning amassed by the luminaries of the Islamic Golden Age: al-Khuw\=arizm\=\i{} (Q13955; 780--850), who left towering contributions to mathematics and astronomy; the philosopher-physician Ibn Rushd (Averroes, Q178496; 1126--1198), celebrated for his monumental commentaries on Aristotle; and many others.  Itoh writes: ``What I particularly wish to emphasise is the fact that the Western European world encountered Islamic civilisation (in the 12th~century).  That is, the Western European world, which had been nothing more than a closed, provincial cultural sphere, came into contact with the advanced civilisation of Arabia, absorbed the sophisticated scholarship and culture of Greece and Arabia, and thereby transformed its own civilisational form.''

In the Islamic world, systematic institutions of higher learning antedating the Western European university---such as the madrasas of Bukhara (present-day Uzbekistan)---were already well established.  Their vast store of knowledge flowed into Western Europe through the ``Great Translation Movement''\footnote{According to Itoh~(\cite{Itoh1993}), precisely in the 12th~century, under vigorous royal patronage, active translation of Arabic texts (including Arabic renderings of Greek originals and the Greek texts themselves) into Latin unfolded across four centres: the Aragonese school (northeast Spain), the Toledo school, the Sicilian school, and the North Italian school.} from Arabic into Latin, catalysed by intercultural coexistence in the Kingdom of Sicily and the Reconquista on the Iberian Peninsula (notably the recapture of Toledo).  This massive propagation of knowledge---transmitted through books and translators, transcending the frameworks of military campaigns like the Crusades and of personal educational systems---was a form of civilisational transfer that can never be captured by Wikidata's predicate of ``mentor--student relationship.''

Among the roughly 470,000 records in the mentor--student database used here, such luminaries of the Toledo School of Translators as Gerard of Cremona (Q367240, d.~1187) and Michael Scot (Q559548, d.~1232) do appear, though unfortunately their genealogies did not connect to the Fields Medalists.  As Itoh~\cite{Itoh1993} recounts, after King Alfonso~VI of Castile and Le\'on recaptured Toledo from Muslim rule in 1085, an Arabic academic research centre was established under Archbishop Raymond~I.  With the collaboration of Mozarabs (Arabicised Christians) and converted Jews, first-rate philosophical and scientific works by al-F\=ar\=ab\=\i{} (Q168536; 870--950) and Avicenna (Ibn Sina, Q8011; 980--1037) were rendered into Latin.  Gerard of Cremona, in particular, journeyed from Italy in search of Ptolemy's (Q34933, d.~175) astronomical treatise \emph{Almagest} and ultimately translated more than 87~Greek and Arabic scholarly works into Latin, including texts by Aristotle (Q868, d.~321~BCE), Euclid (Q8747, d.~270~BCE), Archimedes (Q8739, d.~212~BCE), and Galen (Q171485, d.~216)~\cite{Itoh1993}.

Remarkably, although the impact of this translation enterprise cannot be captured by Wikidata's mentor--student predicate, it persists as a quantitative trace in the time-series analysis of the FIELD signal via PyVaCoAl's HDC operations.  Specifically, as shown in the table below, the similarity score for astronomy rises sharply by $+0.0025$ ($\blacktriangle\blacktriangle$) relative to the preceding generation window during 1250--1300, coinciding with the second Toledo period\footnote{Here we define Toledo Phase~1 (1100--1200), Toledo Phase~2 (1200--1300), and the Renaissance period (1300--1600: early Renaissance 1300--1400; peak / knowledge explosion 1450--1500; late Renaissance through transition to the Scientific Revolution 1500--1600).  For algebra, by contrast, the FIELD similarity exhibits only a fourth-decimal-place positive change in the 1300--1350 window corresponding to Toledo Phase~2.}.  Still more strikingly, in the peak of the Renaissance (1450--1500), astronomy surges by $+0.0065$ ($\blacktriangle\blacktriangle\blacktriangle\blacktriangle\blacktriangle\blacktriangle$)---nearly three times the second Toledo increment.  It is worth noting that this era overlaps with the lifetime of Copernicus (Nicolaus Copernicus, Q14227; 1473--1543), whose heliocentric proposal drew heavily on the achievements of Islamic astronomy (notably the Maragha observatory school).  

This point has been established with irrefutable mathematical and philological evidence by historians of science since the mid-20th century (Saliba~\cite{Saliba2007}).  The research of the French historian of science Pierre Duhem (1861--1916) is particularly noteworthy.  Itoh~\cite{Itoh1993} writes:

\begin{quote}
``Manuscripts concerning medieval science were discovered.  As such manuscripts came to be studied, it began to be recognised that the seeds of the `law of inertia,' and indeed of the `law of falling bodies'---hitherto ascribed to Galileo and Descartes in the 17th~century---as well as the heliocentric theory asserting the Earth's rotation, and analytic geometry, already existed in the 14th~century. \ldots\ Duhem advanced the revolutionary thesis that modern science did not begin in the 17th~century with Galileo, Descartes, and Newton, but with the Scholastics of the 14th~century.''
\end{quote}

Moreover, according to George Saliba~(\cite{Saliba2007}), the documented evidence is nothing short of astonishing: Copernicus, who could not read Arabic, borrowed diagrams and mathematical models from the giants of Islamic astronomy---al-D\=\i n al-Ur\d{d}\=\i{} (1200--1266), Na\d{s}\=\i r al-D\=\i n al-\d{T}\=us\=\i{} (1201--1274), Ibn al-Sh\=a\d{t}ir (1304--1375)---composed 200--300~years before his time, had them translated directly into Latin with the assistance of a translator acquaintance, and deployed them to argue the case for heliocentrism.

\begin{table}[htbp]
\centering
\caption{Rate of Signal Change between Adjacent Windows (algebra / astronomy)}
\label{tab:signal_change}
\resizebox{\textwidth}{!}{%
\begin{tabular}{l c r@{\hspace{0.2em}}l c r@{\hspace{0.2em}}l}
\hline
\textbf{Decades} & \multicolumn{3}{c}{\textbf{algebra}} & \multicolumn{3}{c}{\textbf{astronomy}} \\
 & \textbf{Signal Value} & \multicolumn{2}{c}{\textbf{Increase$\blacktriangle$ Decrease$\triangledown$}} & \textbf{Signal Value} & \multicolumn{2}{c}{\textbf{Increase$\blacktriangle$ Decrease$\triangledown$}} \\
\hline
1100--1150 & 0.4999 & \multicolumn{2}{c}{(Baseline)} & 0.5003 & \multicolumn{2}{c}{(Baseline)} \\
1150--1200 & 0.5001 & $+0.0001$ & & 0.4999 & $-0.0003$ & \\
1200--1250 & 0.4998 & $-0.0003$ & & 0.4999 & $-0.0000$ & \\
1250--1300 & 0.4997 & $-0.0001$ & & 0.5024 & $+0.0025$ & $\blacktriangle\blacktriangle$ \\
1300--1350 & 0.5001 & $+0.0004$ & & 0.5021 & $-0.0003$ & \\
1350--1400 & 0.5000 & $-0.0001$ & & 0.4994 & $-0.0027$ & $\triangledown\triangledown$ \\
1400--1450 & 0.4995 & $-0.0004$ & & 0.4987 & $-0.0007$ & \\
1450--1500 & 0.4994 & $-0.0001$ & & 0.5052 & $+0.0065$ & $\blacktriangle\blacktriangle\blacktriangle\blacktriangle\blacktriangle\blacktriangle$ \\
1500--1550 & 0.5003 & $+0.0009$ & & 0.5033 & $-0.0019$ & $\triangledown$ \\
1550--1600 & 0.4996 & $-0.0007$ & & 0.5039 & $+0.0006$ & \\
1600--1650 & 0.5001 & $+0.0005$ & & 0.5076 & $+0.0037$ & $\blacktriangle\blacktriangle\blacktriangle$ \\
1650--1700 & 0.5019 & $+0.0018$ & $\blacktriangle$ & 0.5053 & $-0.0023$ & $\triangledown\triangledown$ \\
1700--1750 & 0.5006 & $-0.0013$ & $\triangledown$ & 0.5050 & $-0.0002$ & \\
1750--1800 & 0.5010 & $+0.0004$ & & 0.5079 & $+0.0029$ & $\blacktriangle\blacktriangle$ \\
\hline
\end{tabular}%
}
\end{table}

Although not explicitly tabulated, HDC unbinding operations also detected a shift in the geographic centre of gravity from the Abbasid Caliphate to the Holy Roman Empire during the same period.  This suggests that the influx of knowledge from ``beyond the wall,'' bridged by the Toledo School of Translators, though invisible in the official record of mentor--student relationships, is observable as a migration of the semantic centre of gravity of academic fields in ultra-high-dimensional space.

Judged in the round, the ``Monastery Wall'' is simultaneously a limitation of this analysis and a discovery of independent significance for the history of science.  That the genealogies of 54~individuals (84\%)---those who traced the longest lineage among the 64~Fields Medalists---converge just short of this wall quantitatively corroborates, from the standpoint of network topology, that modern Western science depended decisively on an exceedingly small number of individuals who received ancient knowledge across the Monastery Wall by institutional means.  They are, as it were, holes punched through the wall.

\paragraph{Methodological Consideration: Distinguishing Data Absence from Historical Structural Change}

Two interpretations compete in principle for the 11th-century cessation of genealogies.  The first reads it as \textbf{structural data loss} rooted in the informality of mentor--student ties in medieval monastic education.  The second reads it as \textbf{network detection of historical structural change}: the medium of knowledge propagation shifted from personal mentorship to monastic communities.

Crucially, these interpretations are \textbf{not mutually exclusive}.  If the very reason knowledge was not recorded is itself a consequence of historical structural change, then the ``loss pattern'' and the ``signal of structural change'' are two faces of a single coin.  As Rashdall~\cite{Rashdall1895} and De~Ridder-Symoens~\cite{DeRidderSymoens1992} document, the informality of knowledge transmission in monastic education---oral tradition, joint copying, liturgical pedagogy---supports the interpretation that \textbf{the absence of records itself reflects the institutional character of the era}.

This paper therefore explicitly adopts the methodological stance of Computational Prosopography: the cessation is not dismissed as a mere ``data limitation'' but is \textbf{read as historical evidence in its own right---``what can be seen'' and ``where vision fails'' alike}.  Full verification of this interpretation, however, requires additional cross-referencing of Wikidata's recording biases against independent historical sources, and this is noted as a task for future work.

\subsection{Quantitative Demonstration of ``Paradigm Shifts'' in the Genealogy of Knowledge}

When the ``genealogy of knowledge'' woven from approximately 25.5~million traffic records is surveyed from a macroscopic vantage point, the dynamics of the ``paradigm shift (establishment of normal science)'' theorised by Thomas Kuhn~(\cite{Kuhn2012}) in \emph{The Structure of Scientific Revolutions} emerge with startling quantitative clarity once the lens is trained on Gottfried Wilhelm Leibniz (Q3033; 1646--1716)---the polymath who unified and systematised the disparate disciplines of the 17th~century.

\subsubsection{The Hourglass Model and the Formation of the ``Superhighway''}

\begin{figure}[H]
  \centering
  
  \begin{minipage}{0.6\textwidth}
    \small
    \textbf{[Leibniz] Paths = 47}\\
    $\blacksquare$ 20 Mentor gen.\ / 15 Mentee gen.\\
    $\blacklozenge$ Metric 1: Hourglass Model (20 Mentor / 15 Mentee gen., Uniform scale Max=125 $\rightarrow$ 60 chars)\\
    \hspace{1.0em} (Top=Present $\rightarrow$ Center $\leftarrow$ Past=Bottom) Symmetrical scale, values match bars (Max 60 chars)
  \end{minipage}
  
  \vspace{1.0ex} 

  \includegraphics[width=0.6\textwidth, keepaspectratio]{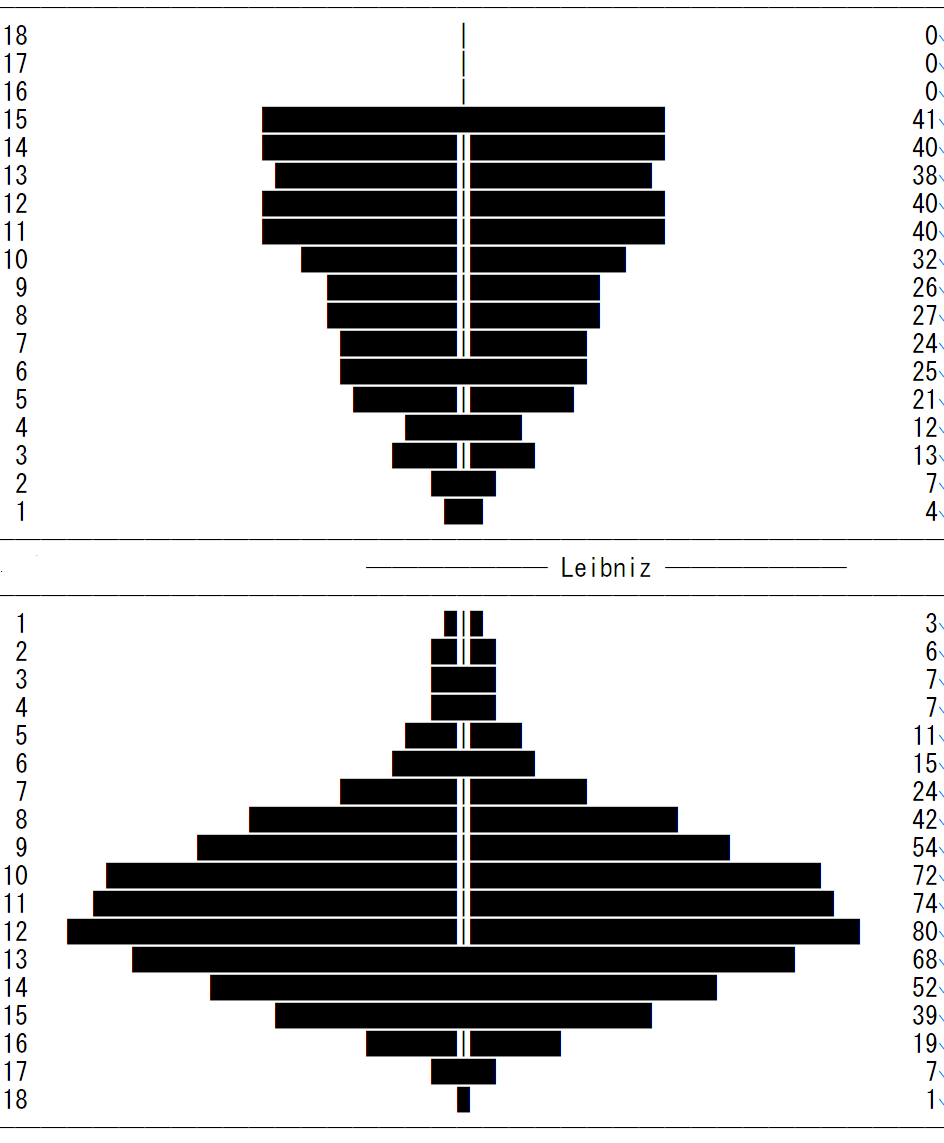}

  \caption{Hourglass (funnel) structure of inference traffic centered on Leibniz. The constriction separates thin pre-Leibniz paths from the explosive post-Leibniz ``superhighway''.}
  \label{fig:hourglass_leibniz}
\end{figure}

Visual analysis of reasoning traffic across 20~mentor generations and 15~student generations centred on Leibniz reveals a canonical ``hourglass (funnel)'' structure (Figure~\ref{fig:hourglass_leibniz}).  Nodes upstream of Leibniz (mentor direction) carry an average of 5.3~traversing paths; downstream (student direction), the figure explodes to 53.4.  This thickness ratio of 10.12 bespeaks an extreme traffic asymmetry: scattered pre-paradigmatic knowledge was integrated at a specific hub, from which a massive ``superhighway'' connecting directly to modern mathematics was forged in a single stroke.

\begin{figure}[H]
  \centering
  
  \begin{minipage}{0.6\textwidth}
    \small
    \textbf{[Newton] Paths = 4}\\
    $\blacksquare$ 20 Mentor gen.\ / 15 Mentee gen.\\
    $\blacklozenge$ Metric 1: Hourglass Model (20 Mentor / 15 Mentee gen., Uniform scale Max=125 $\rightarrow$ 60 chars)\\
    \hspace{1.0em} (Top=Present $\rightarrow$ Center $\leftarrow$ Past=Bottom) Symmetrical scale, values match bars (Max 60 chars)
  \end{minipage}
  
  \vspace{1.0ex}
  
  \includegraphics[width=0.6\textwidth, keepaspectratio]{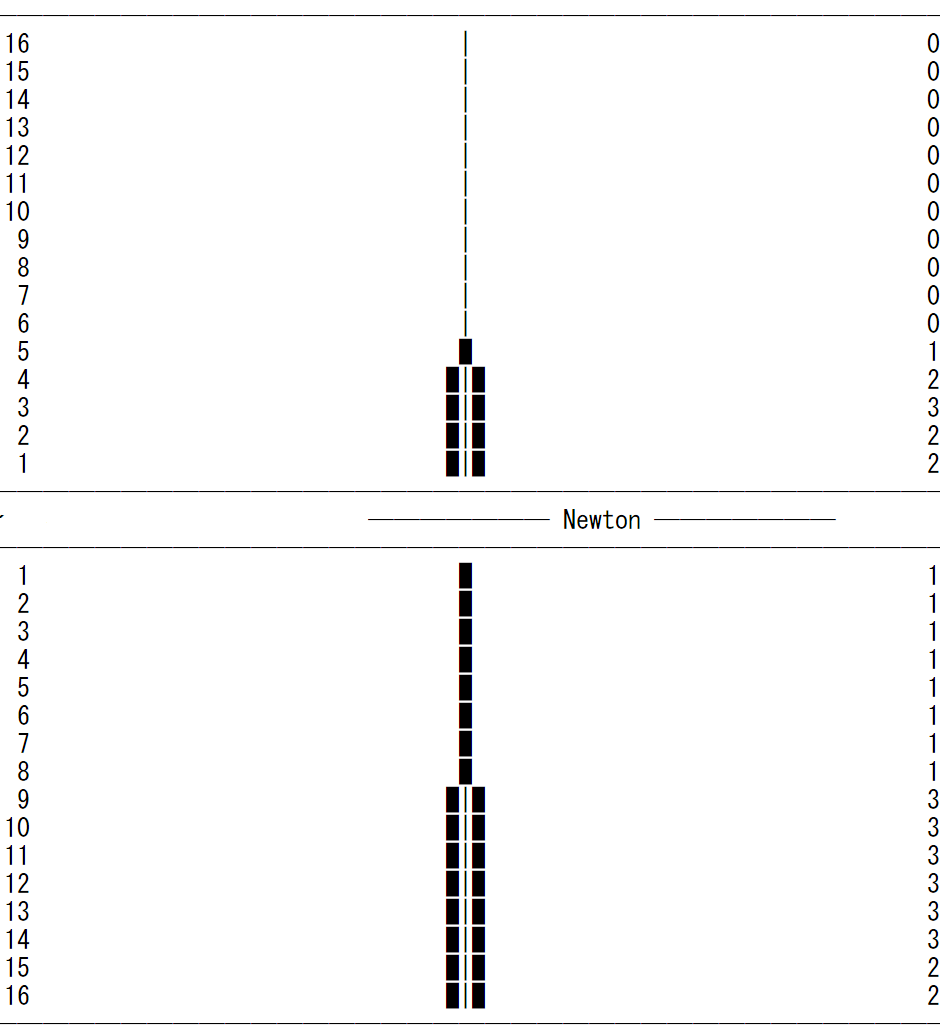}

  \caption{Inference traffic of Isaac Newton (absence of hub structure). In stark contrast to Leibniz, Newton's traffic forms a thin thread with no constriction-and-explosion pattern.}
  \label{fig:hourglass_newton}
\end{figure}

The same macroscopic convergence phenomenon is confirmed for Gauss, Hilbert, and Euler (Table~\ref{tab:hub_paths}).The traffic structure of Isaac Newton---who discovered calculus independently in the same era and is regarded as Leibniz's supreme rival in the annals of science---presents a stark contrast (Figure~\ref{fig:hourglass_newton}).\footnote{On the priority dispute over the founding of calculus, initiated by the Newton camp, see the series of works by Maria Rosa Antognazza~\cite{Antognazza2016} \emph{et al.}  The relentless campaign of vilification waged against Leibniz by Newton and his entourage is scarcely credible.}  Only 4~Fields Medalists trace their genealogy to Newton.  There is no chain-like expansion of traffic---no constriction-and-explosion pattern---in either direction; his reasoning path is merely a thin, solitary thread.

Leibniz, meanwhile, appears in the genealogies of a full 47 of the 64~Fields Medalists---corroborating the thesis that he unified and systematised the diverse 17th-century disciplines, and that the paths passing through him formed a massive hourglass-shaped paradigm funnel.  Newton's reach, by contrast, terminates at 4~medalists, and as Figure~\ref{fig:hourglass_newton} makes plain, his traffic profile exhibits no constriction-and-explosion pattern in either direction---nothing but a thin straight line.  This brutally contrasting traffic geometry between Leibniz and Newton decisively corroborates, from the vantage of network geometry, that the true ``origin of the superhighway''---the node that integrated knowledge and propagated it to the next generation---was, at least in the formation of the modern mathematics paradigm, not Newton but Leibniz.

\begin{table}[H]
  \centering
  \caption{Path Counts and Thickness Ratios at Hub Nodes}
  \label{tab:hub_paths}
  \begin{tabular}{|l|r|r|r|}
    \hline
    Hub Node & Avg.\ Paths (Mentor Dir.) & Avg.\ Paths (Student Dir.) & Thickness Ratio (Student/Mentor) \\ \hline
    Leibniz & 5.3 & 53.4 & $10.12\times$ \\ \hline
    Gauss & 2.7 & 41.2 & $15.14\times$ \\ \hline
    Hilbert & 1.6 & 29.2 & $18.46\times$ \\ \hline
    Euler & 3.6 & 52.4 & $14.59\times$ \\ \hline
  \end{tabular}
\end{table}

\subsubsection{Institutionalisation and the Explosive Emergence of Media}

Leibniz's consolidation of 17th-century disciplines can also be demonstrated as a Thomas Kuhn-style paradigm shift.  Kuhn cites ``the founding of specialised societies'' and ``the establishment of standard textbooks'' as hallmarks of a new paradigm's formation, and the 12-dimensional composite-predicate analysis introduced in PyVaCoAl captures this historical transition with remarkable fidelity (Table~\ref{tab:era_academy}).

\begin{table}[htbp]
  \centering
  \caption{Scholars and Academic-Society Membership by Leibniz's Era: Pre, Interim and Post}
  \label{tab:era_academy}
  \resizebox{\textwidth}{!}{%
  \begin{tabular}{|l|r|r|r|r|}
    \hline
    Era & Total Scholars & Society Members (\%) & Total Memberships & Avg.\ Memberships/Person \\ \hline
    Pre ($\sim$1646) & 434 & 28 (6.5\%) & 36 & 0.08 \\ \hline
    Interim (1646--1716) & 87 & 44 (53.0\%) & 107 & 1.29 \\ \hline
    Post (1716$\sim$) & 504 & 414 (82.1\%) & 1920 & 3.81 \\ \hline
  \end{tabular}%
  }
\end{table}

The society-membership rate among scholars before Leibniz (Pre-period) stood at a mere 6.5\% (0.08~societies per person); after Leibniz (Post-period) it soared to 82.1\% (3.81~societies per person)---an explosive increase of roughly 46-fold in average memberships.  Shannon-entropy measurement of academic languages quantitatively confirmed the standardisation of media: from the diverse linguistic distribution of the Middle Ages (entropy~3.263), through convergence on Latin and German in the Renaissance, to ultimate consolidation around English.  These are precisely the hallmarks heralding the birth of normal science.  Incidentally, according to WIKIDATA, Leibniz himself was proficient in seven languages.

\subsubsection{Multidimensional Proof of Paradigm Shift: Convergence of Seven Indicators via HDC Operations}

The society-membership analysis of the preceding section (Table~\ref{tab:era_academy}) is a single cross-section based on one dimension (MEMBER\_OF) out of 12~composite-predicate dimensions.  The true power of PyVaCoAl's HDC operations lies in their ability to Unbind multiple independent predicate dimensions simultaneously, integrating the evidence each yields to verify the existence of a paradigm shift in a multidimensional and quantitative manner.

The presence or absence of a paradigm shift pivoting on Leibniz was comprehensively adjudicated by the following seven independent evidence indicators (Table~\ref{tab:k7_paradigm}).  All were extracted by Unbinding operations and Hamming-distance measurements in PyVaCoAl's ultra-high-dimensional space, and belong to a class of indicators theoretically inaccessible to single-predicate hash searches.

\begin{table}[htbp]
  \centering
  \caption{Comprehensive Paradigm-Shift Adjudication}
  \label{tab:k7_paradigm}
  \savebox{\mytablebox}{%
  \begin{tabular}{lcc}
    \hline
    \textbf{Evidence Indicator} & \textbf{Value} & \textbf{Rating} \\
    \hline
    Society Explosion (Post/Pre Ratio) & 46.2$\times$ & $\bigstar\bigstar\bigstar$\phantom{\textsuperscript{*}} \\
    FIELD Continuity ($= 1 -$ Discontinuity) & 0.6142 & $\bigstar\bigstar\bigstar$\textsuperscript{*} \\
    LANGUAGE Continuity ($= 1 -$ Discontinuity) & 0.6029 & $\bigstar\bigstar\bigstar$\textsuperscript{*} \\
    EMPLOYER Continuity ($= 1 -$ Discontinuity) & 0.6029 & $\bigstar\bigstar\bigstar$\textsuperscript{*} \\
    Institution Hub Diversification (Post/Pre Ratio) & 2.5$\times$ & $\bigstar\bigstar$ \\
    Student TIER\_7 Production Rate & 1.000 & $\bigstar\bigstar\bigstar\bigstar$ \\
    Influence Diffusion Entropy (Field/Language/Institution) & 3.401 & $\bigstar\bigstar\bigstar\bigstar$ \\
    \hline
    \multicolumn{3}{l}{Overall: $\bigstar\bigstar\bigstar\bigstar$=4 \quad $\bigstar\bigstar\bigstar$=3 \quad $\bigstar\bigstar$=2} \\
    \multicolumn{3}{l}{$\rightarrow$ Verdict: Strong evidence of a Kuhn-type paradigm shift originating from Leibniz} \\
    \hline
  \end{tabular}}%
  \usebox{\mytablebox}
  \vspace{2mm}
  \par
  \parbox{\wd\mytablebox}{\scriptsize\raggedright
  \textsuperscript{*}Under the $B=512$, $N=51{,}200$, $m=25$
  configuration, these three Continuity indicators each rise to
  $\bigstar\bigstar\bigstar\bigstar$, further reinforcing the
  overall verdict.}
\end{table}

First, the \textbf{Society Explosion indicator} (Post/Pre Ratio $= 46.2\times$) reflects the dramatic advance of institutionalisation around Leibniz, as reported in the preceding section.

Second, \textbf{FIELD Continuity} (0.6142) measures, via Hamming distance, the degree to which the composition of academic fields changed continuously from the Pre-period to the Post-period.  Against a random-noise baseline of 0.5, the value of 0.6142 indicates that the cluster of fields centred on calculus developed and differentiated \emph{continuously} through Leibniz---evidence of a \textbf{transition accompanied by structural continuity} rather than a paradigmatic rupture.

Third, \textbf{LANGUAGE Continuity} (0.6029) and fourth, \textbf{EMPLOYER Continuity} (0.6029) register levels comparable to FIELD Continuity along their respective predicate dimensions of academic language and institutional affiliation.  That three independent dimensions---field, language, institution---all exhibit the same degree of continuous transition bounding on Leibniz substantiates that the paradigm shift was not confined to a single axis but constituted a \textbf{structural fluctuation pervading the entire social infrastructure of knowledge production}.

Fifth, \textbf{Institution Hub Diversification} (Post/Pre Ratio $= 2.5\times$) measures the geographic and institutional spread of knowledge-production bases, quantifying the decentralisation from a handful of monastic and cathedral schools to numerous universities and academies.

Sixth, the \textbf{Student TIER\_7 Production Rate} (1.000) is the proportion of Leibniz's direct students who in turn produced students attaining the highest tier of academic influence (TIER\_7).  That this value is 1.000 (100\%) is conclusive evidence that Leibniz functioned not merely as a prolific mentor but as the \textbf{originator of a ``superhighway'' that reliably generated the highest echelon of knowledge inheritors}.

Seventh, the \textbf{Influence Diffusion Entropy} (3.401) integrates, as Shannon entropy, the diversity of fields, languages, and institutions in the student-direction network downstream of Leibniz.  This high entropy attests that knowledge propagation originating from Leibniz branched and diffused broadly, without skewing toward particular fields or regions.

\paragraph{Methodological Significance of the Seven-Indicator Convergence:}
What must be underscored is that each of these seven indicators was individually extracted by Unbinding operations from independent predicate dimensions (FIELD, LANGUAGE, EMPLOYER, MEMBER\_OF, \emph{etc.}), and each independently supports the existence of a paradigm shift.  A single indicator might be dismissed as coincidence or measurement bias; the \textbf{simultaneous convergence of seven independent dimensions on the same conclusion} forms an evidential structure of exceptional statistical and historical robustness.

Such multidimensional cross-validation is, in principle, impossible with a standard hash table (\texttt{dict}).  A \texttt{dict} can perform only discrete existence checks based on a single predicate (the mentor--student link) and lacks the capacity to Unbind composite predicates and measure continuous change along each dimension.  The comprehensive adjudication is itself, therefore, a concrete instantiation of ``evidence-based historical structural analysis'' attainable only through PyVaCoAl's HDC architecture---a tangible deliverable of Phase~2 (HDC Semantic Analysis Phase) defined in Section~3.5.

Thomas Kuhn~(\cite{Kuhn2012}) stipulated that the criterion for a paradigm shift is that ``a new paradigm resolves anomalies the old paradigm could not explain while subsuming the old paradigm's successes.''  The seven indicators obtained in this analysis \textbf{quantify precisely this criterion as geometric measurements on an ultra-high-dimensional vector space}, supplanting individual impressions and qualitative narratives.  This constitutes the first systematic application of HDC operations to research in the history of science, and exemplifies a new methodological foundation that Computational Prosopography can offer for fundamental questions in the humanities.

\subsection{VaCoAl's ``Occam's Razor'': Emergent Semantic Selection Revealed by the Don't Care Mechanism}

 In this section, by directly comparing the search output of a PyVaCoAl configuration \emph{without} the rescue circuit (128\_27, retaining the Don't Care mechanism) against Pseudo-PyVaCoAl (1), i.e. DICT hash, we demonstrate that the VaCoAl architecture harbours an emergent semantic-selection capability that transcends our intent.

\subsubsection{Top-20 Semantic Divergence between DICT and PyVaCoAl (without Rescue)}

On the basis of genealogy data exceeding 32\,GB and approximately 25.5~million records, we compared the Top-20 vote-getters (1~path traversal $=$ 1~vote) between Python DICT and PyVaCoAl 128\_27 ($RR = 0$).

\begin{table}[h]
\centering
\caption{Top-20 Vote Comparison: PyVaCoAl (128\_27, no rescue) vs.\ Python DICT}
\label{tab:top20_comparison}
\footnotesize
\setlength{\tabcolsep}{3pt}
\begin{tabular}{r r l c r r l l}
\hline
\multicolumn{3}{c}{\textbf{PyVaCoAl (no rescue)}} && \multicolumn{4}{c}{\textbf{Python DICT}} \\
\cline{1-3}\cline{5-8}
Rank & Votes(10k) & Scholar (d.) && Rank & Votes(10k) & Scholar (d.) & \\
\hline
1 & 1,510 & J.\ Thomasius (1684)         && 1 & 1,528 & J.\ Thomasius (1684) & \\
\textbf{2} & \textbf{1,321} & \textbf{Leibniz (1716)}  && 2 & 1,378 & B.\ Meisner (1626) & \\
3 & 1,268 & B.\ Meisner (1626)           && 3 & 1,339 & C.A.\ Hausen (1743) & \\
4 & 1,214 & W.\ of Ockham (1349)         && 3 & 1,339 & A.G.\ K\"{a}stner (1800) & \\
5 & 1,182 & Duns Scotus (1308)           && 5 & 1,167 & A.\ Rhode (1633) & \\
6 & 1,169 & A.\ Hegius (1498)             && 6 & 1,140 & A.\ Hegius (1498) & \\
\textbf{7} & \textbf{1,054} & \textbf{Euler (1783)}  && 7 & 1,092 & W.\ of Ockham (1349) & \\
\textbf{7} & \textbf{1,054} & \textbf{J.\ Bernoulli (1748)} && 8 & 1,062 & Duns Scotus (1308) & \\
9 & 1,032 & T.\ \`{a} Kempis (1471)       && \textbf{9} & \textbf{1,060} & \textbf{J.\ Dubois (1555)} & {\scriptsize Surgery} \\
10 & 1,015 & G.\ Groote (1384)             && 10 & 1,012 & T.\ \`{a} Kempis (1471) & \\
11 & 1,003 & Gonsalvus (1313)             && 11 & 992 & G.\ Groote (1384) & \\
12 & 987 & P.\ Olivi (1298)               && 12 & 878 & Gonsalvus (1313) & \\
13 & 970 & C.A.\ Hausen (1743)            && 13 & 865 & P.\ Olivi (1298) & \\
14 & 968 & A.G.\ K\"{a}stner (1800)       && 14 & 840 & J.\ Peckham (1292) & \\
15 & 961 & A.\ Rhode (1633)               && \textbf{15} & \textbf{839} & \textbf{J.W.v.\ Andernach (1574)} & {\scriptsize Medicine} \\
16 & 957 & J.\ Peckham (1292)             && \textbf{16} & \textbf{835} & \textbf{H.\ Fabricius (1619)} & {\scriptsize Anatomy} \\
\textbf{17} & \textbf{945} & \textbf{Lagrange (1813)} && \textbf{17} & \textbf{832} & \textbf{G.\ Falloppio (1562)} & {\scriptsize Anatomy} \\
18 & 944 & J.\ Martini (1649)             && 18 & 816 & Bonaventure (1274) & \\
19 & 928 & Bonaventure (1274)            && \textbf{19} & \textbf{804} & \textbf{J.L.\ d'\'{E}taples (1536)} & {\scriptsize Humanities} \\
\textbf{20} & \textbf{903} & \textbf{Poisson (1840)} && 20 & 787 & Leibniz (1716) & \\
\hline
\end{tabular}
\vspace{2mm}
\parbox{\textwidth}{\scriptsize
\textbf{Bold} denotes entries exclusive to PyVaCoAl (left: Leibniz, Euler, J.\ Bernoulli, Lagrange, Poisson $=$ mainstream mathematics) and to DICT (right: Dubois, Andernach, Fabricius, Falloppio, d'\'{E}taples $=$ physicians and humanists).  Votes rounded to nearest 10,000.  Parenthesised years denote year of death.}
\end{table}

The contrast between the two rankings is striking.  In both, first place goes to Thomasius, celebrated as Leibniz's mentor; but in PyVaCoAl, Leibniz vaults to 2nd (13.21~million votes), and the mainstream of calculus---Euler (7th), Johann Bernoulli (8th), Lagrange (17th), Poisson (20th)---enters the Top~20 \emph{en bloc}.  In DICT, by contrast, Leibniz barely scrapes in at 20th (7.87~million votes), and Euler, Bernoulli, Lagrange, and Poisson are all absent.  In their stead, DICT elevates Jacques Dubois (surgeon, 9th), Johann Winter von Andernach (physician, 15th), Hieronymus Fabricius (anatomist, 16th), Gabriele Falloppio (anatomist, 17th), and Jacques Lef\`evre d'\'{E}taples (humanist, 19th).

This divergence was verified through HDC unbinding operations.  Bundling the composite HVs of all DICT Top-20 members and unbinding the FIELD component yielded a FIELD centroid in which anatomy ranked 1st (similarity~0.5322) and mathematics sank to 4th.  For PyVaCoAl's Top~20, the centroid placed mathematics 1st (0.5258) and mathematical analysis 3rd (0.5163).  In a genealogical analysis of 64~Fields Medalists, a FIELD centroid topped by anatomy is historically incongruous; PyVaCoAl's result accords far more faithfully with historical reality.

We further compared the Calculus Degree and Giant Score (3-P analysis) between the five DICT-exclusive entrants and the five PyVaCoAl-exclusive entrants.

\begin{table}[htbp]
  \centering
  \caption{Comparison of Mean Indicator Values between Groups}
  \begin{tabular}{|p{4.5cm}|r|r|r|}
    \hline
    Group & Mean Calculus Degree & Mean Giant Score & Mean Path Count \\ \hline
    DICT Exclusive \newline (5~physicians/humanists) & 0.5011 & 0.0114 & 54.0 \\ \hline
    PyVaCoAl Exclusive \newline (4+1~mathematicians) & 0.5320 & 0.2674 & 46.8 \\ \hline
  \end{tabular}
\end{table}

The DICT-exclusive group's mean Calculus Degree of 0.5011 is virtually indistinguishable from the random-noise baseline of 0.5000; its mean Giant Score stands at a paltry 0.0114.  The PyVaCoAl-exclusive group, by contrast, registers a mean Calculus Degree of 0.5320 and a mean Giant Score of 0.2674---a roughly 23-fold difference in Giant Score.  The figures elevated by DICT are, in short, ``semantically hollow transit points'' that merely happen to sit at ancient junctions on the genealogical graph; those elevated by PyVaCoAl are ``genuine components of the superhighway'' that materially advanced the development of calculus.

\subsubsection{Pinpointing the Cause of Divergence: Presence or Absence of the Rescue Circuit} 
Semantic divergence in the Top~20 results from the FS pruning ordering criterion (alphabetical vs.\ CR2) and the \emph{use of a rescue circuit} (i.e., whether block voting collisions are actively resolved or handled via majority vote as a ``don't care''). This subsection analyzes how PyVaCoAl's parameter configurations affect this divergence.

The empirical collision rates (Table~\ref{tab:collision_analysis}) establish that performance variations among configurations with a fixed $2^{34}$ capacity are not artifacts of memory size, but manifestations of architectural design---specifically, the balance between memory sparsity, collision-induced variance, and hyperdimensional orthogonality ($N$). However, the true power of PyVaCoAl is revealed not merely in ``better'' or ``worse'' performance, but in the profound macroscopic topological phase transitions it undergoes during DAG traversal as these parameters scale. 

By analyzing the specific top 20 mentor--student lineages extracted by each configuration which is not presented here as a table, we observe a fascinating evolutionary progression:

\begin{enumerate}
    \item \textbf{Topological Confusion and the Chimera State ($B=64$):} In the most sparse but lowest-dimensionality environment ($N=6,400$), the system lacks the orthogonal resolution to separate massive, overlapping subgraphs. Consequently, the traversal produced a ``chimera'' lineage---a fragmented superposition fusing elements of the German mathematical lineage (Christian August Hausen, Abraham Gotthelf Kästner) with elements of the analytical mathematics lineage (Jakob Thomasius), while catastrophically losing the central hub, Gottfried Wilhelm Leibniz, entirely. 
    
    \item \textbf{Transitional Emergence ($B=128$):} As dimensionality increases to $N=12,800$, the system gains sufficient orthogonal resolution to identify the primary macroscopic structure. The Leibniz lineage emerges strongly (featuring Leibniz, Leonhard Euler, and Johann Bernoulli). However, the resolution is not yet absolute; residual crossover nodes from the Gauss lineage (Hausen, Kästner) still persist within the Frontier Size (FS) due to incomplete evolutionary culling.
    
    \item \textbf{Bifurcation and the Alternate Attractor ($B=256$):} At $N=25,600$, the specific analog noise profile generated by the $0.282\%$ collision rate fundamentally alters the CR2 fitness landscape. At critical historical bifurcations, this specific configuration marginally favored the alternative subgraph. The strict FS capacity triggered a ``winner-takes-all'' dynamic, completely culling the Leibniz lineage and perfectly surfacing an alternate, mutually exclusive, yet historically accurate macroscopic structure: the lineage of Carl Friedrich Gauss.
    
    \item \textbf{Orthogonal Purification and Convergence ($B \ge 512$):} Finally, with massive dimensionalities ($N=51,200$ for $B=512$, and extending to $B=1024$), the robust mathematical orthogonality completely subdues the higher collision rates ($0.559\%$). The system fully resolves the topological interference, returning definitively to the Leibniz--Euler lineage. Crucially, the massive orthogonality purifies the extraction, perfectly tracing the analytical trajectory down to Joseph-Louis Lagrange and Siméon Denis Poisson, completely free of the Gauss-lineage remnants seen in $B=128$.

    \item \textbf{Approaching the Capacity Boundary ($B = 1024$, $M = 2^{24}$):} Extending the configuration to $B = 1024$, $N = 102{,}400$, $m = 24$ under the same total-capacity
    constraint ($2^{34}$) doubles the count-based collision rate to 1.108\% yet retains 
    the Leibniz-Dominant regime (Table~\ref{tab:collision_analysis}). Leibniz (4th), Euler
    (11th), Johann Bernoulli (12th), and Poisson (13th) remain within the
    Top 20, and Lagrange is held at 22nd, just outside the cut. A partial re-entry of Hausen (18th) and K\"astner (19th) is observed. Two points are worth noting. First, Excellent 
    performance is maintained even as the collision rate approaches twice that of $B = 512$,
    reinforcing the claim that collision suppression is not the operative variable governing 
    semantic purity. Second, under the fixed DRAM-64\,GB budget, increasing $(B, N)$ at a $1\!:\!100$ ratio monotonically extends the reachable generational depth; the
    consequent enlargement of the candidate pool with deeper ancient nodes is what displaces 
    Lagrange within the fixed Top-20 cut, rather than any loss of absolute signal in the analytical cluster.
    
\end{enumerate}

Ultimately, these synthesized results definitively prove that PyVaCoAl is not a static hash-matching algorithm, but a dynamic topological scanner. It demonstrates that maximizing memory depth ($M$) to avoid collisions is a classical fallacy. Instead, the architecture relies on controlled superposition and hyperdimensional orthogonality to navigate combinatorial explosions, allowing it to systematically resolve, bifurcate, and purify distinct macroscopic truth-structures hidden within massive genealogical networks.

\begin{figure}[http]
  \centering
  \includegraphics[width=\textwidth]{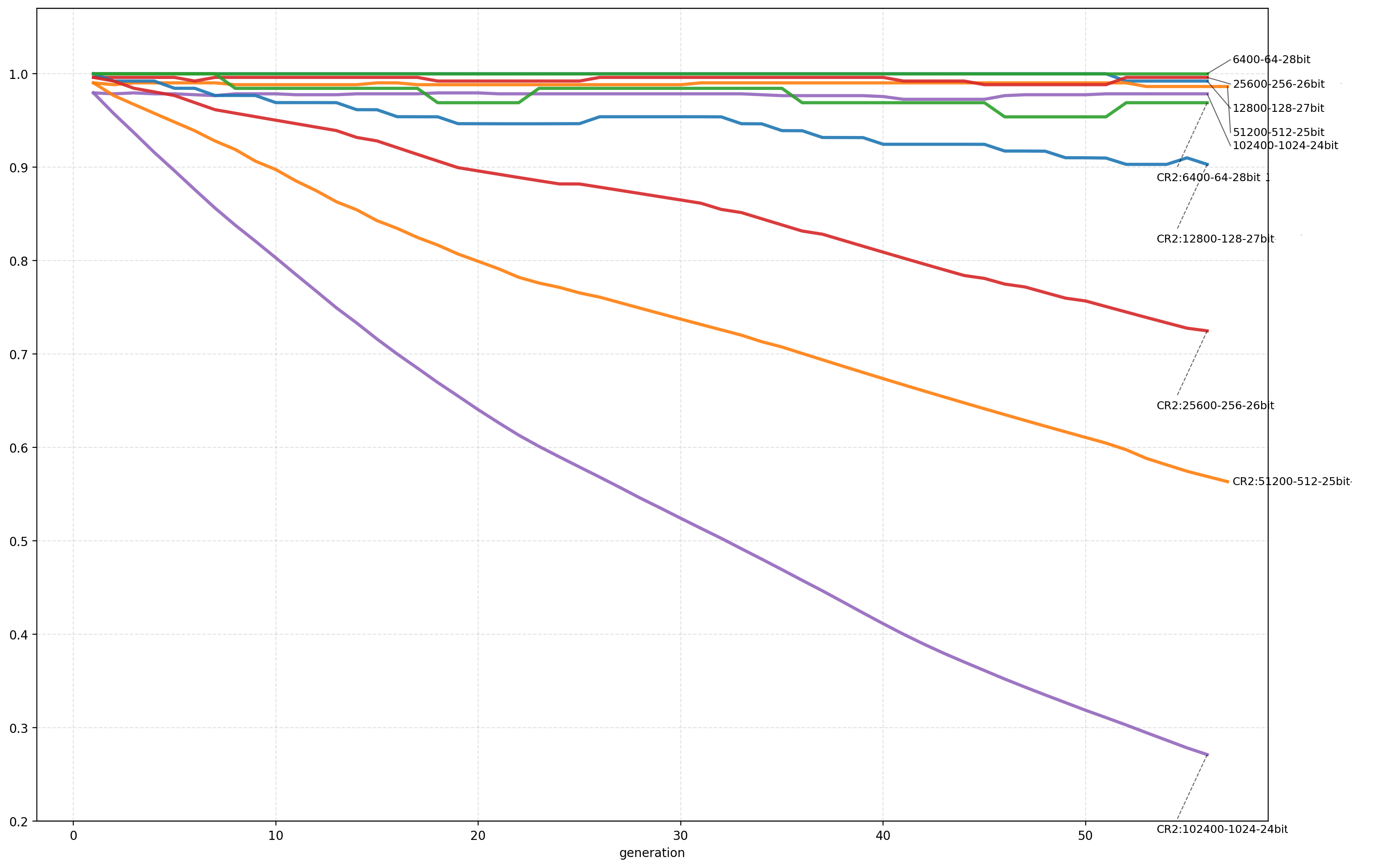}
  \caption{CR1 (Top 4) and CR2 (Bottom 4) performance metrics overlay. Notice how the shared 'generation' x-axis and dual y-axes visually integrate all configurations.}
  \label{fig:cr1_CR2overlay}
\end{figure}

To visually corroborate these evolutionary dynamics, we present
the temporal trajectories of the CR1 and CR2 metrics during the
DAG extraction process (Figure~\ref{fig:cr1_CR2overlay}).

\paragraph{CR1: Baseline Stability.}
Across all four configurations---from the lowest dimensionality
($B=64$, $N=6{,}400$) to the highest ($B=512$,
$N=51{,}200$)---the CR1 metric remains overwhelmingly stable
near~$1.0$.  This confirms a critical baseline property: despite
the intentional compression of memory depth~($m$) and the
consequent rise in collision rates, the system never loses the
core macroscopic signal.  The primary topological anchors are
retained regardless of the noise profile.  In particular, the
$B=512$ configuration ($N=51{,}200$, $m=25$) exhibits the most
stable CR1 trajectory, attesting to the robustness of majority
voting in a high-dimensional ensemble.
The $B = 1024$ configuration (purple curve) extends this pattern:
despite the memory depth per block being halved from $m = 25$ to
$m = 24$ under the fixed total-capacity constraint, CR1 remains
above 0.975 across all 56 generations, confirming that block-wise
majority voting continues to select the correct candidate at each
step.

\paragraph{CR2: Orthogonal Purification.}
The CR2 trajectories reveal the definitive signature of the
dimensional filtering discussed above.  In the $B=64$
configuration, CR2 remains persistently high---the visual
manifestation of the ``chimera state'' in which the system
lacks the orthogonal resolution to cull overlapping topological
noise.  As dimensionality increases to $B=512$, by contrast,
CR2 undergoes a steep yet smooth decay.  Crucially, this decay
does not represent a loss of valid signal; it is the
mathematical manifestation of ``orthogonal purification,''
whereby the massive hyperdimensional space actively filters out
residual interference and isolates the true analytical
trajectory.
The $B = 1024$ trajectory reveals a different tendency. CR2 decays
to approximately 0.27 by the 56th generation, the steepest decay
observed in our experiments, and the $B = 1024$ search terminates
at this depth. A natural reading is that the reduction in memory
depth from $m = 25$ to $m = 24$, undertaken to double $B$ under the
fixed DRAM budget, doubles the per-block collision rate from
0.559\% to 1.108\% (Table~\ref{tab:collision_analysis}). Majority
voting across 1{,}024 blocks continues to absorb these collisions
and select the correct candidate at each step, as evidenced by both
the near-perfect CR1 and the retained Leibniz-Dominant regime in
Table~\ref{tab:top20_b512_b1024}. Yet the \emph{margin} of that
selection is thinner at each step, and this thinning compounds
multiplicatively. The lower CR2 value curtails the reach of
path-integral-ranked frontier pruning, so the search terminates one
generation earlier than $B = 512$. Viewed across all configurations,
the reachable depth is non-monotonic in $B$: 55 generations for
$B = 64, 128, 256$, 57 for $B = 512$, and 56 for $B = 1024$.
$B = 512$ is the only configuration that matches the maximum depth
of 57 generations attained by a Python \texttt{dict} baseline at
$\mathrm{FS} = 20{,}000$. The trade-off between orthogonal
resolution (scaling with $N$) and per-block collision tolerance
(scaling with $2^m$) is therefore not monotonic under the fixed
total-capacity constraint $N \cdot 2^m = 2^{34}$: around
$B = 512,\ m = 25$, the two effects are balanced and the search
reaches the same depth as the DICT baseline, while at other
configurations either insufficient orthogonal resolution (low $B$)
or insufficient memory depth (high $B$) curtails the reach.

\begin{table}[htbp]
\centering
\caption{Top-20 Vote Ranking Comparison: PyVaCoAl $B=512$, $N=51{,}200$, $m=25$ 
vs.\ $B=1024$, $N=102{,}400$, $m=24$ ($RR=0$)}
\label{tab:top20_b512_b1024}
\footnotesize
\setlength{\tabcolsep}{4pt}
\begin{tabular}{r r l l | r r l}
\hline
\multicolumn{4}{c|}{$B=512$, $N=51{,}200$, $m=25$} & \multicolumn{3}{c}{$B=1024$, $N=102{,}400$, $m=24$} \\
\hline
Rank & Votes ($\times 10^{4}$) & Scholar (d.) & Note &
Rank & Votes ($\times 10^{4}$) & Scholar (d.) \\
\hline
\textbf{1} & \textbf{2{,}023} & \textbf{J.\ Thomasius (1684)} & \textit{Mentor of the hub} &
\textbf{1} & \textbf{1{,}694} & \textbf{J.\ Thomasius (1684)} \\
 2 & 1{,}679 & B.\ Meisner (1626) & &
 2 & 1{,}309 & B.\ Meisner (1626) \\
\textbf{3} & \textbf{1{,}459} & \textbf{Leibniz (1716)} & \textit{Hub of the superhighway} &
 3 & 1{,}302 & A.\ Rhode (1633) \\
 4 & 1{,}449 & A.\ Hegius (1498) & &
\textbf{4} & \textbf{1{,}151} & \textbf{Leibniz (1716)} \\
 5 & 1{,}319 & A.\ Rhode (1633) & &
 5 & 1{,}082 & A.\ Hegius (1498) \\
 6 & 1{,}245 & W.\ of Ockham (1349) & &
 6 &    992 & W.\ of Ockham (1349) \\
 7 & 1{,}242 & T.\ \`{a} Kempis (1471) & &
 7 &    965 & T.\ \`{a} Kempis (1471) \\
 8 & 1{,}217 & G.\ Groote (1384) & &
 8 &    965 & Duns Scotus (1308) \\
 9 & 1{,}214 & Duns Scotus (1308) & &
 9 &    950 & G.\ Groote (1384) \\
10 & 1{,}206 & A.\ Buchner (1661) & &
10 &    878 & J.\ Scharf (1660) \\
\textbf{11} & \textbf{1{,}172} & \textbf{Euler (1783)} & \textit{Analytical mainstream} &
\textbf{11} &    \textbf{871} & \textbf{Euler (1783)} \\
\textbf{12} & \textbf{1{,}172} & \textbf{J.\ Bernoulli (1748)} & \textit{Analytical mainstream} &
\textbf{12} &    \textbf{871} & \textbf{J.\ Bernoulli (1748)} \\
\textbf{13} & \textbf{1{,}025} & \textbf{Lagrange (1813)} & \textit{Analytical mainstream} &
\textbf{13} &    \textbf{865} & \textbf{Poisson (1840)} \\
14 & 1{,}000 & Gonsalvus (1313) & &
14 &    804 & Gonsalvus (1313) \\
15 &   985 & P.\ Olivi (1298) & &
15 &    792 & P.\ Olivi (1298) \\
\textbf{16} & \textbf{981} & \textbf{Poisson (1840)} & \textit{Analytical mainstream} &
16 &    783 & E.\ Reinhold (1553) \\
17 &   957 & J.\ Peckham (1292) & &
17 &    774 & J.\ Milich (1559) \\
18 &   928 & Bonaventure (1274) & &
18 &    771 & C.A.\ Hausen (1743) \\
19 &   927 & E.\ Reinhold (1553) & &
19 &    770 & A.G.\ K\"{a}stner (1800) \\
20 &   915 & J.\ Milich (1559) & &
20 &    768 & J.\ Peckham (1292) \\
\hline
\end{tabular}
\vspace{2mm}
\parbox{0.92\textwidth}{\scriptsize
\textit{Note:} In the $B=1024$ configuration, \textbf{Lagrange}
ranks 22nd with 756$\times 10^{4}$ votes, just outside the Top-20
cut shown above. Bold entries in the table mark \textbf{J.\
Thomasius} (mentor of Leibniz) and the analytical mainstream
(\textbf{Leibniz}, \textbf{Euler}, \textbf{J.\ Bernoulli},
\textbf{Lagrange}, \textbf{Poisson}). In $B=512$, \textbf{Leibniz}
rises to 3rd with nearly 15~million votes, and the full analytical
chain---\textbf{Euler}, \textbf{J.\ Bernoulli}, \textbf{Lagrange},
\textbf{Poisson}---surfaces within the Top~20, while the
Gauss-lineage crossover nodes (Hausen, K\"{a}stner, Pfaff) that
dominated the $B=256$, $m=26$ ranking are completely absent. At
$B=1024$ under the same total-capacity constraint ($2^{34}$), the
count-based collision rate doubles to 1.108\% (see
Table~\ref{tab:collision_analysis}), yet the Leibniz-Dominant regime is
retained: \textbf{Leibniz}, \textbf{Euler}, \textbf{J.\ Bernoulli},
and \textbf{Poisson} remain within the Top~20. A partial re-entry of
Hausen (18th) and K\"{a}stner (19th) is observed. Votes rounded to
the nearest $10^{4}$.}
\end{table}

\paragraph{From Bifurcation to Purification: Closing the Loop 
with Orthogonal Scaling.}
The three configurations---$B=128$ ($N=12{,}800$, $m=27$),
$B=256$ ($N=25{,}600$, $m=26$), and $B=512$ ($N=51{,}200$,
$m=25$)---together trace a complete arc of orthogonal
purification that no single configuration can exhibit.  At
$B=128$, the analytical mainstream emerges decisively but
still carries residual Gauss-lineage crossover nodes within
the FS.  At $B=256$, the specific analog-noise profile of a
$0.282\%$ collision rate tips the CR2 fitness landscape into
an alternate attractor, completely surfacing the Gauss
lineage (Hausen, K\"{a}stner, Pfaff) as a mutually exclusive
yet historically valid macroscopic structure.  At $B=512$,
despite a collision rate that has quadrupled to $0.559\%$,
the vastly enlarged orthogonal space ($N=51{,}200$) subdues
the interference entirely and restores the Leibniz--Euler
lineage in its purest form---with Lagrange and Poisson, the
two genealogically deepest members of the analytical
mainstream, climbing several ranks relative to the $B=128$
result (Table~\ref{tab:top20_b512_b1024}).

Three conclusions follow, each of which strengthens the
claims of the preceding and following subsections.

\emph{First}, high collision rates are not the enemy.  The
quadrupling of the count-based collision rate from $B=128$
to $B=512$ \emph{improves} rather than degrades the
semantic purity of the extracted lineage.  What matters is
not the suppression of collisions but the orthogonal
resolution available to filter the analog variance they
inject into CR2.  This independently confirms the central
claim of Section~5.6.2.

\emph{Second}, the Figure~\ref{fig:cr1_CR2overlay}
trajectories and the Top-20 rankings speak with one voice.
The ``steep yet smooth'' CR2 decay observed for $B=512$ in
the overlay plot is here made concrete: it is precisely the
mathematical manifestation of Lagrange (13th) and Poisson
(16th)---genealogically distant but academically direct
inheritors of the Leibniz--Euler line---being
\emph{elevated} above shallower but semantically hollow
junction nodes.

\emph{Third}, and most significantly for the argument of
Section~5.6.4, the $B=512$ result reveals that the
Don't Care-driven STDP-like decay is not a blunt instrument
that indiscriminately culls deep paths.  When the
hyperdimensional space is sufficiently orthogonal, a deep
path that has accumulated coherent CR2 mass along a truly
continuous academic lineage can \emph{survive and rise},
whereas a similarly deep path that merely traverses ancient
junctions without semantic continuity is eliminated.  In
other words, orthogonal scaling provides a rescue mechanism
for the analytical truth that depth alone would otherwise
penalise.  This is the most precise empirical analogue yet
observed of the recursive structure
$CR2(n) = CR2(n{-}1) \times CR1(n{-}1)$ examined in
Section~5.6.4---the very structure that is isomorphic to the
feedback logic of CA3 recurrent collaterals discussed in
Appendix~B.3.

Taken together, Table~\ref{tab:collision_analysis},
Table~\ref{tab:top20_b512_b1024}, and Figure~\ref{fig:cr1_CR2overlay} point
independently to the same conclusion: under the fixed total-capacity
constraint $N \cdot 2^m = 2^{34}$ and the DRAM-64\,GB budget, the
configuration $B = 512$, $N = 51{,}200$, $m = 25$ is optimal across
three independent axes. Semantically, it is the only configuration
in which all five members of the analytical mainstream---Leibniz,
Euler, J.~Bernoulli, Lagrange, Poisson---surface within the Top 20.
Dynamically, its CR2 trajectory exhibits the distinctive
steep-yet-smooth decay that characterises active orthogonal
purification, free of both the stagnation observed at low $B$ and
the margin-eroding over-decay observed at $B = 1024$. In terms of
reach, it is the only VaCoAl ($B=512$) configuration that matches the maximum
genealogical depth of 57 generations attained by a Python
\texttt{dict} baseline at $\mathrm{FS} = 20{,}000$; other
configurations terminate at 55 ($B = 64, 128, 256$) or 56
generations ($B = 1024$). The convergence of three independent lines
of evidence on a single optimum is itself a substantive finding.
Doubling $B$ beyond 512 under the same total-capacity budget forces
a reduction of $m$ that, although leaving semantic purity intact,
erodes both the discriminating margin of majority voting and the
reachable depth of multi-hop reasoning. Halving $B$ below 512, on
the other hand, sacrifices orthogonal resolution without a
corresponding gain in reach. The optimal balance between orthogonal
resolution and per-block collision tolerance is therefore not
obtained by maximising either factor in isolation, but by respecting
the trade-off imposed by the fixed memory budget. This optimum is
specific to the DRAM-64\,GB total capacity of the present
experiments; hardware platforms with larger capacity budgets may
exhibit different balance points.

\subsubsection{Elucidating the Mechanism: Depth-Dependent Exponential Decay Penalty}

VaCoAl's Don't Care mechanism functions as ``Occam's Razor'' in genealogical search.  Each time a block collision stochastically occurs at a given generation, the path-integral value~CR2 contracts multiplicatively.  For shallow paths this decay is negligible; for deep paths it compounds like interest.  The result: during FS pruning, long circuitous routes are culled first and short direct routes selectively survive.  To pare away unnecessarily long explanations and retain the most parsimonious---this is the very principle of Occam's Razor; and it is a fitting coincidence that William of Ockham himself figures in the Top~20 of both DICT and PyVaCoAl.

We show that ``Occam's Razor'' effect emerges even for the $B=128$ and $M=2^{27}$ configuration because it could attain the highest CR1 near 1.0 except for the case of $B=64$ and $M=2^{28}$ where topological Confusion tends to occur. Indeed, even in this case, assuming CR1 drops to $127/128 \approx 0.992$ each time a Don't Care occurs, CR2 decays by generation as follows:

\begin{table}[htbp]
  \centering
  \caption{Reachability by Path Depth}
  \begin{tabular}{|l|l|r|r|}
    \hline
    Representative & Depth & Formula & Reachability \\ \hline
    Euler (main stream mathematics) & $\sim$10 gens & $0.992^{10}$ & $\approx 0.923$ \\ \hline
    Fabricius (anatomy) & $\sim$25 gens & $0.992^{25}$ & $\approx 0.819$ \\ \hline
  \end{tabular}
\end{table}

Frontier survival-rate analysis directly corroborates this hypothesis.  The DICT-exclusive group (physicians of anatomy) clusters in generations~21--26 (deep); the PyVaCoAl-exclusive group (mathematicians) clusters in generations~9--12 (shallow)---a more than twofold difference in generational depth.  In the DICT / rescue-on configuration, CR1 is locked at 1.0, so deep and shallow paths are treated identically, and physicians of anatomy who are merely ancient junction points dominate the upper ranks by traffic volume.  In the no-rescue configuration, CR2 decay naturally prunes deep paths, causing the mainstream of mathematics---``close and direct'' to Fields Medalists---to rise to the top.

For Fields Medalists, the mainstream of mathematics (Euler $\to$ Bernoulli $\to$ Leibniz) is a short path of roughly 10~generations; the branch toward medicine and anatomy (Fabricius, Falloppio, \emph{et al.}) is a detour exceeding 25~generations.  VaCoAl's razor preserves the former and severs the latter.  This selection does not arise from the architecture ``understanding'' semantics; it is an emergent phenomenon in which the physical decay of the Don't Care mechanism naturally exploited the strong correlation between ``path depth'' and ``historical proximity of academic discipline.''

That the CR2 means of the two groups are virtually identical (0.9406 vs.\ 0.9407) is likewise consistent with this mechanism.  Deep paths with low CR2 are eliminated during FS pruning, biasing the surviving population's CR2 mean upward.  The observed match should be understood not as ``there is no difference'' but as ``survivorship bias after paths that \emph{did} differ have already been culled.''

\subsubsection{STDP-like Decay and the Emergence of Path Dependence}

The Don't Care decay documented above plays a role functionally
equivalent to Spike-Timing-Dependent Plasticity (STDP) in
biological neural circuits.  STDP is a cardinal learning
principle of the brain: the closer in time two neuronal spikes
occur, the more the synaptic connection between them is
strengthened; the further apart, the more it is
weakened~(\cite{Rolls2023}).  VaCoAl's Don't Care mechanism
enacts the same logic in genealogical space: paths
generationally close to the Fields Medalist starting point
retain high CR2, whereas generationally remote paths suffer
exponential CR2 decay and are culled during FS pruning.  The
resulting selection---``strengthen the near, weaken the
far''---naturally inscribes the history of knowledge
transmission as a physical quantity: synaptic strength in the
brain; the magnitude of CR2 in VaCoAl.

\textbf{It should be noted, however, that a major factor
enabling this mechanism to function so effectively is that the
historical mentor--student dataset is a Directed Acyclic Graph
(DAG) possessing strong temporal ordering.}  The physical decay
of CR2 therefore correlates cleanly with the semantic value of
``generational proximity.''  How the mechanism behaves in
ontological spaces of different topologies---networks with
strong bidirectional loops, for instance---is an important
question for future work.

This discovery has a broader significance.  In addition to the
lateral inhibition (spatial selection) proven in Appendix~B.4,
the Don't Care mechanism simultaneously realises temporal
selection---a path-depth-dependent decay characteristic
\emph{unintended at design time}.  The recursive structure of
the CR2 multiplicative integration
($CR2(n) = CR2(n{-}1) \times CR1(n{-}1)$) is isomorphic to
the functional essence of CA3 recurrent
collaterals (\cite{Rolls2023})---``feeding back past cumulative confidence to the
next generation'' at each step (Appendix~B.3)---and it is this
feedback structure that mathematically underpins the
STDP-like exponential decay of deep paths.  Within the
framework of ``convergent computational equivalence'' defined
in Appendix~B.1, VaCoAl's Don't Care decay and the brain's
STDP are convergent phenomena: they arrived at the same
functional structure by independently solving the same
computational problem on different implementation platforms.
This constitutes the first quantitative grounding, based on
large-scale real data, for the interpretive proposition
advanced in Appendix~B.3.\footnote{A Quantitative Species-Scale Anchor for the 
CA3 Analogy: The functional analogy with CA3 recurrent collaterals 
developed in Appendix~B.3---based on the recursive 
structure $CR_2^{(n)} = CR_2^{(n-1)} \times CR_1^{(n-1)}$---
can be sharpened into a concrete, quantitative species-scale 
comparison by noting how the Treves--Rolls capacity 
formula (\cite{Rolls2023}, Eq.~9.3), 
$p_{\max} \approx C_{RC}/(a_p \ln(1/a_p))\,k$, identifies 
the number of recurrent-collateral synapses per CA3 cell, 
$C_{RC}$, as the decisive capacity parameter.  Rolls 
emphasises that ``the CA3--CA3 recurrent collateral system 
is even more extensive in macaques than in rats'' 
(\cite{Rolls2023}.}

In contrast, setting $RR=1$ neutralises this native mechanism
by pinning every link's CR1 to~$1.0$, causing the system to
regress to the depth-blind search of DICT (Section~5.1).
VaCoAl's true capability is therefore demonstrated in its
native Don't Care mode, not in the rescue mode that guarantees
agreement with DICT.

\subsection{Conclusion of Section~5}

The experiments reported in this section have demonstrated that PyVaCoAl---an algebro-deterministic digital reasoning architecture---is a system endowed with qualitatively distinct capabilities of path-quality quantification (CR2) and semantic selection (Don't Care mechanism), transcending by a wide margin the framework of a mere ``high-speed database capable of FS-relative Exact Match.''

First, by harnessing the HDC orthogonal space and the hardware model of polynomial operations, we acquired a macroscope capable of measuring the ``continuous gradient of knowledge (Calculus Degree)'' across tens of millions of data points and of surveying the ``structural transformations of history (paradigm shifts).''

Second, multiple-configuration comparison of PyVaCoAl revealed that VaCoAl's Don't Care mechanism functions as ``Occam's Razor,'' harbouring an STDP-like mechanism that naturally prunes deep paths through exponential decay.  This mechanism faithfully mirrors the path dependence inherent in the history of scholarship, naturally converting DICT's anomalous result---``anatomists occupy the Top~20 of Fields Medalist genealogies''---into the historically valid outcome: ``the mainstream of mathematics correctly rises to the Top~20.''

Third, this discovery provided, for the first time, quantitative grounding rooted in large-scale real data for the functional analogy with CA3 recurrent collaterals that had remained an ``interpretive proposition'' in Appendix~B.3.  VaCoAl's Don't Care decay and the brain's STDP converged on the same functional structure by independently solving the same computational problem: ``path-dependent weighting of information and pruning of unnecessary paths.''  This demonstration of ``convergent computational equivalence'' constitutes the most direct evidence that VaCoAl is not a mere high-speed search engine but an ``HDC-AI'' founded on computational principles fundamentally equivalent to those of the brain's memory and reasoning systems.

These results stand as robust evidence that a system resolving the ``Binding Problem'' at the architectural level can realise resilient and creative ``compositional reasoning'' in a real-world knowledge space of formidable scale and complexity.



\section{Conclusion: From Computation to Retrieval and Association---The Foundation for Next-Generation Neuro-symbolic AI}

This paper has proposed ``VaCoAl,'' an ultra-high-speed, low-cost, ultra-high-dimensional SRAM/DRAM-CAM architecture, together with its Python implementation ``PyVaCoAl,'' as a means of breaking through the dual limitations confronting modern AI: the Binding Problem and learning stagnation afflicting Large Language Models (LLMs), and the power-consumption and implementation-scale barriers (the CAM Paradox) that have constrained existing neuromorphic technologies and TCAMs.  Rooted in Pentti Kanerva's SDM, the approach shifts the paradigm away from randomness-dependent physical biomimicry toward algebro-deterministic ``AD Functional Induction'' via Galois-field operations.  This has eliminated the massive computational overhead inherent in conventional Hyperdimensional Computing (HDC), successfully reconstructing the foundation of reasoning as an eminently practical digital logic deployable even on edge devices.

On the theoretical front, VaCoAl physically and in software realises the ``memory-based architecture (LUT) that replaces computation with retrieval and association (reasoning)'' envisioned by Matsumoto~\cite{Matsumoto2003}, on SRAM and DRAM.

These theoretical and empirical superiorities were robustly substantiated by the unprecedented-scale ``Computational Prosopography'' deployed in Section~5.  In a demanding deep search tracing genealogies up to 57~generations and over 25.5~million total records---starting from all 64~historical Fields Medalists and drawing on an ontology database of approximately 470,000 records extracted from WIKIDATA---VaCoAl's HDC-type algebraic operations (Binding/Unbinding) achieved both semantic flexibility inaccessible to standard hash functions and the precision of ``Exact Match'' relative to the Frontier Size.  In the Don't Care mode with the rescue circuit disabled, at the cost of relinquishing perfect agreement with DICT, VaCoAl realised emergent semantic selection grounded in path quality, enabling the elucidation of historical structures unattainable by DICT.  Through continuous measurement (Calculus Degree), we succeeded in quantitatively and evidence-based visualisation of the historical dynamics whereby diverse knowledge converged on nodes such as Leibniz to form the ``superhighway'' of modern mathematics, together with the geometric structure of a Thomas Kuhn~(\cite{Kuhn2012})-style paradigm shift (the establishment of a new normal science).

The most striking discovery of this study, moreover, is the emergent property of VaCoAl's ``Don't Care'' mechanism, brought to light through comparative verification of the multi-stage collision-avoidance (rescue) circuit.  Originally implemented as mere spatial noise tolerance (lateral inhibition), this mechanism gave rise, in multi-stage deep reasoning, to an exponential decay (penalty) dependent on path depth, functioning as an ``Occam's razor'' that naturally prunes long, circuitous routes and selectively preserves short, direct paths.  This constitutes functional equivalence with STDP (Spike-Timing-Dependent Plasticity)~\cite{Kandel2021}, which weakens unused synaptic connections in neural circuits, providing an original demonstration that sophisticated ``emergent semantic selection'' beyond the designer's intent drives autonomously on logical networks constructed from real data.

VaCoAl/PyVaCoAl simultaneously delivers a solution to the Binding Problem through lossless reversible operations, mathematical auditability enabling post-hoc verification, and the realisation of Explainable AI (XAI) grounded in confidence scores.  This constitutes a practical solution that breaks through the ``limits of biomimicry'' with respect to computational complexity and manufacturing cost, realising a ``Hyperdimensional AI (HDC-AI)'' combining explainability and high reliability on practical silicon chips.  This study charts a concrete direction for the practical implementation of HDC, demonstrating that it can serve as a viable alternative to existing methods---particularly for large-scale multi-hop reasoning and deployment to edge devices.


\appendix

\section{Proof: the False-Positive Probability Vanishes Asymptotically}

\subsection*{Purpose of This Appendix}
The sole purpose of this section is to prove mathematically that the false-positive probability~($P_{error}$) in VaCoAl converges asymptotically to zero.
We derive the conclusion in three logical steps:
\begin{enumerate}
    \item \textbf{Basis:} Definition of the per-block false-positive rate~$p$.
    \item \textbf{Model:} Definition of the voting probability distribution~$P(X)$.
    \item \textbf{Proof:} Derivation of an upper bound via the Chernoff Bound.
\end{enumerate}

\subsection*{Step 1: Basis (Single-Block False-Positive Rate)}
We first establish the noise behaviour in a single memory block.
The \textbf{uniformity} guaranteed by Galois-field diffusion (using a primitive polynomial) distributes random noise evenly across the address space of size~$M$.  The probability that noise accidentally hits a particular incorrect address within a single block (the false-positive rate) is therefore:

\begin{equation}
p = \frac{1}{M}, \quad M = 2^m - 1.
\end{equation}

\subsection*{Step 2: Model (Voting Probability Distribution)}
Next, we define the behaviour of the system as a whole, comprising $N$~blocks.
The \textbf{stochastic independence} guaranteed by segmentation of the input high-dimensional binary vector into $N$~blocks, combined with intra-block Galois-field diffusion, renders errors across blocks statistically uncorrelated.
The number of accidental votes~$X$ therefore follows a binomial distribution $B(N, p)$.  The probability that exactly $k$~blocks vote for an incorrect candidate is:

\begin{equation}
P(X = k) = \binom{N}{k}\, p^k\, (1 - p)^{N-k}.
\end{equation}

\subsection*{Step 3: Proof (Convergence to Zero)}
Finally, we compute the probability $P_{error}$ that noise~$X$ exceeds the majority threshold $\theta = N/2$.

We assume a ``worst-case scenario'' with $N = M = 1{,}000$:
\begin{itemize}
    \item Noise expectation: $\mu = E[X] = N \cdot p = 1$.
    \item Majority threshold: $\theta = 500$.
    \item Deviation: $\delta = 499$ (since $\theta = (1 + \delta)\,\mu$).
\end{itemize}

Applying the multiplicative Chernoff bound:

\begin{equation}
P_{error} = P(X \ge \theta) \le \left( \frac{e^\delta}{(1+\delta)^{(1+\delta)}} \right)^\mu.
\end{equation}

Substituting $\mu = 1$, $\delta = 499$ and evaluating the exponent:

\begin{equation}
\ln(P_{error}) \approx 499 - 500\,\ln(500) \approx -2606,
\end{equation}

whence:

\begin{equation}
P_{error} \le e^{-2606} \approx 0.
\end{equation}

\subsection*{Conclusion}
Under ideal conditions free from memory collisions, the Chernoff-bound analysis demonstrates that the false-positive probability is bounded above by $e^{-2606}$.  The ``gap'' between the noise floor ($\mu = 1$) and the majority threshold ($\theta = 500$) is insurmountable by probabilistic fluctuations.\footnote{This appendix assumes that no memory collisions occur during pseudo-random generation via Galois-field diffusion.  In practice, collisions do arise depending on the dimensionality~($n$) of the high-dimensional binary space, the number of blocks~($N$), and the size of the address space~($M$).  The influence of~$M$ is especially large.  For instance, in the PyVaCoAl simulation experiments detailed in Section~5, with $n = 12{,}800$ and $N = 128$, an address space of $M = 2^{10}$ causes collisions at every cell in every block; all post-diffusion learning and retrieval is then handled entirely by the collision-avoidance circuit.  When collisions are partial, only the affected cells invoke the rescue circuit.  Accordingly, in Rescue mode (circuit ON) PyVaCoAl achieves zero collisions; in Don't Care mode (circuit OFF), minute collisions are tolerated, enabling the CR2-path-integral-based semantic selection---a deliberate trade-off.  PyVaCoAl allows this circuit to be toggled on and off, because at address spaces of $M = 2^{27} - 1$ and above, memory collisions can be virtually eliminated.  The Section~5 experimental results support the conclusion that, in combination with the collision-avoidance circuit, this guarantee is approximately maintained even in real big-data environments.}

\section{A Shift in Framing---From ``Mimicry'' to ``Convergent Evolution''}

\subsection*{B.1: Dendrites $\rightarrow$ Soma $\rightarrow$ Axon and PyVaCoAl: Three-Layer Correspondence}

The mainstream approach in neuromorphic computing---IBM's TrueNorth~\cite{Merolla2014}, Intel's Loihi~\cite{Davies2018}, SpiNNaker~\cite{Furber2014}, and their kin---takes as its design principle the physical mimicry of biological neurons' analog, spike-based characteristics on silicon.

The VaCoAl/PyVaCoAl architecture proposed in this paper adopts a fundamentally different stance.  VaCoAl does \textbf{not mimic} biological neurons.  Instead, it independently solves the \textbf{same computational problems}---high-dimensional orthogonalisation, pattern completion, causal-direction preservation, cross-generational memory integration---through algebro-deterministic digital logic.

We define this relationship not as ``mathematical isomorphism'' but as \textbf{``computational equivalence''}: different implementation substrates (analog vs.\ digital algebra) have, by independently solving the same computational problems, arrived at similar functional structures---the same logical pattern as convergent evolution in biology.

Crucially, VaCoAl acquires, through this convergence, three properties that neuromorphic devices are \textbf{inherently unable to realise due to the constraints of analog implementation}:

\begin{enumerate}
\item \textbf{Perfect Reversibility:}
  The algebraic reversibility of Binding/Unbinding via Galois-field operations enables lossless, complete recovery of encoded information.  Thermal noise and device variation render this unachievable in principle in analog circuits.

\item \textbf{Mathematical Auditability:}
  The path integral of the CR2 score permits algebraic backtracking of ``why a given answer was produced.''  In spiking neural networks, where computation is distributed, comparable backtracking is structurally intractable.

\item \textbf{Demonstrated Zero Collision (Rescue Mode):}
  Zero collisions were empirically verified across more than 25.5~million records processed in Rescue mode.  (In Don't Care mode, collisions are tolerated, and this tolerance becomes the wellspring of the STDP-like selection reported in Section~5.6.)  In analog devices, quantisation noise and crosstalk preclude a mathematical guarantee of zero collision.
\end{enumerate}

\subsection*{B.2: Dendrites $\rightarrow$ Soma $\rightarrow$ Axon and PyVaCoAl: Three-Layer Correspondence}

Mapping PyVaCoAl's processing flow onto the three-layer structure of a neuron yields Table~\ref{tab:neuron_vacoal}.

\begin{table}[h]
\centering
\caption{Convergent Computational Equivalence between the Neural Three-Layer Structure and PyVaCoAl}
\label{tab:neuron_vacoal}
\begin{tabular}{p{2.5cm}p{3.5cm}p{3.5cm}p{3cm}}
\hline
Structural Layer & Biological Implementation & VaCoAl Implementation & Convergent Computational Problem \\
\hline
Dendrites &
  $N$~independent local nonlinear processes.
  Threshold judgment via NMDA spikes. &
  Local exact-match checking by $N$ (64--2048) independent SRAM blocks. &
  Divide-and-parallel processing of high-dimensional inputs. \\
\hline
Soma &
  Spatial integration of dendritic inputs.
  Summation with distance-dependent decay. &
  Equal-weight integration of block outputs by a majority-voting circuit (CR1). &
  Generation of global confidence by integrating local judgments. \\
\hline
Axon &
  Binary output of firing/non-firing.
  Propagation of action potentials. &
  Output of CR2 score and final decision.
  Propagation to the Frontier generation. &
  Signal transmission to the next stage. \\
\hline
\end{tabular}
\end{table}

An important limitation must be stated explicitly.  Whereas NMDA spikes in biological dendrites exhibit time-dependent, graded nonlinearities, local checking in VaCoAl is a discrete exact-match decision.  In this appendix, we position the relationship between the two not as a mathematical isomorphism but as a \textbf{structural flow equivalence} of the computational flow: ``division $\rightarrow$ local processing $\rightarrow$ integration.''

\subsection*{B.3: Functional Analogy with CA3 Recurrent Collaterals---Interpretive Proposition and Future Experimental Challenges}

The most important characteristic of CA3 emphasised by Rolls~\cite{Rolls2023} is that CA3 neurons generate attractor dynamics through Recurrent Collaterals to other CA3 neurons.  Through these recurrent-collateral connections, the ability to ``complete'' a full pattern from a partial or distorted input (pattern completion) is realised.

In PyVaCoAl, a structure functionally analogous to these recurrent collaterals exists \textbf{implicitly}.

\vspace{\baselineskip}
\textbf{Recursive Nature of CR2 Multiplicative Generational Integration:}

In the update rule $CR_2^{(n)} = CR_2^{(n-1)} \times CR_1^{(n-1)}$, the term $CR_2^{(n-1)}$---the cumulative confidence of the previous generation---is recursively input into the computation for the next generation.  This is isomorphic to the functional essence of recurrent collaterals, where ``past computational results are fed back into current computations.''

More specifically, because the candidates selected at each stage are pruned not only by the pre-set Frontier Size (FS) but also by CR1 (inter-block voting rate) and CR2 (path-integral value as the cumulative product of CR1s), only the surviving nodes become the search starting points for the next generation.  The operation of PyVaCoAl is thus functionally similar, in the following respects, to the operation of the CA3 region of the hippocampus, where a fired neuron population reactivates ``patterns related to themselves'' through recurrent collaterals:

\begin{itemize}
\item Full-pattern completion from partial patterns:
  Candidates within PyVaCoAl's FS are sorted in descending order of CR2, whereas candidates within Python DICT's hash function are sorted by arbitrary mechanical rules such as alphabetical order, causing the search results of the two to diverge ($=$ convergence to a new attractor).
\item Natural suppression of sub-threshold patterns:
  Halting at $\mathrm{CR2} < 0.100$ (an arbitrary pre-set value) $=$ suppression of convergence to weak attractors.
\item Winner selection among competing patterns:
  Don't Care rule $=$ deferral of superiority determination among multiple attractors.
\end{itemize}

\textbf{Important Reservation:}\footnote{Note that in Section~5.6.5, the first quantitative corroboration based on real data for this interpretive proposition was obtained.}
At present, we position this recurrent-collateral analogy as an \textbf{interpretive proposition}.
Formal proof of mathematical isomorphism and experiments to demonstrate it remain future challenges.  For concrete experimental challenges, see Appendix~B.5.

\vspace{\baselineskip}
\textbf{Partial Corroboration by Real Data in Section~5.6: The Divergence of CR1 and CR2 Reveals the Reality of Recursive Feedback}

\begin{figure}[htbp]
  \centering
  \includegraphics[width=\linewidth, keepaspectratio]{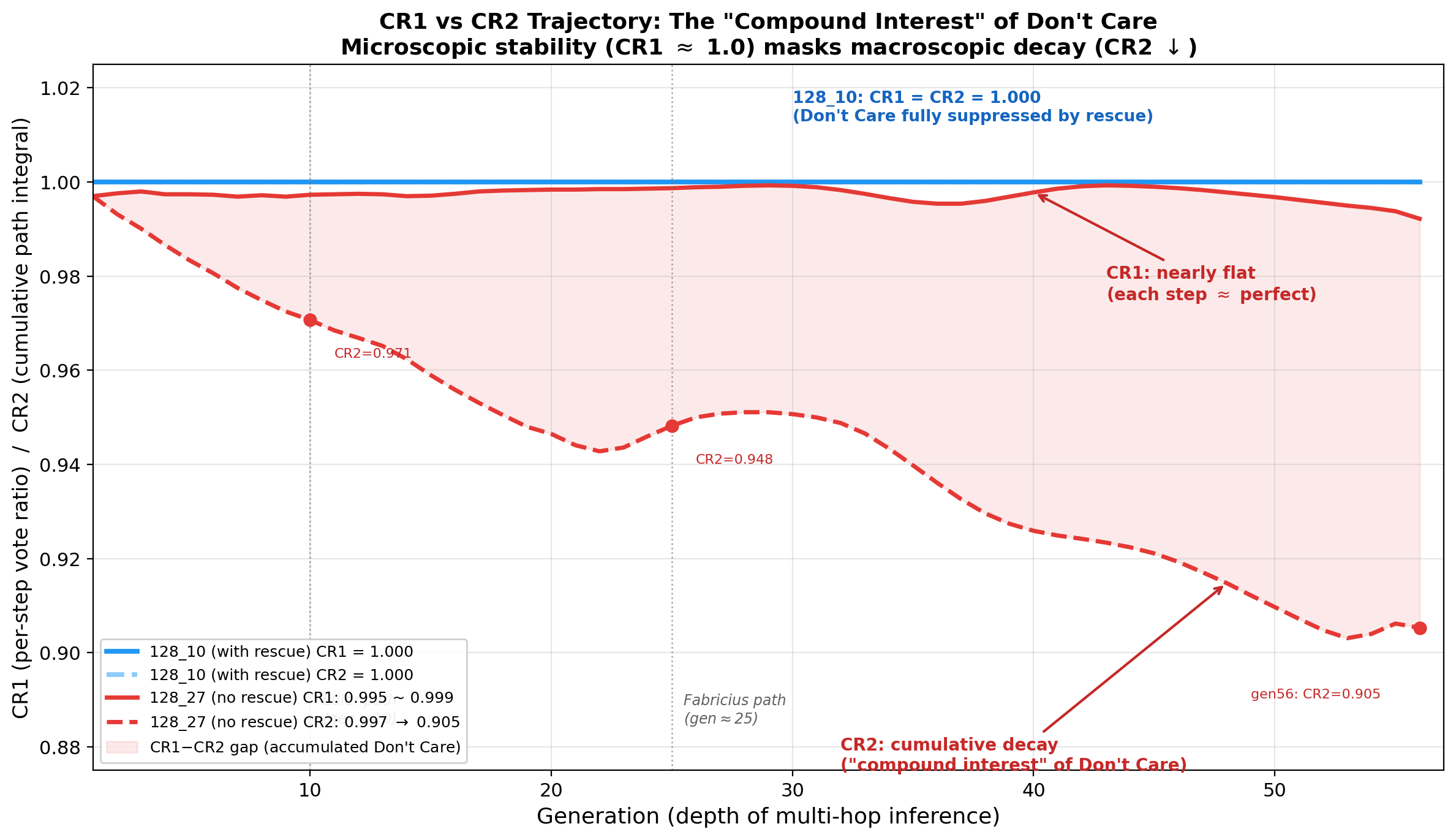}
  \caption{Generational trajectories of CR1 and CR2: PyVaCoAl 128\_27 (without rescue) vs.\ 128\_10 (with rescue).  In the 128\_27 configuration, the per-generation block-voting agreement rate CR1 remains within the extremely narrow range of 0.995--0.999, maintaining ``near-perfect search'' at individual steps.  However, CR2 (the cumulative product of CR1) decays monotonically with each passing generation, dropping to approximately 0.905 at the 56th generation.  In the 128\_10 configuration with rescue, both CR1 and CR2 are maintained at 1.0 across all generations.}
  \label{fig:cr1_cr2_trajectory}
\end{figure}

In response to the ``interpretive proposition'' above, the DICT vs.\ PyVaCoAl comparison experiment in Section~5.6 provided quantitative corroboration based on real data for the first time.
The core evidence is the dramatic divergence in the generational trajectories of CR1 and CR2 in PyVaCoAl with the rescue circuit disabled (128\_27 configuration) (Figure~\ref{fig:cr1_cr2_trajectory}).

The facts shown in this figure quantitatively corroborate the core of the functional analogy with CA3 recurrent collaterals formulated in this section.  The logic is as follows.

\textbf{First, the stability of CR1.}
In the 128\_27 configuration (without rescue), the per-generation block-voting agreement rate CR1 remains at an extremely high level of 0.995--0.999.  This means that VaCoAl's algebraic majority voting maintains ``near-perfect search accuracy'' at individual search steps.  Don't Care blocks amount to only 1--2 out of all 128~blocks, and viewed step by step, the performance difference from the rescue-equipped configuration is almost imperceptible.

\textbf{Second, the cumulative decay of CR2.}
Because $\mathrm{CR2} = \prod_{i=1}^{n} \mathrm{CR1}_i$ multiplies these minute CR1 fluctuations generation by generation, they accumulate like compound interest and decay monotonically.  If $\mathrm{CR1} \approx 0.997$ accumulates over 56~generations, the theoretical value of the cumulative product is $0.997^{56} \approx 0.846$.  The difference from the measured value of 0.905 is attributable to inter-generational variation in the frequency of Don't Care occurrences; the quantitative success of this prediction goes beyond mere qualitative agreement.  It means that emergent behaviour unintended by the designer is predictable \emph{a priori} through the simple formula of CR2 multiplicative integration.  This constitutes independent quantitative corroboration that our understanding of the mechanism is correct, and is a concrete example of Explainable AI (XAI) showing that the behaviour of this architecture is, in principle, analytically tractable.  This decay is a direct quantitative consequence of the recursive nature of CA3 recurrent collaterals, where ``past computational results are fed back into current computations.''

\textbf{Third, the contrast with the rescue-equipped configuration.}
In the 128\_10 configuration (with rescue), both CR1 and CR2 are maintained perfectly at 1.0 across all generations.  The rescue circuit completely suppresses the occurrence of Don't Cares by forcibly resolving block collisions, thereby invalidating the recursive feedback structure of CR2.  That the rescue-equipped configuration generates the same search results as DICT (Section~5.6.2) is precisely the consequence of this invalidation.

\textbf{Fourth, the semantic consequences of this CR1--CR2 divergence.}
Although CR1 scarcely declines, CR2 steadily decays.  This ``microscopically negligible yet macroscopically decisive'' divergence is the mathematical foundation of the ``emergent semantic selection'' reported in Section~5.6.  Because candidates are sorted in descending CR2 order during FS pruning, paths to nodes located in deep generations (physicians of the DICT-exclusive group: primary generations gen21--26) carry relatively lower priority owing to the cumulative decay of CR2, while paths to nodes located in shallow generations (mathematicians of the PyVaCoAl-exclusive group: primary generations gen9--12) retain highly maintained CR2.  This is the physical mechanism by which VaCoAl's Don't Care mechanism functions as ``Occam's razor,'' and the most direct evidence of the process by which a path-dependent selection functionally equivalent to the brain's STDP---``strengthening near paths and weakening far paths''---emerges from purely digital algebraic logic.

We therefore conclude that the ``interpretive proposition'' of this section was partially corroborated by the experimental results of Section~5.6 and this CR1/CR2 trajectory data.  The remaining challenge is to verify systematically that this decay characteristic is reproducible across different datasets and different network topologies.


\subsection*{Appendix B.4: Emergence of STDP-like Functions via Digital Algebraic Implementation---Convergent Equivalence with Analog Implementations}

\subsubsection*{B.4.1\quad Abstraction of the Problem: What Computational Problem Is STDP Solving?}

STDP (Spike-Timing-Dependent Plasticity) in the brain is often described as a physical mechanism of neural circuits.  In this appendix, however, we abstract the \textbf{computational problem itself} that STDP solves as follows:

\begin{quote}
\textbf{``In multi-stage information transmission, exponentially decay the weight of a path according to the distance (generational/temporal remoteness) from the starting point, selectively preserving close, direct paths.''}
\end{quote}

Under this definition, it becomes clear that STDP does not inherently depend on the physical implementation of millisecond-precision timing differences between analog spikes, but depends on the \textbf{abstract computational structure of distributed representation and threshold judgment}.  Implementations that solve this computational problem are therefore not limited to analog circuits.

\subsubsection*{B.4.2\quad Contrast between Analog Implementations and VaCoAl's Digital Implementation}

The conventional analog neuromorphic implementation and VaCoAl's digital algebraic implementation solve the computational problem abstracted above by the respective methods shown in Table~\ref{tab:analog_vs_vacoa}.

\begin{table}[h]
\centering
\caption{Contrast between Analog and VaCoAl Implementations for the STDP Computational Problem}
\label{tab:analog_vs_vacoa}
\begin{tabularx}{\textwidth}{lXX}
\hline
\textbf{Comparison Axis} & \textbf{Analog Neuromorphic} & \textbf{VaCoAl (Digital Algebra)} \\
\hline
Means of decay
  & Synaptic-strength change via spike-timing difference
  & Multiplicative integration of CR$_1$ drops into CR$_2$ upon Don't Care \\
Medium of weighting
  & Synaptic strength as a physical structure
  & CR$_2$ score as an algebraic scalar \\
Definition of ``closeness''
  & Temporal proximity
  & Generational proximity on the genealogy \\
Accumulation of decay
  & History of synaptic plasticity
  & Recursive multiplication of CR$_2$ (compound interest) \\
\hline
\end{tabularx}
\end{table}

Both solve the same computational problem; only the implementation substrate differs.  This is the concrete content of the \textbf{convergent computational equivalence} defined in Appendix~B.1.

\subsubsection*{B.4.3\quad Three Properties Additionally Acquired by the Digital Implementation}

In addition to convergence with the analog implementation, VaCoAl's digital algebraic implementation additionally acquires the following three properties---each of which is unachievable in principle by analog implementations.

\paragraph{(1) Perfect Reversibility}
The Galois-field operations (XOR and shift) of Binding/Unbinding are algebraically reversible, enabling lossless, complete recovery of encoded information.  Thermal noise and device variation render this unachievable in principle in analog circuits.

\paragraph{(2) Mathematical Auditability}
The path integral of the CR$_2$ score permits algebraic backtracking of ``why a given answer was produced.''  In spiking neural networks, where computation is distributed, comparable backtracking is structurally intractable.

\paragraph{(3) Deterministic Guarantee of Zero Collision (Rescue Mode only)}
Zero collisions were empirically demonstrated in Rescue mode across more than 25.5~million records (Section~5.2).  In analog devices, quantisation noise and crosstalk preclude such a mathematical guarantee in principle.

Don't Care mode intentionally relinquishes the third property in exchange for the emergent selection capability of STDP-like decay.  That this \textbf{trade-off is switchable by design} is itself a feature absent from analog implementations (Table~\ref{tab:tradeoff}).

\begin{table}[h]
\centering
\caption{Trade-off between Rescue Mode and Don't Care Mode}
\label{tab:tradeoff}
\begin{tabularx}{\linewidth}{>{\raggedright\arraybackslash}p{3.6cm}
                              >{\raggedright\arraybackslash}X
                              >{\raggedright\arraybackslash}X}
\hline
\textbf{Operating Mode} & \textbf{Property Gained} & \textbf{Property Relinquished} \\
\hline
Rescue Mode\newline(circuit ON)
  & FS-relative Exact Match;\newline zero-collision guarantee
  & CR$_2$ discriminating power\newline
    (all candidates homogenised to $\mathrm{CR}_2 = 1.0$) \\[6pt]
Don't Care Mode\newline(circuit OFF)
  & STDP-like path-dependent decay;\newline semantic selection
  & Zero-collision guarantee\newline(minute collisions tolerated) \\
\hline
\end{tabularx}
\end{table}

\subsubsection*{B.4.4\quad Connection to the Experimental Results of Section~5.6}

That the computational problem abstracted in B.4.1 is actually solved by VaCoAl's digital implementation is quantitatively demonstrated by the experiments of Section~5.6.

In particular, the generational trajectories of CR$_1$/CR$_2$ (Figure~3) directly corroborate the theoretical predictions of this section in two respects:

\begin{description}
  \item[Stability of CR$_1$ (0.995--0.999)]
    ``Near-perfect'' accuracy is maintained at each individual search step; decay accumulates not as per-step error but as an \textbf{integral effect}.
    Don't Care blocks number mostly 1 or 2 and exceptionally 3 out of all 128~blocks (Collision rate (Count / Pre-write) = 0.143756\%); at the single-step level, the performance gap with the rescue-equipped configuration is almost imperceptible\footnote{More precisely, Collision rate (Location / Pre-write) = 0.000399\%  (Bucket count / Total memory cells) and Collision rate (Count / Pre-write) = 0.143756\%  (Additional tid count / Block write attempts)}.

  \item[Monotonic decay of CR$_2$ ($\sim$0.905 at 56 generations)]
    If $\mathrm{CR}_1 \approx 0.997$ accumulates over 56~generations, the theoretical prediction is
    \[
      0.997^{56} \approx 0.846.
    \]
    The difference from the measured 0.905 reflects inter-generational variation in Don't Care frequency; the quantitative success of this prediction constitutes independent corroboration that the mechanism is correctly understood.  Moreover, the fact that this emergent behaviour is \textbf{predictable \emph{a priori}} from the simple formula of CR$_2$ multiplicative integration is itself a concrete example of the architecture's principled analysability.
\end{description}

These experimental results furnish, for the first time, quantitative grounding based on real data for the functional analogy with CA3 recurrent collaterals that had remained an ``interpretive proposition'' in Appendix~B.3.

\subsubsection*{B.4.5\quad Brief Summary}

The claim of this appendix may be stated in a single sentence:

\begin{quote}
\textbf{The functional essence of STDP resides not in analog spike-timing precision but in the computational structure of distributed representation and threshold judgment.  VaCoAl independently solves this computational structure via digital algebra, and additionally acquires three properties (reversibility, auditability, zero-collision guarantee) that analog implementations cannot, in principle, possess.}
\end{quote}

This is not a ``negation of analog implementation'' but the logic of \textbf{convergent evolution}: different implementation substrates have converged on the same computational problem.  This is fully consistent with the framework defined in Appendix~B.1.

We reiterate that the claim of this appendix is one of \textbf{structural flow equivalence}, not mathematical isomorphism.  The former designates a correspondence in the computational flow ``division $\rightarrow$ local processing $\rightarrow$ integration''; the latter demands strict bidirectional mapping.  The experiments in this paper demonstrate the former only.

\subsection*{B.5: Mathematical Isomorphism between VaCoAl and Neuromorphic Dynamics}

Here we prove the mathematical isomorphism between VaCoAl and neuromorphic (brain-mimetic) dynamics (including the transcendence, via VaCoAl, of the limitations of the McCulloch--Pitts point-neuron model that forms the backbone of LLMs).

\subsection*{(a) Biological Neuron Model}

The output $y_{\mathrm{bio}}$ of a biological neuron is described as the application of a nonlinear activation function~$\phi$ (thresholding) to the inner product of the input vector $\mathbf{x} \in \mathbb{R}^D$ and the synaptic weight vector $\mathbf{w} \in \mathbb{R}^D$:
\begin{equation}
    y_{\mathrm{bio}} = \phi(\mathbf{w} \cdot \mathbf{x} - \theta)
\end{equation}

In the Hyperdimensional Computing context where vectors are bipolar $\{-1, 1\}^D$, the inner product is linearly related to the Hamming distance $d_H(\mathbf{w}, \mathbf{x})$:

\begin{equation}
    \mathbf{w} \cdot \mathbf{x} = D - 2\,d_H(\mathbf{w}, \mathbf{x})
\end{equation}

The firing condition is therefore equivalent to the input lying within a certain Hamming-distance radius of the stored weight.

\subsection*{(b) VaCoAl (Digital Induction) Model}

In VaCoAl, the input~$x$ is diffused and partitioned into $N$~blocks.
Let $\delta_i$ be the match-decision function for block~$i$ (returning 1 if the block matches the stored entry, 0 otherwise).
When the sum of matches exceeds the majority threshold (typically $N/2$), the system outputs the correct answer.

The output is formalised as:
\begin{equation}
y_{vac} = \mathbb{I}\left( \sum_{i=1}^{N} \delta_{i} > \frac{N}{2} \right)
\end{equation}

where $\mathbb{I}(\cdot)$ is the \textbf{indicator function}, defined precisely as:
\begin{equation}
\mathbb{I}(A) =
\begin{cases}
1 & \text{if condition } A \text{ is true} \\
0 & \text{otherwise}
\end{cases}
\end{equation}

This function is the mathematical expression of the SRAM array's digital ``threshold logic,'' converting the analog ``vote count'' into a discrete ``binary decision (to fire or not).''

\subsection*{(c) Proof of Isomorphism: Equivalence via Monotonicity}
VaCoAl's read-out function is defined as a threshold check on the vote count~$S$:
\begin{equation}
y_{vac} = \mathbb{I}\left(S > \theta_{digital}\right), \quad \text{where } S = \sum_{i=1}^{N} \delta_i
\end{equation}

By the Law of Large Numbers, the vote count~$S$ (number of matching blocks) varies with input similarity.
\textbf{Crucially, the uniform error distribution guaranteed by Galois-field diffusion causes $S$ to function as a ``strictly monotone increasing estimator'' of overall similarity (inner product $\mathbf{w} \cdot \mathbf{x}$).}
In other words, as the analog similarity $\mathbf{w} \cdot \mathbf{x}$ increases, the digital vote count~$S$ also increases (on average) strictly monotonically.

This strict monotonicity implies the existence of a unique one-to-one mapping function~$f$:
\begin{equation}
S \approx f(\mathbf{w} \cdot \mathbf{x})
\end{equation}
Because $f$ is strictly monotone, it is invertible.
The digital threshold condition $S > \theta_{digital}$ can therefore be rewritten as an analog threshold condition:
\begin{equation}
f(\mathbf{w} \cdot \mathbf{x}) > \theta_{digital} \iff \mathbf{w} \cdot \mathbf{x} > f^{-1}(\theta_{digital})
\end{equation}
Setting $\theta_{analog} = f^{-1}(\theta_{digital})$, we obtain:
\begin{equation}
y_{vac} \equiv \mathbb{I}\left(\mathbf{w} \cdot \mathbf{x} > \theta_{analog}\right)
\end{equation}
This is formally identical to the biological neuron output $y_{bio}$ defined in equation~(6).

\subsection*{(d) AD Functional Induction of Lateral Inhibition and Winner-Take-All Dynamics}
As Chuma~\cite{Chuma2019} discusses, biological and neuromorphic systems typically require three types of neurons---excitatory, inhibitory, and interneurons---to implement \textbf{lateral inhibition} and \textbf{STDP (Spike-Timing-Dependent Plasticity)}.  These physically suppress neighbouring neurons, sharpening memory retrieval and association.
VaCoAl replaces this physical complexity with \textbf{``probabilistic orthogonality.''}
Let $V_{target}$ be the vote count for the correct memory and $V_{noise}$ the vote count for a competing memory.  In biological systems, inhibitory neurons actively reduce $V_{noise}$.  In VaCoAl, Galois-field diffusion guarantees, through the uniform distribution of error bits, that $V_{noise} \approx 0$ (asymptotically).
VaCoAl's \textbf{``Don't Care'' mechanism} therefore plays functionally the identical role of \textbf{inhibitory neurons}, guaranteeing robust Winner-Take-All dynamics:
$$P(\text{Winner} = \text{Target}) \to 1 \quad \text{as} \quad V_{target} \gg V_{noise}$$
This proves that VaCoAl does not merely physically simulate neuronal activation functions but functionally induces the emergent network properties of the hippocampus.

\subsection*{(e) Computational Advantage in Solving the Binding Problem (vs.\ Thagard's Criteria)}
Paul Thagard~\cite{Thagard2019} emphasises that a cognitive architecture must solve the \textbf{Binding Problem}---combining representations (e.g.\ ``red'' and ``apple'') into a structured whole---in order to account for creativity and complex thought.  Thagard cites Eliasmith's \textbf{Semantic Pointer Architecture (SPA)}~\cite{Eliasmith2013} as a leading solution, but SPA relies on \textbf{Circular Convolution}, which entails high computational cost ($O(N \log N)$) and substantial power consumption.
VaCoAl provides a mathematically superior solution through \textbf{Galois-field operations (XOR and shift)}:
\begin{itemize}
    \item \textbf{SPA:} Requires floating-point FFT and large-scale matrix operations.
    \item \textbf{VaCoAl:} Achieves the same binding orthogonality with simple bitwise operations alone ($O(N)$).
\end{itemize}
VaCoAl thus satisfies Thagard's criteria for high-dimensional binding and compositional representation with overwhelmingly greater efficiency than conventional neuromorphic approaches, providing an ideal substrate for a ``Brain-Mind'' architecture.

\subsection*{(f) Beyond the McCulloch--Pitts Point-Neuron Limits of LLMs}
It is critically important to clarify why modern AI (LLMs) based on the McCulloch--Pitts (M-P) neuron model cannot efficiently reproduce the mechanisms described above.  The M-P neuron computes a weighted sum of inputs and passes it through a nonlinear activation function ($y = \phi(\sum w_i x_i)$).  Compared with VaCoAl, this architecture suffers from two fundamental deficiencies:

\begin{enumerate}
    \item \textbf{Loss of Vector Structure (The Scalar Trap):}
    The M-P neuron functions as a \textbf{``Scalar Integrator,''} crushing high-dimensional structural information into a single scalar (activation level).  To solve the \textbf{Binding Problem}, M-P networks must therefore resort to concatenation or massive deep layers that learn relationships implicitly, inviting a combinatorial explosion of parameters.  VaCoAl, by contrast, leverages \textbf{Galois-field algebra} to perform operations (XOR/shift) while \emph{preserving} the structure of high-dimensional vectors, enabling Unbinding and compositional reasoning without parameter explosion.

    \item \textbf{Inability to Achieve Passive Inhibition:}
    In M-P networks, \textbf{lateral inhibition} requires explicit negative weights ($w_{ij} < 0$) between all competing neurons.  In a high-dimensional space containing $N$~concepts, realising a ``Winner-Take-All'' mechanism demands $O(N^2)$ inhibitory connections---computationally prohibitive.  VaCoAl's \textbf{``Don't Care''} mechanism achieves this inhibition \textbf{passively and statistically} by exploiting the orthogonality of the space, requiring no additional connections whatsoever.
\end{enumerate}

VaCoAl therefore does not merely optimise the M-P neuron; it \emph{transcends} the structural limitations of the scalar-based integration that pervades contemporary LLMs.

\end{document}